\documentclass[10pt,journal,compsoc]{IEEEtran}
\ifCLASSOPTIONcompsoc
  \usepackage[nocompress]{cite}
\else
  \usepackage{cite}
\fi
\ifCLASSINFOpdf
   \usepackage[pdftex]{graphicx}
\else
\fi
\usepackage[dvipsnames]{xcolor}

\usepackage{amsmath}
\usepackage{amsthm}
\usepackage{amsfonts}

\usepackage{amsmath,amsfonts,bm, mathrsfs}
\usepackage{amssymb}

\def\1{\bm{1}}

\DeclareMathAlphabet{\mathsfit}{\encodingdefault}{\sfdefault}{m}{sl}
\SetMathAlphabet{\mathsfit}{bold}{\encodingdefault}{\sfdefault}{bx}{n}

\def\gA{{\mathcal{A}}}

\def\gN{{\mathcal{N}}}

\def\gX{{\mathcal{X}}}
\def\gY{{\mathcal{Y}}}

\def\sR{{\mathbb{R}}}

\newcommand{\E}{\mathbb{E}}

\newcommand{\indep}{\perp \!\!\! \perp}
\DeclareMathOperator*{\argmax}{arg\,max}

\usepackage{tablefootnote}
\usepackage{multirow}
\usepackage[caption=false,font=footnotesize,labelfont=sf,textfont=sf]{subfig}
\usepackage{enumitem}
\usepackage{soul}

\usepackage{float}
\newcommand{\myparagraph}[1]{\smallskip\noindent\textbf{#1}}

\newtheorem{proposition}{Proposition}
\newtheorem{lemma}{Lemma}

\begin{document}
%
\title{Bare Demo of IEEEtran.cls for\\ IEEE Computer Society Journals}
\title{Information Pursuit: An Information Theoretic Approach to Learning Interpretable Representations}
\title{Learning Interpretable Predictive Representations via Information Pursuit}
\title{Pursuing Short Explanations: A Framework for Interpretable Deep Learning}
\title{Pursuing Minimal-Length Explanations: A Framework for Interpretable Deep Learning}
\title{Learning Explainable Representations via Information Pursuit}
\title{Learning Explainable Predictors by Composing
Explainable Queries with Information Pursuit}

\title{Explainable by Design: Composing Interpretable Concepts for Explainable Models}
\title{Explainable by Design: Predictors from Explainable Queries}
\title{Designing Explainable Predictors by Composing Explainable Queries}
\title{Explainable by Design: Predictors from Explainable Queries}
\title{Composing Explainable Queries for Explainable Predictors}
\title{Explainable by Design: Composing Explainable Queries for Explainable Predictors}

\title{Interpretable by Design: Learning Predictors by 
Composing Interpretable Queries}

%
%
%
%

\author{Aditya Chattopadhyay,
        Stewart Slocum, 
        Benjamin D. Haeffele,\\
        Ren\'e Vidal,~\IEEEmembership{Fellow,~IEEE} and
        Donald Geman,~\IEEEmembership{Life Senior Member,~IEEE}
\IEEEcompsocitemizethanks{\IEEEcompsocthanksitem 
The authors are with the Mathematical Institute for Data Science and the Center for Imaging Science of The Johns Hopkins University, MD, 21218.
E-mail: \{achatto1, sslocum3, bhaeffele, rvidal, geman\}@jhu.edu
}}

%
%

\markboth{Journal of \LaTeX\ Class Files,~Vol.~14, No.~8, August~2015}%
{Shell \MakeLowercase{\textit{et al.}}: Bare Demo of IEEEtran.cls for Computer Society Journals}
%



\IEEEtitleabstractindextext{%
\begin{abstract}
There is a growing concern about typically opaque decision-making with high-performance machine learning algorithms. Providing an explanation of the reasoning process in domain-specific terms can be crucial for adoption in risk-sensitive domains such as healthcare. We argue that machine learning algorithms should be interpretable by design and that the language in which these interpretations are expressed should be domain- and task-dependent. Consequently, we base our model’s prediction on a family of user-defined and task-specific binary functions of the data, each having a clear interpretation to the end-user. We then minimize the expected number of queries needed for accurate prediction on any given input. As the solution is generally intractable, following prior work, we choose the queries sequentially based on information gain. However, in contrast to previous work, we need not assume the queries are conditionally independent. Instead, we leverage a stochastic generative model (VAE) and an MCMC algorithm (Unadjusted Langevin) to select the most informative query about the input based on previous query-answers. This enables the online determination of a query chain of whatever depth is required to resolve prediction ambiguities. Finally, experiments on vision and NLP tasks demonstrate the efficacy of our approach and its superiority over post-hoc explanations.

\end{abstract}

}

\maketitle

\IEEEdisplaynontitleabstractindextext

%
\IEEEpeerreviewmaketitle

\section{Introduction}
\label{sec:introduction}

\IEEEPARstart{I}{n} recent years, interpreting large machine learning models has emerged as a major priority, particularly for transparency in making decisions or predictions that impact human lives \cite{gunning2019xai, rudin2019stop, murdoch2019interpretable}. In such domains, understanding \textit{how} a prediction is made may be as important as achieving high predictive accuracy.  For example, medical regulatory agencies have recently emphasized the need for computational algorithms used in diagnosing, predicting a prognosis, or suggesting treatment for a disease, to explain why a particular decision was made \cite{eureg2019,FDAminutes2021}. 

On the other hand, it is widely believed that there exists a fundamental trade-off in machine learning between interpretability and predictive performance  \cite{johansson2011trade,wanner2021stop,dovsilovic2018explainable,arrieta2020explainable,gunning2019darpa}. Simple models like decision trees and linear classifiers are often regarded as \textit{interpretable}\footnote{Although later in the paper we will discuss situations in which even these simple models need not be interpretable.} but at the cost of potentially reduced accuracy compared with larger \textit{black box} models such as deep neural networks. 
As a result, considerable effort has been given to developing methods that provide \textit{post-hoc} explanations of black box model predictions, i.e., given a prediction from a (fixed) model provide additional annotation or elaboration to explain how the prediction was made. As a concrete example, for image classification problems, one common family of post-hoc explanation methods produces attribution maps which seek to estimate the regions of the image that are \textit{most important} for prediction. This is typically approached by attempting to capture the effect or sensitivity of perturbations to the input (or intermediate features) on the model output \cite{baehrens2010explain, simonyan2013deep, kolek2022rate, shrikumar2017learning, zeiler2014visualizing, selvaraju2017grad, smilkov2017smoothgrad, subramanya2019fooling}. 
However, post-hoc analysis has been critiqued for a variety of issues
\cite{adebayo2018sanity, yang2019benchmarking,kindermans2019reliability, shah2021input, slack2020fooling,rudin2019stop} (see also \S \ref{sec:related_work})
%
and often fails to provide explanations in terms of concepts that are intuitive or interpretable for humans~\cite{koh2020concept}. 

This naturally leads to the question of what an \textit{ideal} explanation of a model prediction would entail; however, this is potentially highly \textit{task-dependent} both in terms of the task itself as well as what the user seeks to obtain from an explanation.  For instance, a model for image classification is often considered interpretable if its decision can be explained in terms of patterns occurring in salient parts of the image \cite{chen2019looks} (e.g., the image is a car because there are wheels, a windshield, and doors), whereas in a medical task explanations in terms of causality and mechanism could be desired (e.g., the patient's chest pain and shortness of breath is likely not a pulmonary embolism because the blood D-dimer level is low, suggesting thrombosis is unlikely). Note that some words or patterns may be \textit{domain-dependent} and therefore not interpretable to non-experts, and hence what is interpretable ultimately depends on the end user, namely the person who is trying to understand or deconstruct the decision made by the algorithm \cite{rudin2022interpretable}.

In addition to this \textit{task-dependent} nature of model interpretation, there are several other desirable intuitive aspects of interpretable decisions that one can observe.  The first is that meaningful interpretations are often \textit{compositional} and can be constructed and explained from a set of \textit{elementary units} \cite{janssen1997compositionality}. For instance, words, parts of an image, or domain-specific concepts \cite{lakkaraju2016interpretable, wan2021nbdt, mu2020compositional} could all be a suitable basis to form an explanation of a model's prediction depending on the task. Moveover, the basic principle that \text{simple} and \textit{concise} explanations are preferred (i.e., Occam's razor) suggests that interpretablity is enhanced when an explanation can be composed from the smallest number of these elementary units as possible. Finally, we would like this explanation to be \textit{sufficient} for describing model predictions, meaning that there should be no external variables affecting the prediction that are not accounted for by the explanation.

\begin{figure}[t]
\centering
\includegraphics[width=1.0\linewidth]{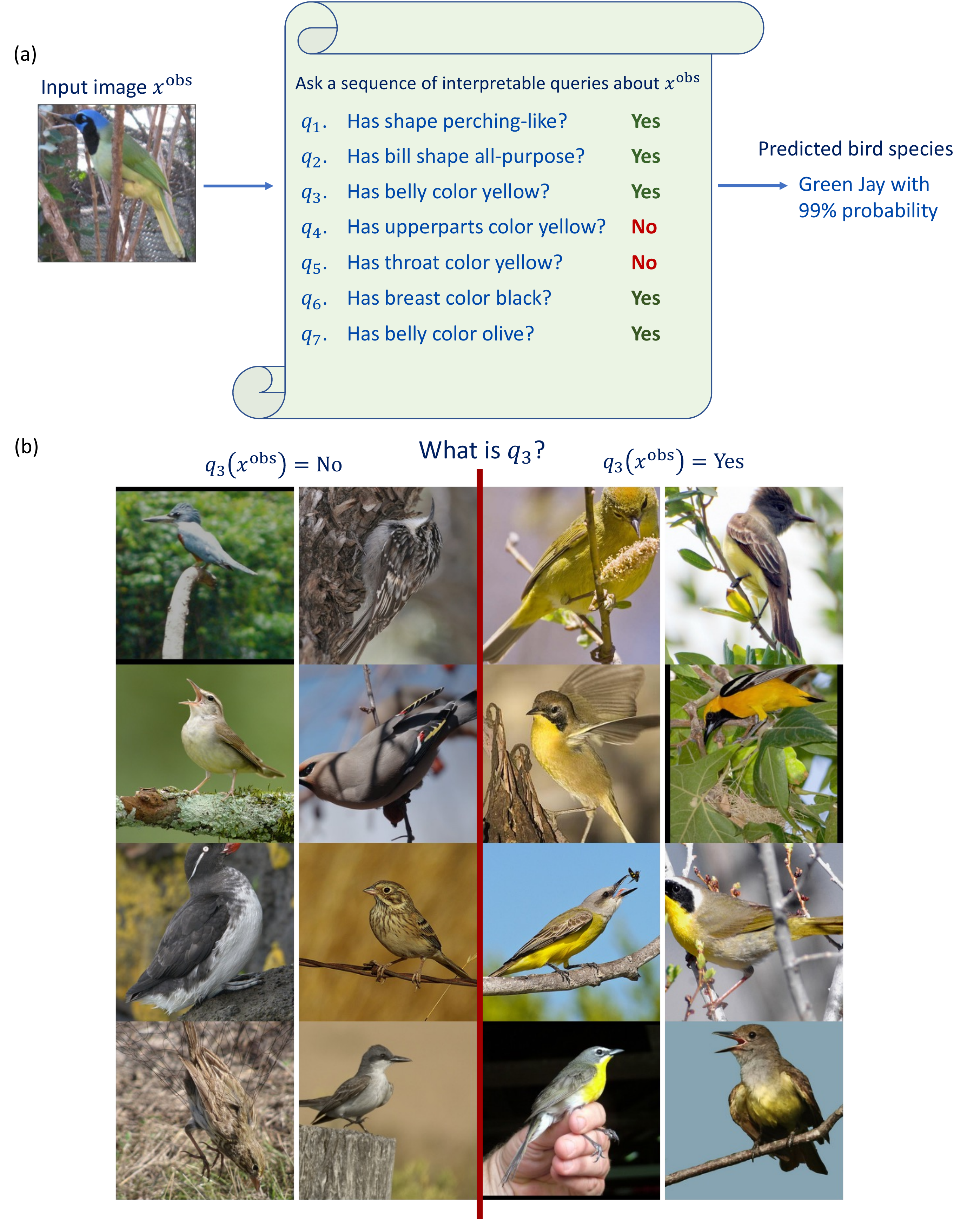}
\caption{\textbf{(a) An illustration of our proposed learning framework.} The prediction of a bird species is explained through a short sequence of interpretable queries, $(q_1, q_2, ..., q_7)$, derived from a user-defined query set of domain-specific attribute for birds. \textbf{(b) Interpretable queries.} Each query in this case corresponds to a well-defined bird attribute. For instance, $q_3$ asks ``Does the bird have belly color yellow?". We visualize some example images which evaluate to ``Yes" and observe that all of them correspond to birds with a yellow belly. Similarly, all images which evaluate to ``No" corresponds to birds which do not have a yellow belly.}
\label{Fig: teaser}
\vspace{-1mm}
\end{figure}

Inspired by these desirable properties, we propose a framework for learning predictors that are \textit{interpretable by design}. The proposed framework is based on composing a subset of user-defined \textit{concepts}, i.e., functions of the input data which we refer to as \textit{queries}, to arrive at the final prediction. 
Possible choices for the set of queries $Q$ based on the style of interpretation that is desired include:
\begin{enumerate}
\item \textbf{Salient image parts:}
For vision problems, if one is interested in explanations in terms of salient image regions then this can be easily accomplished in our framework by defining the query set to be a collection of small patches (or even single pixels) within an image. This can be thought of as a generalization of the pixel-wise explanations generated by attribution maps.

\item \textbf{Concept-based explanations:}
In domains such as medical diagnosis or species identification, the user might prefer explanations in terms of concepts identified by the community to be relevant for the task. For instance, a ``Crow" is determined by the shape of the beak, color of the feathers, etc. In our framework, by simply choosing a query for each such concept, the user can easily obtain concept-based explanations (see Fig. \ref{Fig: teaser}(b)).

\item \textbf{Visual scene interpretation:}
In visual scene understanding, one seeks a rich semantic description of a scene by accumulating the answers to queries about the existence of objects and relationships, perhaps generating a scene graph \cite{jahangiri2017information}. One can design a query set $Q$ by instantiating these queries with trained classifiers. The answers to chosen queries in this context would serve as a semantic interpretation of the scene. 

\item \textbf{Deep neuron-based explanations:} The above three examples are query sets based on domain knowledge. Recent techniques \cite{mu2020compositional, bau2017network,hernandez2022natural} have shown the ability of different neurons in a trained deep network to act as concept detectors. These are learnt from data by solving auxiliary tasks without any explicit supervisory signal. One could then design a $Q$ in which each query corresponds to the activation level of a specific concept neuron. Such a query set will be useful for tasks in which it is difficult to specify interpretable functions/queries beforehand. 
\end{enumerate}

Given a user-specified set of queries $Q$, our framework makes its prediction by selecting a short sequence of queries such that the sequence of query-answer pairs provides a complete explanation for the prediction.
More specifically, the selection of queries is done by first learning a generative model for the joint distribution of queries and output labels and then using this model to select the ``most informative" queries for a given input. The final prediction is made using the Maximum A Posteriori (MAP) estimate of the output given these query-answer pairs. Fig. \ref{Fig: teaser}(a) gives an illustration of our proposed framework, where the task is to predict the bird species in an image and the queries are based on color, texture and shape attributes of birds. We argue that the sequence of query-answer pairs provides a meaningful explanation to the user that captures the subjective nature of interpretability depending on the task at hand, and that is, by construction, compositional, concise and sufficient.    


\begin{figure}[t]
\centering
\includegraphics[width=1.0\linewidth]{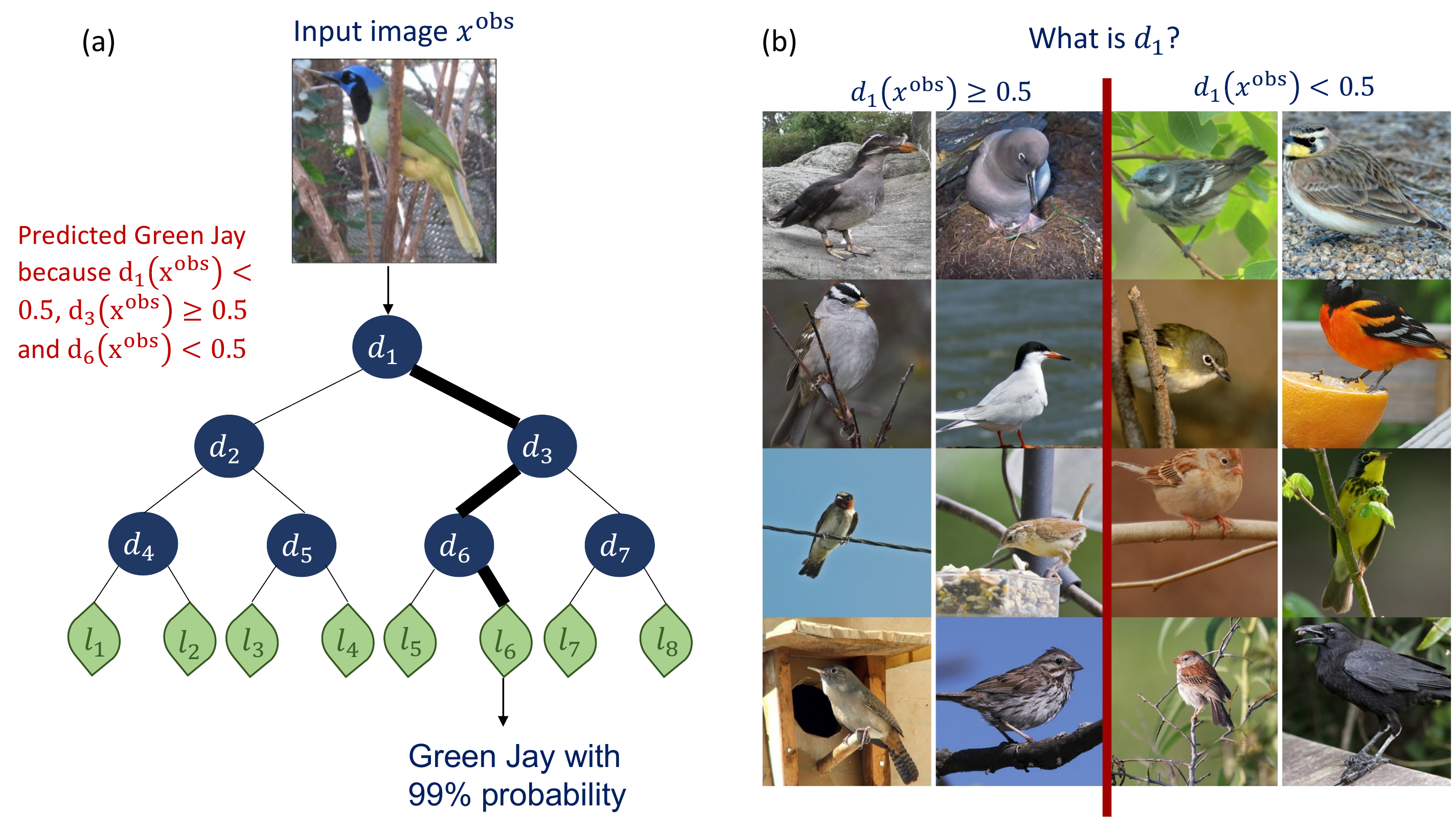}
\caption{\textbf{The interpretability of an explanation depends on how interpretable the queries are.} \textbf{(a)} An illustration of a Deep Neural decision tree \cite{kontschieder2015deep} trained on the CUB-2011 dataset of bird images. The bold path denotes the trajectory the input image $x^{\textrm{obs}}$ takes through the tree. Each $d_i$ corresponds to an internal node of the tree and is a black-box function/query learnt from data. Each $l_i$ denotes a leaf and computes the final classification for $x^{\textrm{obs}}$. The prediction can be explained as a conjunction of internal node functions, but is it really interpretable? \textbf{(b)} Example images that get routed to the left sub-tree ($d_1 \geq 0.5$) and right sub-tree ($d_1 < 0.5$) of the root node. Notice that the interpretation of $d_1$ is not clear from these examples. Compare this to Fig. \ref{Fig: teaser} where the semantics of each query is unambiguous to the end-user.}
\label{Fig: non-interpretable query set}
\end{figure}

At first glance, one might think that classical decision trees \cite{breiman2017classification, quinlan1986induction} based on $Q$ could also produce interpretable decisions by design. However, the classical approach to determining decision tree branching rules based on the empirical distribution of the data is prone to over-fitting due to data fragmentation.
Whereas random forests \cite{amit1997shape, breiman2001random} are often much more competitive than classical decision trees in accuracy \cite{caruana2006empirical,fernandez2014we,xu2021deep}, they sacrifice interpretability, the very property we want to hardwire into our decision algorithm. Similarly, the accuracy of a single tree can be improved by using deep networks to learn queries directly from data, as in Neural Decision Trees (NDTs) \cite{kontschieder2015deep}. However, the opaqueness of the interpretation of these learnt queries makes the explanation of the final output, in terms of  logical operations on the queries at the internal nodes, unintelligible. Figure \ref{Fig: non-interpretable query set} illustrates this with an example.

In this paper we make the following contributions;
\begin{itemize}
    \item We propose a novel framework for prediction that is \emph{interpretable by design}.  We allow the end-user to specify a set $Q$ of queries about input $X$ and formulate learning as the problem of selecting a minimal set of queries from $Q$ whose answers are sufficient for predicting output $Y$. We formulate this query selection problem as an optimization problem over strategies that minimize the number of queries needed on average to predict $Y$ from $X$. A prediction for $Y$ is then made based on the selected query-answer pairs, which provide an explanation for the prediction that is by construction interpretable. The set of selected query-answer pairs can be viewed as a \textit{code} for the input. However, a major difference between our framework and coding theory is that, due to the constraint of interpretability, $Q$ is a vanishingly small collection of the functions of $X$, whereas coding theory typically considers $Q$ to be all possible binary functions of $X$.

\item Since computing the exact solution to our optimization problem is computationally challenging, we propose to greedily select a minimal set of queries by using the \textit{Information Pursuit} (IP) algorithm \cite{geman1996active}. IP sequentially selects queries in order of maximum \emph{information gain} until enough evidence is gathered from the query-answer pairs to predict $Y$. This sequence of query-answer pairs serves as the \textit{explanation} for predicting $Y$ from $X$. To ameliorate the computational challenge of computing information gain for high-dimensional input and query spaces, prior work \cite{geman1996active} had assumed that query answers were conditionally independent given $Y$, an assumption that is largely inadequate for most prediction tasks we encounter in practice. In this paper, we propose a latent variable graphical model for the joint distribution of queries and outputs, $p(Q(X),Y)$, and learn the required distributions using Variational Autoencoders (VAEs). We then use the Unadjusted Langevin Algorithm (ULA) to generate samples required to carry out IP. This gives us a tractable algorithm for any task and query set. To the best of our knowledge, ours is the first implementation of IP that uses deep generative models and does \textit{not} assume that query answers are conditionally independent given $Y$.

\item Finally, we demonstrate the utility of our framework on various vision and NLP tasks. In binary image classification using MNIST, Fashion-MNIST \& KMNIST, and bird species identification using CUB-200, we observe that IP finds succinct explanations which are highly predictive of the class label. We also show, across various datasets, that the explanations generated by our method are shorter and more predictive of the class label than state-of-the-art post-hoc explanation methods like Integrated Gradients and DeepSHAP.
\end{itemize}

\section{Related Work}
\label{sec:related_work}
Methods for interpretable deep learning can be separated into those that seek to explain existing models (post-hoc methods) and those that build models that are interpretable by design. Because they do not negatively impact performance and are convenient to use, post-hoc explanations have been the more popular approach, and include a great diversity of methods.

Saliency maps estimate the contribution of each feature through first-order derivatives \cite{baehrens2010explain, simonyan2013deep, chattopadhay2018grad, selvaraju2017grad, smilkov2017smoothgrad}. Linear perturbation-based methods like LIME \cite{ribeiro2016should} train a linear model to locally approximate a deep network around a particular input, and use the coefficients of this model to estimate the contribution of each feature to the prediction. Another popular set of methods use game-theoretic Shapley values as attribution scores, estimating feature contributions by generating predictions on randomly sampled subsets of the input \cite{lundberg2017unified}. We provide quantitative comparisons between IP and these methods in Section \ref{section: comparison to attribution methods}. Recently, there has been interest in concept-based analogues of these methods that leverage similar approaches to measure the sensitivity of a prediction to high-level, human-friendly concepts as opposed to raw features \cite{kim2018interpretability, zhou2018interpretable, yeh2020completeness}.

Despite certain advantages, what all the above post-hoc methods have in common is that they come with little guarantee that the explanations they produce actually reflect how the model works \cite{rudin2019stop}. Indeed, several recent studies \cite{adebayo2018sanity,yang2019benchmarking, kindermans2019reliability, shah2021input, subramanya2019fooling} call into question the veracity of these explanations towards the trained model. Adebayo \emph{et al.} \cite{adebayo2018sanity} show that several popular attribution methods act similar to edge detectors and are insensitive to the parameters of the model they attempt to explain! Yang \emph{et al.}\cite{yang2019benchmarking} find that these methods often produce false-positive explanations, assigning importance to features that are irrelevant to the prediction of the model. It is also possible to adversarially manipulate post-hoc explanations to hide any spurious biases the trained model might have picked up from data \cite{slack2020fooling}. 

\myparagraph{Interpretability by design.} These issues have motivated recent work on deep learning models which are \textit{interpretable by design}, i.e., constrained to produce explanations that are faithful to the underlying model, albeit with varying conceptions
of ``faithfulness". Several of these models are constructed so they behave similarly to or can be well-approximated by a classically interpretable model, such as a linear classifier \cite{alvarez2018towards, bohle2021convolutional} or a decision tree \cite{wu2021optimizing}. This allows for an approximately faithful explanation in raw feature space. In a similar vein, Pillai \& Pirsiavash \cite{pillai2021explainable} fix a post-hoc explanation method (e.g. Grad-CAM \cite{selvaraju2017grad}), and regularize a model to generate consistent explanations with the chosen post-hoc method. However, our method does not just behave \textit{like} a fully interpretable model or generate \textit{approximately} faithful explanations, but rather it produces explanations that are guaranteed to be faithful and fully explain a given prediction.

Another approach to building interpretable models by design is to generate explanations in terms of high-level, interpretable concepts rather than in raw feature space, often by applying a linear classifier to a final latent space of concepts \cite{chen2019looks, alvarez2018towards, chen2020concept}. However these concepts are learned from data, and may not align with the key concepts identified by the user. For example, Prototypical Part Networks \cite{chen2019looks} take standard convolutional architectures and insert a ``prototype layer" before the final linear layer, learning a fixed number of visual concepts that are used to represent the input. This allows the network to explain a prediction in terms of these ``prototype" concepts. Since these prototypes are learned embeddings, there is no guarantee that their interpretation will coincide with the user's requirements. Furthermore, these explanations may require a very large number of concepts, while in contrast, we seek minimal-length explanations to preserve interpretability. 

Attention-based models are another popular family of models that are sometimes considered interpretable by design \cite{de2019bias, galassi2020attention}. However, attention is only a small part of the overall computation and can be easily manipulated to hide model biases \cite{pruthi2019learning}. Moreover, the attention coefficients are not necessarily a sufficient statistic for the model prediction.

Perhaps most similar to our work are Concept Bottleneck Networks \cite{koh2020concept}, which first predict an intermediate set of human-specified concepts $c$ and then use $c$ to predict the final class label. Nevertheless, the learnt mapping from concepts to labels is still a black-box. To remedy this, the authors suggest using a linear layer for this mapping but this can be limiting since linearity is often an unrealistic assumption \cite{janssen1997compositionality}. In contrast, our framework makes no linearity assumptions about the final classifier and the classification is explainable as a sequence of interpretable query-answer pairs obtained about the input (see Fig. \ref{Fig: teaser}(a)).

\myparagraph{Neural networks and decision trees.} Unlike the above methods, which can be thought of as deep interpretable linear classifiers, our method can be described as a deep decision tree that branches on responses to an interpretable query set. Spanning decades, there has been a variety of work building decision trees from trained neural networks \cite{craven1995extracting, dancey2004decision, frosst2017distilling, wan2021nbdt} and using neural networks within nodes of decision trees \cite{kontschieder2015deep, roy2016monocular, biccici2018conditional, murthy2016deep}. Our work differs from these in three important aspects. First, rather than allowing arbitrary splits, we branch on responses to an interpretable query set. Second, instead of using empirical estimates of information gains based on training data (which inevitably encounter data-fragmentation \cite{vilalta1997global} and hence overfitting), or using heuristics like agglomerative clustering on deep representations \cite{wan2021nbdt}, we calculate information gain from a generative model, leading to strong generalization. Third, for a given input, say $x^{\textrm{obs}}$, we use a generative model to compute the queries along the branch traversed by $x^{\textrm{obs}}$ in an online manner. The entire tree is never constructed. This allows for much very deep terminal nodes when necessary to resolve ambiguities in prediction. As an example, for the task of topic classification using the HuffPost dataset (\S \ref{Section: Word-based Queries}), our framework asks about $199$ queries (on average) before identifying the topic. Such large depths are impossible in standard decision trees due to memory limitations.

\myparagraph{Information bottleneck and minimal sufficient statistics.} The problem of finding minimal-length, task-sufficient codes is not new. For example, the \emph{information bottleneck} method \cite{tishby2000information} seeks a minimum-length encoding for $X$ that is (approximately) sufficient to solve task $Y$. Our concept of description length differs in that we constrain the code to consist of interpretable query functions rather than {\it all functions} of the input, as in the information bottleneck and classical information theory. Indeed, arbitrary subsets of the input space (e.g. images) are overwhelmingly {\it not} interpretable to humans. 

\myparagraph{Sequential active testing and hard attention.} The \textit{information pursuit} (IP) algorithm we use was introduced in \cite{geman1996active} under the name "active testing," which sequentially observes parts of an input (rather than the whole input at once), using mutual information to determine "where to look next," which is calculated online using on a scene model.  Sequentially guiding the selection of partial observations has also been independently explored in Bayesian experimental design \cite{chaloner1995bayesian}. Subsequent works in these two areas include many ingredients of our approach (e.g. generative models \cite{sznitman2010active,jahangiri2017information} and MCMC algorithms \cite{cuturi2020noisy}). Of particular interest is the work of Branson \emph{et al.} \cite{branson2014ignorant} which used the CUB dataset to identify bird species by sequentially asking pose and attribute queries to a human user. They employ IP to generate the query sequence based on answers provided by the user, much like our experiments in \S\ref{Sec: Concept-Based Queries Birds}. However, for the sake of tractability, all the above works assume that query answers are independent conditioned on $Y$. We do not. Rather, to the best of our knowledge, ours is the first implementation of the IP algorithm that uses deep generative models and only assumes that queries are independent given $Y$ \textit{and some latent variable $Z$}. This greatly improves performance, as we show in \S\ref{section: Experiments}.

The strategy of inference through sequential observations of the input has been recently re-branded in the deep learning community as \textit{Hard Attention} \cite{mnih2014recurrent, elsayed2019saccader, li2016glance}. However, high variance in gradient estimates and scalability issues have prevented widespread adoption. In the future, we wish to explore how our work could inform more principled and better-performing reward functions for Hard Attention models.

\myparagraph{Visual question answering.} Although it may appear that our work is also related to the Visual Question Answering (VQA) literature \cite{li2019visual,malinowski2017ask,mao2019neuro,shih2016look,lu2016hierarchical,andreas2016neural}, we note that our work addresses a very different problem. VQA focuses on training deep networks for \textit{answering} a large set of questions about a visual scene. In contrast, our framework is concerned with \textit{selecting} a small number of queries to ask about a given image to solve a task, say classification. As we move on to more complex tasks, an interesting avenue for future work would involve using VQA systems to supply answers to the queries used in our framework. However, this would  require significantly more complex generative models that the ones considered here.

\section{Learning Interpretable Predictors by Composing Queries via Information Pursuit}

\label{section: Quantifying e-entropy}
Let $X$ and $Y$ be the input data and the corresponding output/hypothesis, both random variables assuming values in $\gX$ and $\gY$ respectively. In supervised learning, we seek to infer $Y$ from $X$ using a finite set of samples drawn from the joint distribution $p_{XY}(x, y)$.\footnote{We denote random variables by capital letters and their realizations with small letters.} As motivated in Section~\ref{sec:introduction},  useful explanations for prediction should be \textit{task-dependent}, \textit{compositional}, \textit{concise} and \textit{sufficient}.  We capture such properties through a suitably rich set $Q$ of binary functions $q(x)$, or \textit{queries}, whose answers $\{q(x)\}_{q\in Q}$  collectively determine the task $Y$.  More precisely, a query set $Q$ is \textit{sufficient} for $Y$ if
\begin{equation}
p (y \mid x) = p(y \mid \{x' \in \gX : q(x') = q(x) \ \forall q\in Q\}).
\end{equation}
In other words, $Q$ is sufficient for $Y$ if whenever two inputs $x$ and $x'$ have identical answers for all queries in $Q$, their corresponding posteriors are equal, i.e., $p(y \mid x) = p (y \mid x')$. 

Given a fixed query set $Q$, how do we compose queries into meaningful representations that are predictive of $Y$? We answer this by first formally defining an explanation strategy $\pi$ and then formulating the task of composing queries as an optimization problem. 

\myparagraph{Explanation strategies based on composing queries.} 
An {\it explanation strategy}, or just {\it strategy}, is a function, $\pi : K^* \rightarrow Q$, where $K^*$ is the set of all finite-length sequences generated using elements from the set $K = \{(q, q(x)) \mid q \in Q, x \in \gX\}$ of query-answer pairs. We require that $Q$ contains a special query, $q_{STOP}$, which signals the strategy to stop asking queries and output $expl_Q^{\pi}(x)$, the set of query-answer pairs asked before $q_{STOP}$. More formally, a strategy $\pi$ is recursively defined as follows; given input sample $x^{\textrm{obs}}$
\begin{enumerate}
    \item $q_1 = \pi(\emptyset)$. The first query is independent of $x^{\textrm{obs}}$.
    \item $q_{k+1} = \pi(\{q_i, {q_i}(x^{\textrm{obs}})\}_{1:k})$. All subsequent queries depend on the query-answer pairs observed so far for~$x^{\textrm{obs}}$.  
    \item If $q_{L + 1}=q_{STOP}$ terminate, and return 
    \begin{equation}
    \label{def: expl}
    expl_Q^{\pi}(x^{\textrm{obs}}) := \{q_i, {q_i}(x^{\textrm{obs}})\}_{1:L}.
    \end{equation}
\end{enumerate} 

Notice that each $q_i$ depends on $x^{\textrm{obs}}$, but we drop this dependency in the notation for brevity. 
We call the number of pre-STOP queries for a particular $x^{\textrm{obs}}$ as the explanations' description length and denote it by $t^{\pi}(x^{\textrm{obs}}) := |expl^{\pi}_Q(x^{\textrm{obs}})|$. Computing a strategy on $x^{\textrm{obs}}$ is thus akin to traversing down the branch of a decision tree dictated by $x^{\textrm{obs}}$. Each internal node encountered along this branch computes the query proposed by the strategy based on the path (query-answer pairs) observed so far.

Notice also that we restrict out attention to {\it sequential} strategies so that the resulting explanations satisfy the property of being \textit{prefix-free}.\footnote{The term prefix-free comes from the literature on instantaneous codes in information theory.} This means that explanations generated for predictions made on an input signal $x_1$ cannot be a sub-part for explanations generated for predictions on a different input signal $x_2$; otherwise, the explanation procedure is ambiguous because a terminal node carrying one label could be an internal node of a continuation leading to a different label. Sequential strategies generate prefix-free explanations by design. For non-sequential strategies, which are just functions mapping an input $X$ to a set of queries in $Q$, it is not clear how to effectively encode the constraint of generating prefix-free explanations.

\myparagraph{Concise and approximately sufficient strategies.} 
In machine learning, we are often interested in solving a task \textit{approximately} rather than \textit{exactly}. Let $Q$ be sufficient for $Y$, choose a distance-like metric $d$ on probability distributions and let $\epsilon > 0$.  We propose the following optimization problem to efficiently compose queries for prediction,
\begin{align}
    \label{eq: complexity definition}
    &\min_{\pi} \E_{X}\left[|expl^{\pi}_Q(X)|\right] =: H^{\epsilon}_Q(X; Y)  \\
    &\textrm{s.t.} \; \mathbb{E}_X[d\left(p(Y \mid X), p(Y \mid expl^{\pi}_Q(X)\right)] \leq \epsilon \;\; (\epsilon\textit{-Sufficiency)}, \nonumber
\end{align}
where the minimum is taken over all strategies $\pi$. The solution $\pi^*$ to \eqref{eq: complexity definition} provides a criterion for an optimal strategy for the task of inferring $Y$ approximately from $X$. The \textit{minimal expected description length} objective, $H^{\epsilon}_Q(X; Y)$, ensures the conciseness of the explanations, while the constraint ensures approximate sufficiency of the explanation. The ``metric" $d$ on distributions could be KL-divergence, total variation, Wasserstein distance, etc. The hyper-parameter $\epsilon$ controls how approximate the explanations are. The posterior $p(y \mid expl^{\pi}_Q(x^{\textrm{obs}}))$ should be interpreted as the conditional probability of $y$ given the event 
\begin{equation}
    [x^{\textrm{obs}}]_{\pi, Q} := \{x \in \gX\ \mid expl^{\pi}_{Q}(x) = expl^{\pi}_Q(x^{\textrm{obs}})\}.
    \label{eq: eq. class}
\end{equation}
$expl^{\pi}_Q(X)$ can also be interpreted as a random variable which maps input $X$ to its equivalence class $[X]_{\pi, Q}$.

\begin{figure}
\centering
\includegraphics[width=0.95\linewidth]{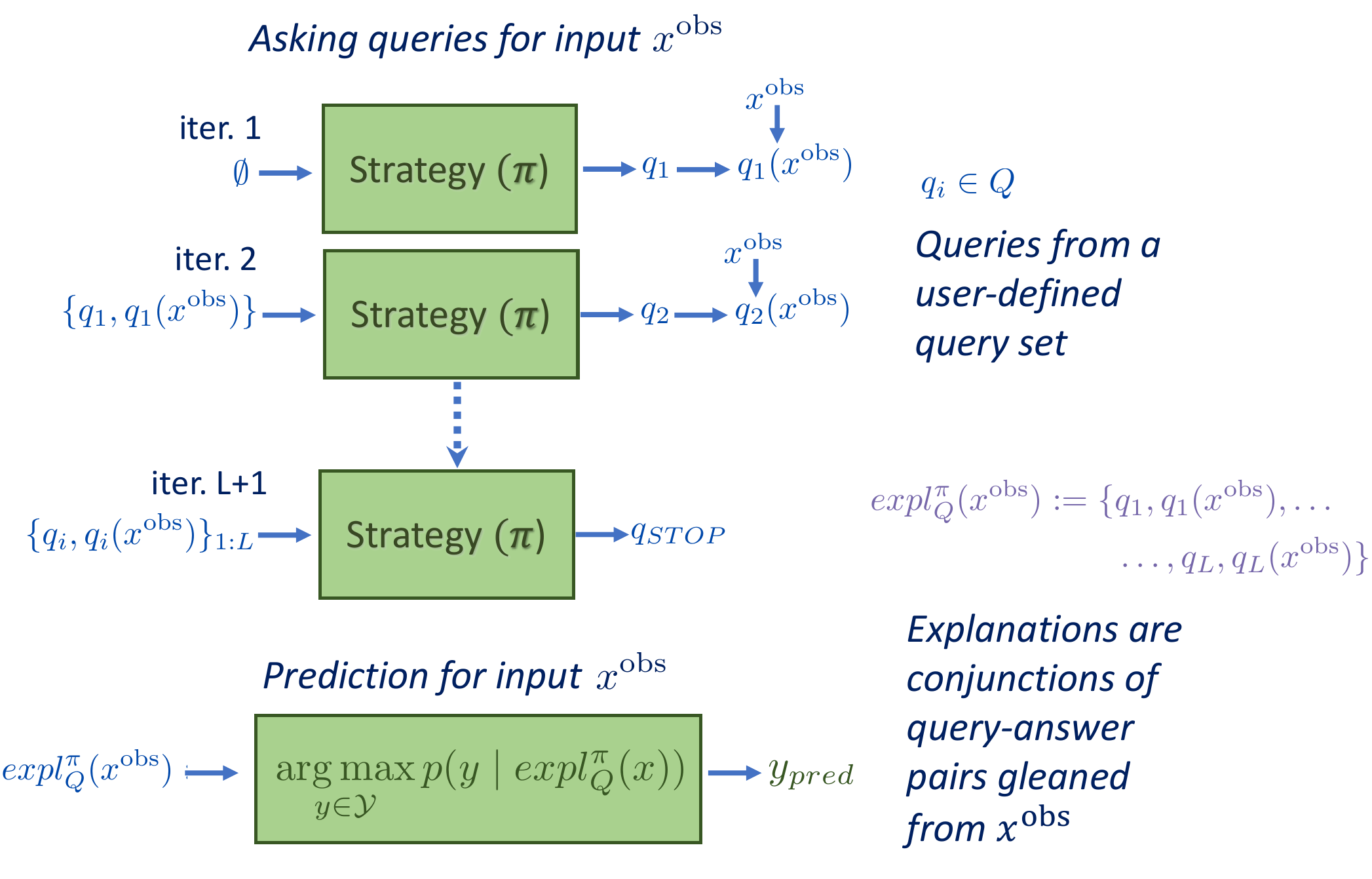}
    \caption{Schematic view of the overall framework for quantifying explanations for predicting $y$ from $x^{\textrm{obs}}$. For details see Sec. \ref{section: Quantifying e-entropy}.}
    \label{Fig: framework}
\end{figure}

The final prediction/inference for the input $x^{\textrm{obs}}$ is then taken to be the usual MAP estimator, namely
\begin{equation}
y_{pred} = \argmax_{y \in \gY}p(y \mid expl^{\pi}_Q(x^{\textrm{obs}})).
\end{equation}
The sequence of query-answers streams obtained by $\pi$ on $x^{\textrm{obs}}$ serves as the explanation for $y_{pred}$. One could also monitor the posterior over the labels $Y$ evolving as successive queries get asked to gain more insight into the strategy's decision-making process. Fig. \ref{Fig: framework} illustrates the overall framework in detail. 


\myparagraph{Information Pursuit: a greedy approximation.} Unfortunately, solving \eqref{eq: complexity definition} is known to be NP-Complete and hence generally intractable \cite{laurent1976constructing}. As an approximate solution to \eqref{eq: complexity definition} we propose to use a greedy algorithm called Information Pursuit (IP). IP was introduced by Geman \& Jedynak in 1996 \cite{geman1996active} as a model-based, online construction
of a single but deep branch. The IP strategy, that is, $\pi = \textrm{IP}$, is recursively defined as follows, 
\begin{align}
    \label{def. IP algorithm}
    q_1 &= \textrm{IP}(\emptyset) = \argmax_{q \in Q} I({q}(X); Y) \\
    q_{k+1} &= \textrm{IP}(\{q_i, {q_i}(x^{\textrm{obs}})\}_{1:k})) = \argmax_{q \in Q} I({q}(X); Y \mid S^{\textrm{IP}}_{k}(x^{\textrm{obs}}))\nonumber
\end{align}
where $I$ denotes mutual information and $S^{\textrm{IP}}_{k}(x^{\textrm{obs}})$ corresponds to the event $\{x \in \gX \mid \{q_i, {q_i}(x^{\textrm{obs}})\}_{1:k} = \{q_i, {q_i}(x)\}_{1:k}\}$. Ties in choosing $q_{k+1}$ are broken arbitrarily if the maximum is not unique. 

The algorithm stops when there are no more informative queries left in $Q$, that is, it satisfies the following condition:
\begin{equation}
\begin{aligned}
q_{L+1} = q_{STOP} \quad \textrm{ if } \; &\max_{q \in Q} I({q}(X); Y \mid S^{\textrm{IP}}_{m}(x^{\textrm{obs}})) \leq \epsilon \\& \; \forall m \in \{L, L + 1, ..., L + T\},
\label{eq: practical terminating condition}
\end{aligned}
\end{equation}
where hyper-parameter $T > 0$ is chosen via cross-validation. This termination criteria corresponds to taking the distance-like metric $d$ in \eqref{eq: complexity definition} as the KL-divergence between the two distributions. Further details about the relation between this termination criteria and the $\epsilon$-Sufficiency constraint in  \eqref{eq: complexity definition} are provided in Appendix \ref{Appendix: termination criteria for IP}. For tasks in which $Y$ is a function of $X$, a common scenario in many supervised learning problems, we use a simpler alternative,
\begin{equation}
\begin{aligned}
q_{L+1} = q_{STOP} \quad \textrm{ if } \; &\argmax_{y \in \gY} p(y \mid S^{\textrm{IP}}_{m}(x^{\textrm{obs}})) \geq 1 - \epsilon \\& \; \forall m \in \{L, L + 1, ..., L + T\}.
\label{eq: practical terminating condition for deterministic cases}
\end{aligned}
\end{equation}

The key distinction between the information gain criteria used in standard decision tree induction and $\textrm{IP}$ is that the former uses the empirical distributions to compute \eqref{def. IP algorithm} while the latter is based on generative models (as we will see in Section \ref{section: Information Pursuit}). The use of generative models guards against data fragmentation \cite{vilalta1997global} and thus allows for asking longer sequences of queries without grossly over-fitting.

\myparagraph{How does $\textrm{IP}$ compare to the optimal strategy $\boldsymbol{\pi^*}$?} We begin by characterizing the constraint in \eqref{eq: complexity definition} in terms of mutual information, the quantity that drives $\textrm{IP}$.

\begin{proposition}
Let $S^{\pi}_{k}(X)$ be a random variable where any realization $S^{\pi}_{k}(x^{\textrm{obs}})$, $x^{\textrm{obs}} \in \gX$, denotes the event 
\[ S^{\pi}_k(x^{\textrm{obs}}) := \{x' \in \gX \mid \{q_i, q_i(x^{\textrm{obs}})\}_{1:k} = \{q_i, q_i(x')\}_{1:k} \},\]
where $q_i$ is the $i^{th}$ query selected by $\pi$ for input $x^{\textrm{obs}}$.
Here we use the convention that $S^{\pi}_{0}(X) = \Omega$ (the entire sample space) and $S^{\pi}_{l}(X) = S^{\pi}_{t^{\pi}(X)}(X)$ $\; \forall l > t^{\pi}(X)$. If $Q$ is finite\footnote{The assumption of $Q$ being a finite set is benign. Many interested applications can be addressed with a finite $Q$ as we show in our experiments.} and $d$ is taken to be the KL-divergence, then objective \eqref{eq: complexity definition} can be rewritten as
\begin{equation}
\label{eq: complexity definition MI}
\begin{split}
     &H_Q^{\epsilon}(X; Y) := \min_{\pi} \E_{X}\left[|expl^{\pi}_Q(X)|\right] \\
    &\textrm{s.t.} \;  \sum_{k=1}^{\tau^{\pi}}I(Y; S^{\pi}_k(X) \mid S^{\pi}_{k - 1}(X)) \geq I(X; Y) - \epsilon,
    \end{split}
\end{equation}
where $\tau^{\pi} = \max\{t^{\pi}(x): x \in \gX\}$ and $t^{\pi}(X)$ is defined as the number of queries selected by $\pi$ for input $X$ until $q_{STOP}$. 
\end{proposition}

See Appendix \ref{Appendix: Characterizing the optimal strategy} for a detailed proof. The objective in \eqref{eq: complexity definition MI} can be alternatively stated as, 
\begin{equation}
    \label{eq: complexity definition MI 2}
\begin{split}
    &\max_{\pi} \sum_{k=1}^{\tau^{\pi}}I(Y; S^{\pi}_k(X) \mid S^{\pi}_{k - 1}(X)) \\
    & \textrm{s.t.} \; \E_{X}\left[|expl^{\pi}_Q(X)|\right] \leq \gamma, 
\end{split}
\end{equation}
where $\gamma > 0$ is a user-defined hyper-parameter. From \eqref{eq: complexity definition MI 2} it is clear that the optimal strategy $\pi^*$ would ask a sequence of queries about $X$ that would maximize the cumulative sum of the 
mutual information each additional query provides about $Y$, conditioned on the history of query-answers observed so far, subject to a constraint on the average number of queries that can be asked. As stated before, solving for $\pi^*$ is infeasible but a greedy approximation that makes locally optimal choices is much more amenable.

Suppose that one has been given the answers to $k$ queries about a given input, the locally optimal choice would then be to ask the most informative query about $Y$ conditioned on the history of these $k$ query-answers observed. This greedy choice at each stage gives rise to the $\textrm{IP}$ strategy. Obtaining approximation guarantees for $\textrm{IP}$ is still an open problem; however in the special case where $Q$ is taken to be the set of all possible binary functions of $X$, it is possible to show that $\textrm{IP}$ \textit{asks at most $1$ query more} than $\pi^*$ on average. More formally, we have the following result, whose proof can be found in Appendix \ref{Appendix: proof of performance guarantee IP}.
\begin{proposition}\label{prop: IP guarantees} Let $Y$ be discrete. Let $\tilde{H}_Q(X; Y)$ be the expected description length obtained by the $\textrm{IP}$ strategy. If $H(Y | X) = 0$ and $Q$ is the set of all possible binary functions of $X$ such that $H(q(X) \mid Y) = 0$ $\forall q \in Q$, then 
\begin{equation}
    H(Y) \leq \tilde{H}_Q(X; Y) \leq H(Y) + 1
\end{equation}
\end{proposition} 

Having posed the problem of finding explanations as an optimization problem and proposed a greedy approximation to solving it, in the next section we propose a tractable implementation of IP based on deep generative models. 




\section{Information Pursuit using Variational Autoencoders and Unadjusted Langevin}
\label{section: Information Pursuit}

IP requires probabilistic models relating query-answers and data to compute the required mutual information terms in \eqref{def. IP algorithm}. Specifically, computing $q_{k + 1}$ in \eqref{def. IP algorithm} (for any iteration number $k$) requires computing the mutual information between $q(X)$ and $Y$ given the history $S^{\textrm{IP}}_{k}(x^{\textrm{obs}})$ till time $k$. As histories become longer, we quickly run out of samples in our dataset which belong to the event $S^{\textrm{IP}}_{k}(x^{\textrm{obs}})$. As a result, non-parametric sample-based methods to estimate mutual information (such as \cite{belghazi2018mine}) would be impractical. In this section, we propose a model-based approach to address this challenge for a general supervised learning task and query set $Q$. In \S\ref{section: Experiments} we adapt this model to the specific cases where $Q$ is taken to be image patches or task-based concepts.


\myparagraph{Information Pursuit Generative Model.} To make learning tractable, we introduce latent variables $Z$ to account for all the dependencies between different query-answers, and we posit the following factorization of $Q(X), Y, Z$
\begin{align}
    \label{eq: factorization of query answers}
p_{Q(X)ZY} &(Q(x), z, y) \\
&= \prod_{q \in Q}p_{q(X)\mid ZY}(q(x) \mid z, y)p_Y(y)p_Z(z),\nonumber
\end{align}
where $Q(X) = \{q(X): q \in Q\}$, and $z$ and $q(x)$ denote realizations of $Z$ and $q(X)$ respectively. In other words, we assume that the query-answers are conditionally independent given the label $y$ and a latent vector~$z$. The independence assumption in \eqref{eq: factorization of query answers} shows up ubiquitously in many machine learning applications, such as the following.
\begin{enumerate}
    \item \textbf{$\boldsymbol{q(X)}$ as object presence indicators evaluated at non-overlapping windows:} Let $Q$ be a set of non-overlapping windows in the image $X$ with $q(X)$ being a random variable indicating the presence of an object at the $q^{th}$ location. The correlation between the $q$s is entirely due to latent image generating factors $Z$, such as lighting, camera position, scene layout, and texture along with the scene description signal $Y$.  
    \item \textbf{$\boldsymbol{q(X)}$ as snippets of speech utterances:} A common assumption in speech recognition tasks is that the audio frame features ($q(X)$) are conditionally independent given latent phonemes $Z$ (which is often modeled as a Hidden Markov Model). 
\end{enumerate}

The latent space $Z$ is often a lower-dimensional space compared to the original high-dimensional $X$. We learn $Z$ from data in an unsupervised manner using variational inference. Specifically, we parameterize the distributions $\{p_{\omega}(q(x) \mid z, y) \ \forall q \in Q\}$ with a \textbf{\textit{Decoder Network}} with shared weights $\omega$. These weights are learned using stochastic Variational Bayes \cite{kingma2013auto} by introducing an approximate posterior distribution $p'_\phi(z \mid y, Q(x))$ parameterized by another neural network with weights $\phi$ called the \textbf{\textit{Encoder Network}} and priors $p_Y(y)$ and $p_Z(z)$. More specifically, the parameters $\phi$ and $\omega$ are learned by maximizing the Evidence Lower BOund (ELBO) objective. Appendix \ref{Appendix: Network architectures and training} gives more details on this optimization procedure. The learned Decoder Network $p_{\omega^*}(q(x) \mid z, y)$ is then used as a plug-in estimate for the true distribution $p_{q(X)\mid ZY}(q(x) \mid z, y)$, which is in turn used to estimate \eqref{eq: factorization of query answers}.

\myparagraph{Implementing $\boldsymbol{\textrm{IP}}$ using the generative model.}
Once the Decoder Network has been learned using variational inference, the first query $q_1 = \textrm{IP}(\emptyset)$ is the one that maximizes the mutual information with $Y$ as per \eqref{def. IP algorithm}.
The mutual information term for any query $q$ is
completely determined by $p({q}(x), y)$, which is obtained by
numerically marginalizing the nuisances $Z$ from
\eqref{eq: factorization of query answers} using Monte Carlo integration. In particular, we carry out the following computation $\forall q \in Q$,
\begin{equation}
\begin{aligned}
    p_{q(X)Y}({q}(x),y)  &= \int_z p_{Q(X)ZY}(Q(x), z, y)dz \\
                     &= \int_z p_{q(X)\mid ZY}(q(x) \mid z, y)p_Y(y)p_Z(z)dz\\
                     &\approx \frac{1}{N} \sum_{i=1}^N p_{\omega^*}({q}(x)\mid y,z^{(i)})p_Y(y) \\
                     & =: \tilde{p}(q(x), y)).
\end{aligned}
\end{equation}
In the last approximation, $p_{\omega^*}({q}(x)\mid y,z^{(i)})$ is the distribution obtained using the trained decoder network. $N$ is the number of i.i.d. samples drawn
and $z^{i} \sim p_Z(z)$. We then estimate mutual information numerically via the following formula,
\begin{equation}
\label{eq: numerical MI first query}
I(Y; q(X)) = \!\!\sum_{q(x), y}\!\!\tilde{p}(q(x), y) \log \frac{\tilde{p}(q(x), y )}{\tilde{p}(q(x))\tilde{p}(y\!)}.
\end{equation} 

The computation of subsequent queries $q_{k+1}$ requires the mutual information conditioned on observed history $S_{k}^{\textrm{IP}}(x^{\textrm{obs}})$, which can be calculated from the distribution 

\begin{align}
\label{eq: compute intermediate conditionals}
&p({q}(x), y \mid S^{\textrm{IP}}_k(x^{\textrm{obs}})) \\
&= \int   p({q}(x), z, y \mid S^{\textrm{IP}}_k(x^{\textrm{obs}}))dz \nonumber \\
&= \int   p({q}(x) \mid z, y, S^{\textrm{IP}}_k(x^{\textrm{obs}}))p(z \mid y, S^{\textrm{IP}}_k(x^{\textrm{obs}}))p(y \mid S^{\textrm{IP}}_k(x^{\textrm{obs}}))dz \nonumber\\
&= \int   p({q}(x) \mid z, y)p(z \mid y, S^{\textrm{IP}}_k(x^{\textrm{obs}}))p(y \mid S^{\textrm{IP}}_k(x^{\textrm{obs}}))dz. \nonumber
\end{align}
The first equality is an application of the law of total probability. The last equality appeals to the assumption that $\{q(X), q \in Q\}$ are conditionally independent given $Y, Z$ \eqref{eq: factorization of query answers}.

To estimate the right-hand side of \eqref{eq: compute intermediate conditionals} via Monte Carlo
integration, one needs to sample $z^{i} \sim p\left(z \mid y, S^{\textrm{IP}}_{k}(x^{\textrm{obs}})\right)$
and compute 
\begin{align}
\label{eq: marginalize z}
    p(q(x), y \mid S^{\textrm{IP}}_{k}(x^{\textrm{obs}})) 
    &\approx \tilde{p}(q(x), y \mid S^{\textrm{IP}}_{k}(x^{\textrm{obs}}))\\
    &:= \frac{1}{N} \sum_{i=1}^N p_{\omega^*}({q}(x)\!\mid\! z^{(i)}, y)p(y \!\mid\! S^{\textrm{IP}}_{k}(x^{\textrm{obs}})),\nonumber
\end{align}
where the term $p(y \mid S^{\textrm{IP}}_{k}(x^{\textrm{obs}}))$ is estimated recursively via the Bayes' theorem. This computation is as follows, 
\begin{equation}
\label{eq: posterior y}
\begin{aligned}
    p(y \mid S^{\textrm{IP}}_k(x)) &\propto p(y, S^{\textrm{IP}}_k(x)) \\ &= p({q_k}(x), y, S^{\textrm{IP}}_{k-1}(x)) \\ &\propto p({q_k}(x) \mid y, S^{\textrm{IP}}_{k-1}(x))p(y \mid S^{\textrm{IP}}_{k-1}(x))
    \end{aligned}
\end{equation}
$S^{\textrm{IP}}_0(x) = \emptyset$ (since no evidence via queries has been gathered from $x$ yet) and so $p(y \mid S^{\textrm{IP}}_0(x)) = p_Y(y)$. The posterior $p(y \mid S^{\textrm{IP}}_k(x))$ is obtained by normalizing the last equation in \eqref{eq: posterior y} such that $\sum_y p(y|S^{\textrm{IP}}_k(x)) = 1$. This recursive updating of the posterior is similar to the posterior updates used in Bayesian sequential filtering \cite{doucet2009tutorial}. The term
$p({q_k}(x) \mid y, S^{\textrm{IP}}_{k-1}(x))$ is estimated using \eqref{eq: marginalize z}.

Having estimated $p({q}(x), y \mid S^{\textrm{IP}}_{k}(x^{\textrm{obs}}))$, we then numerically compute the mutual information between query-answer $q(X)$ and $Y$ given history for every $q \in Q$ via the formula
\begin{align}
\label{eq: numerical MI}
&I(Y; q(X) \mid S^{\textrm{IP}}_{k}(x^{\textrm{obs}})) = \\
&\!\!\sum_{q(x), y}\!\!\tilde{p}(q(x), y \!\mid\! S^{\textrm{IP}}_{k}(x^{\textrm{obs}})) \log \frac{\tilde{p}(q(x), y \mid S^{\textrm{IP}}_{k}(x^{\textrm{obs}}))}{\tilde{p}(q(x) \!\mid\! S^{\textrm{IP}}_{k}(x^{\textrm{obs}}))\tilde{p}(y\! \mid\! S^{\textrm{IP}}_{k}(x^{\textrm{obs}}))}.\nonumber
\end{align}


\myparagraph{Estimating $\boldsymbol{p\left(z \mid y, S^{\textrm{IP}}_{k}(x^{\textrm{obs}})\right)}$ with the Unadjusted Langevin Algorithm.} 
Next we describe how to sample from this posterior $p(z \mid y, S^{\textrm{IP}}_{k}(x^{\textrm{obs}}))$ using the Unadjusted Langevin Algorithm (ULA). ULA is an iterative algorithm used to approximately sample from any distribution with a density known only up to a normalizing factor. It has been successfully applied to many high-dimensional Bayesian inference problems \cite{jalal2021instance,nijkamp2020anatomy, durmus2019high}. Given an initialization $z^{(0)}$, ULA proceeds by
\begin{equation}
  z^{(i + 1)} = z^{(i)} +  \eta \nabla U(z^{(i)}) + \sqrt{2\eta}\zeta^{(i + 1)}. 
  \label{eq: ULA update}
\end{equation}
Here $(\zeta^{(i)})_{i \geq 1} \sim \gN(0, I)$ and $\eta$ is the step-size. Asymptotically, the chain $(z^{(i)})_{i \geq 1}$ converges to a stationary distribution that is ``approximately" equal to a measure with density $\propto e^{U(z)}$ \cite{welling2011bayesian}.

For IP, we need samples from $p(z \mid y, S^{\textrm{IP}}_{k}(x^{\textrm{obs}})$. This is achieved by initializing $z^{(0)}$ using the last iterate of the ULA chain used to simulate $p(z \mid y, S^{\textrm{IP}}_{k - 1}(x^{\textrm{obs}})$.\footnote{$z^{(0)} \sim \gN(0, I)$ for the first iteration of IP.} We then run ULA for $N$ iterations by recursively applying \eqref{eq: ULA update} with 
\[U(z) := \log p(z, S^{\textrm{IP}}_{k}(x^{\textrm{obs}}) \!\mid\! y) = \log p(S^{\textrm{IP}}_{k}(x^{\textrm{obs}})\! \mid\! z, y)p(z)p(y).\]

The number of steps $N$ is chosen to be sufficiently large to ensure the ULA chain converges ``approximately" to the desired $z \sim p(z \mid y, S^{\textrm{IP}}_{k}(x^{\textrm{obs}}))$. We use the trained decoder network $\prod_{i=1}^{k} p_{\omega}(q_i(x) \mid z, y)$, with $q_i$ being the $i^{th}$ query asked by $\textrm{IP}$ for input $x$, as a proxy for $p\left(S^{\textrm{IP}}_{k}(x^{\textrm{obs}})\right) \mid z, y)$. We then obtain stochastic approximations of \eqref{eq: compute intermediate conditionals} by time averaging the iterates,
\begin{equation}
p\left({q}(x), y \! \mid \! S^{\textrm{IP}}_{k}(x^{\textrm{obs}})\right)\!\! \approx \!\! \frac{1}{N} \sum_{i=1}^N p_{\omega}\!\left({q}(x)\!\mid\! z^{(i)}, y\right)\!p\!\left(y\! \mid\! S^{\textrm{IP}}_{k}(x^{\textrm{obs}})\!\right)\!\!,
\end{equation}
where $(z^{(i)})_{1:N}$ are the iterates obtained using the ULA chain whose stationary distribution is ``approximately" $p\left(z \mid y, S^{\textrm{IP}}_{k}(x^{\textrm{obs}})\right)$. 

\myparagraph{Algorithmic complexity for IP.}
For any given input $x$, the per-iteration cost of the IP algorithm is $\mathcal{O}(N + |Q|m)$\footnote{In this computation we have assumed, for simplicity, a unit cost for any operation that was computed in a batch concurrently on a GPU.}, where $|Q|$ is the total number of queries, $N$ is the number of ULA iterations, and $m$ is cardinality of the product sample space $q(X) \times Y$. For simplicity we assume that the output hypothesis $Y$ and query-answers $q(X)$ are finite-valued and also that the number of values query answers can take is the same. However, our framework can handle more general cases. See Appendix \ref{Appendix: complexity analysis} for more details.

\section{Experiments}
\label{section: Experiments}
In this section, we empirically evaluate the effectiveness of our method. We begin by analyzing the explanations provided by IP for classifying individual input data, in terms of words, symbols, or patterns (the queries). We find in each case that IP discovers concise explanations which are amenable to human interpretation. 
We then perform quantitative comparisons which show that (i) IP explanations are more faithful to the underlying model than existing attribution methods; and (ii) the predictive accuracy of our method using a given query set is competitive with black-box models trained on features provided by the same set. 

\begin{figure*}
\centering
\includegraphics[width=1\textwidth]{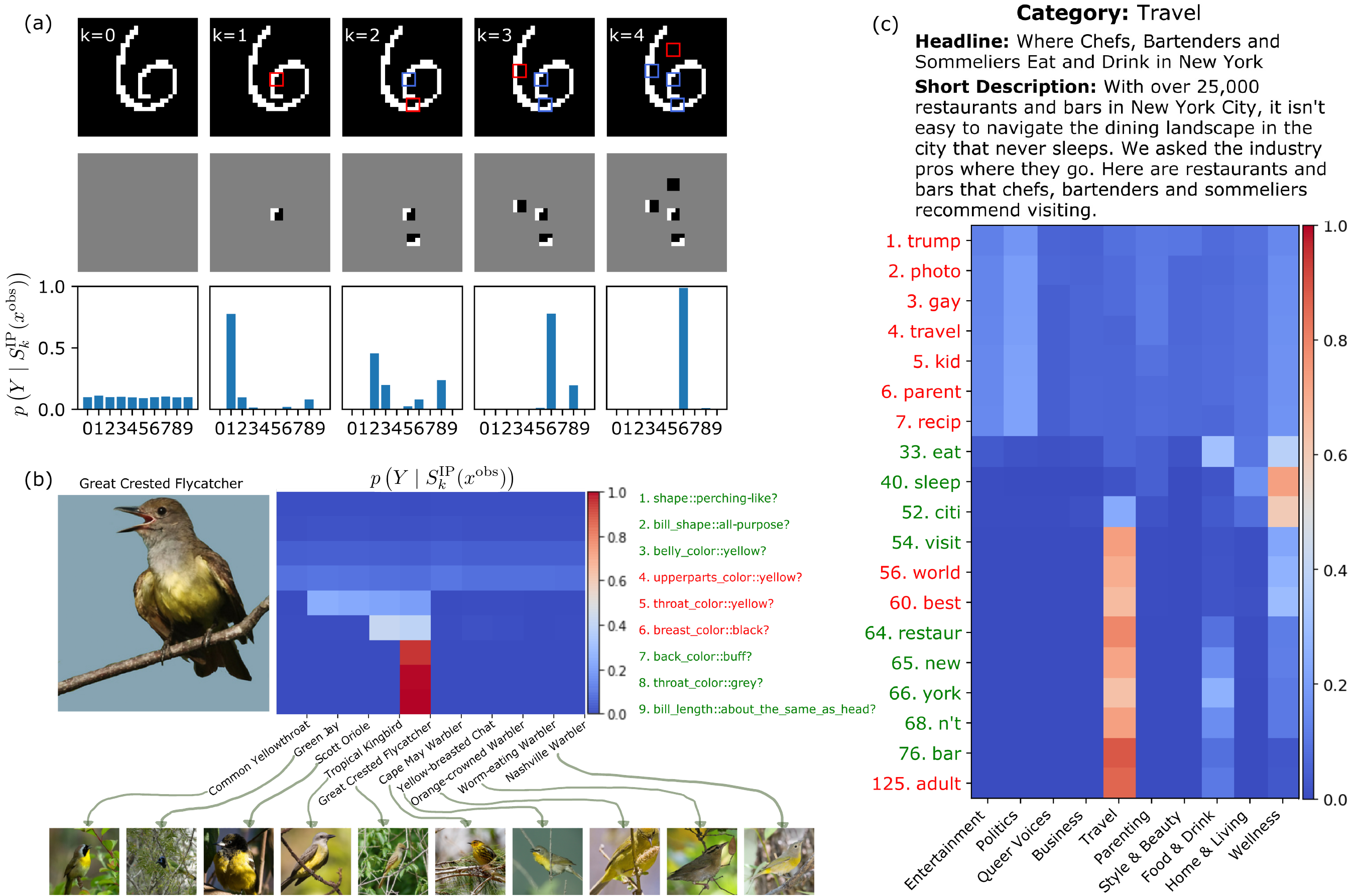}
\caption{\textbf{(a) IP on MNIST.} The top row displays the test image with red boxes denoting the current queried patch and blue boxes denoting previous patches. The second row shows the revealed portion of the image that IP gets to use at each query. The final row shows the model's estimated posteriors at each query, beginning at a nearly uniform prior before converging on the true digit ``6'' after $4$ queries. \textbf{(b) IP on CUB Bird Species Classification.} On the left we show the input image and on the right we have a heatmap of the estimated class probabilities at each iteration. We only show the top $10$ most probable classes out of the $200$. To the right, we display the queries asked at each iteration, with red indicating a ``no'' response and green a ``yes'' response. \textbf{(c) IP on HuffPost News.} We show the input news item and a heatmap depicting the evolution of topic probabilities as IP asks queries and gathers answers. Words colored in red are absent from the sentence while words in green are present. For our visualization, we compute the KL divergence between each successive posterior and plot only the top 20 queries that led to the greatest change in posterior class probabilities.}
\label{Fig: IP in action}
\end{figure*}

\subsubsection{Binary Image Classification with Patch Queries}
\myparagraph{Task and query set.} We start with the simple task of binary image classification. We consider three popular datasets -- MNIST \cite{lecun1998gradient}, Fashion-MNIST\cite{xiao2017fashion} and KMNIST \cite{clanuwat2018deep}. We choose a threshold for binarizing these datasets since they are originally grayscale. 
We choose the query set $Q$ as the set of all $w \times w$ overlapping patch locations in the image. The answer $q(X)$ for any $q \in Q$ is the $w^2$ pixel intensities observed at the patch indexed by location $q$. This choice of $Q$ reflects the user's desire to know which parts of the input image are most informative for a particular prediction, a common practice for explainability in vision tasks \cite{chen2019looks}. We conduct experiments for multiple values of $w$ and conclude that $w=3$ provides a good trade-off between the required number of queries and the interpretability of each query. Note that when $w>1$ the
factorization in \eqref{eq: factorization of query answers} that we use to model $p(Q(x), y, z)$ and compute mutual information no longer holds as the overlapping queries $q(X)$ are now \textit{causally related} (and therefore dependent even when conditioned on $Z$, making them unable to be modeled by a VAE). So instead of training a VAE to directly model the query set $p(Q(x) \mid y, z)$, we train a VAE to model the pixel distribution $p(x \mid y, z)$, and then compute the probability distribution over the patch query $p(q(x) \mid z, y)$ as the product of the probabilities of all pixels in that patch.\footnote{Since the patches overlap in our query set, when computing the conditional probability of a patch query given history we only consider the probability of the pixels in the patch that have not yet been observed in our history.} 

\myparagraph{IP in action.} 
Fig.~\ref{Fig: IP in action}(a) illustrates the decision-making process of IP using $3\times 3$ patch queries on an image $x^{\textrm{obs}}$ of a $6$ from the MNIST test set. The first query is near the center of the image; recall from \eqref{def. IP algorithm} that this choice is independent of the particular input image and represents the patch whose pixel intensities have maximum mutual information with $Y$ (the class label). The updated posterior, $p\left(Y \mid S^{\textrm{IP}}_1(x^{\textrm{obs}})\right)$, concentrates most of its mass on the digit ``1'', perhaps because most of the other digits do not commonly have a vertical piece of stroke at the center patch. However, the next query (about three pixels below the center patch) reveals a horizontal stroke and the posterior mass over the labels immediately shifts to $\{2, 3, 6, 8\}$. The next two queries are well-suited to discerning between these four possibilities and we see that after asking $4$ questions, IP is more than $90\%$ confident that the image is a $6$. Such rich explanations in terms of querying informative patches based on what is observed so far and seeing how the belief $p\left(Y \mid S^{\textrm{IP}}_k(x^{\textrm{obs}})\right)$ of the model evolves over time is missing from post-hoc attribution methods which output static importance scores for every pixel towards the black-box model's final prediction.

\myparagraph{Explanation length vs. task complexity.} Fig. \ref{fig:accuracy_minimality_plot} shows that IP requires an average of $5.2$, $12.9$ and $14.5$ queries of size $3\times 3$ to predict the label with $99\%$ confidence ($\epsilon = 0.01$ in \eqref{eq: practical terminating condition for deterministic cases}) on MNIST, KMNIST and FashionMNIST, respectively. This reflects the intuition that more complex tasks require longer explanations. For reference, state-of-the-art deep networks on these datasets obtain test accuracies in order MNIST $\geq$ KMNIST $\geq$ FashionMNIST (see last row in Table \ref{fig: results table}).  

\begin{figure}
    \centering
    \includegraphics[width=0.48\textwidth]{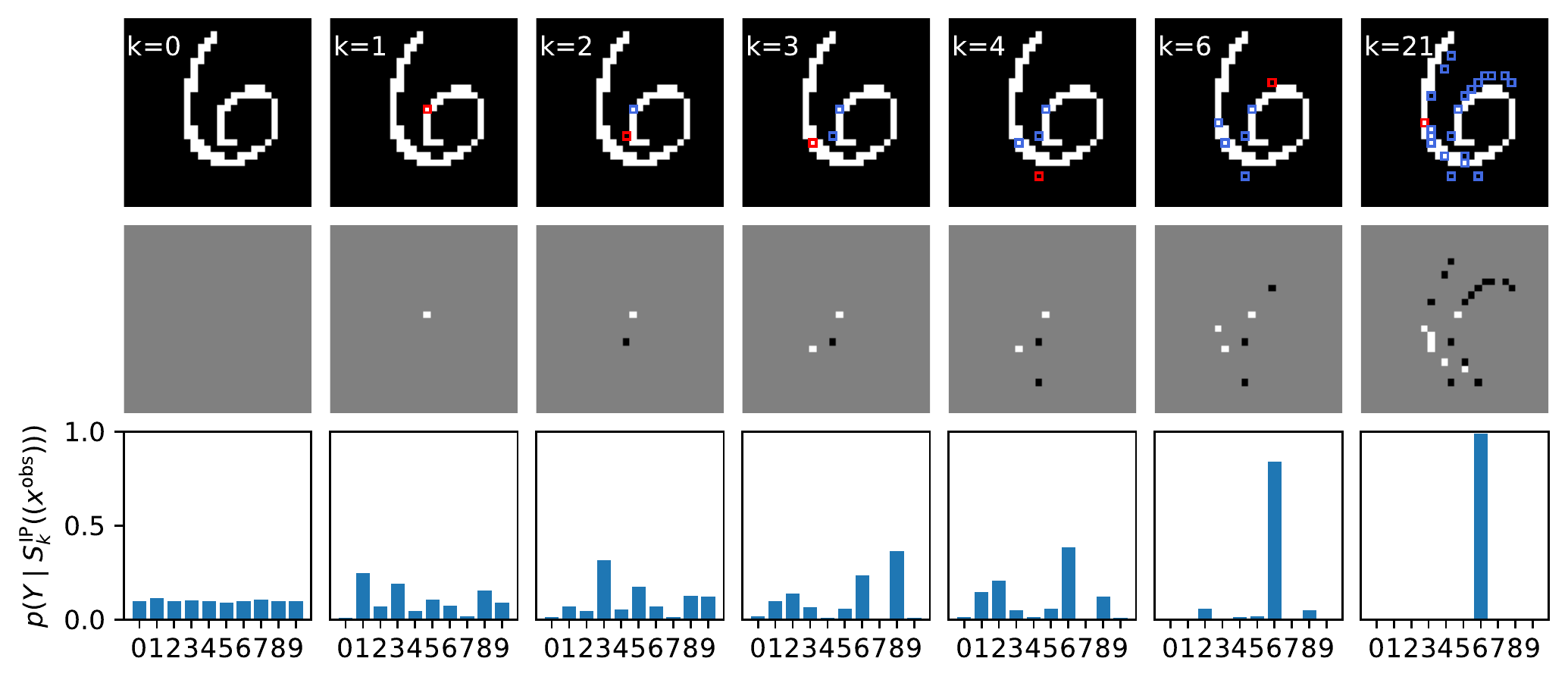}
    \caption{\textbf{IP with $1 \times 1$ patches on MNIST.} Through the first 6 iterations, IP asks queries in the same center vertical region as in Fig. \ref{Fig: IP in action}(a) (which uses $3\times3$ queries), outlining the distinctive loop in the bottom of the ``6''. However, reaching 99\% confidence requires a total of 21 $1\times1$ queries as opposed to just 4 $3\times3$ ones. For conciseness, we show only the 6 queries that led to the greatest KL divergence between successive posterior class probabilities.}
    \label{fig:IP_in_action_MNIST_1x1}
\end{figure}

\myparagraph{Effect of patch size on interpretability.} 
We also run IP on MNIST with patch sizes of $1\times1$ (single pixels), $2\times2$, $3\times3$, and $4\times4$. We observed that IP terminates at 99\% confidence after $21.1$, $9.6$, $5.2$, and $4.6$ queries on average, respectively. While this suggests that larger patches lead to shorter explanations, we note that explanations with larger patches use more pixels (e.g. on MNIST, IP uses $21.1$ pixels on average for $1\times 1$ patches and $54.7$ pixels on average for $4\times4$ patches). That being said, very small patch queries are hard to interpret (see Fig. \ref{fig:IP_in_action_MNIST_1x1}) and very large patch queries are also hard to interpret since each patch contains many image features. Overall, we found that $3 \times 3$ patches represented the right trade-off between interpretability in terms of edge patterns and minimality of the explanations. Specifically, single pixels are not very interpretable to humans but the explanations generated are more efficient in terms of \textit{number of pixels needed to predict the label}. On the other extreme, using the entire image as a query is not interesting from an interpretability point of view since it does not help us understand which parts of the image are salient for prediction. We refer the reader to Appendix \ref{Appendix: IP Patch Scales} for additional patch size examples and quantitative analysis.


\subsubsection{Concept-Based Queries}
\label{Sec: Concept-Based Queries Birds}
\myparagraph{Task and query set.} What if the end-user is interested in a more semantic explanation of the decision process in terms of high-level concepts? This can be easily incorporated into our framework by choosing an appropriate query set $Q$. As an example we consider the challenging task of bird species classification on the Caltech-UCSD Birds-200-2011 (CUB) dataset \cite{wah2011caltech}. The dataset contains images of 200 different species of birds. Each image is annotated with $312$ binary attributes representing high-level concepts, such as the colour and shape of the beak, wings, and head. Unfortunately, these attribute annotations are very noisy. We follow \cite{koh2020concept} in deciding attribute labels by majority voting. For example, if more than 50\% of images in a class have black wings, then we set all images in that class to have black wings. We construct $Q$ by choosing a query for asking the presence/absence of each of these $312$ binary attributes. Unfortunately, attribute annotations are not available at test time. To remedy this, we train a CNN (see \cite{koh2020concept} for details) to answer each query using the training set annotations, which is then used to answer queries at test time. Subsequently, we learn a VAE to model the joint distribution of query-answers supplied by this CNN (instead of the ground truth annotations) and $Y$, so our generative model can account for any estimation errors incurred by the CNN. Finally, we carry out IP as explained in \S\ref{section: Information Pursuit}.

\myparagraph{IP in action.} Consider the image of a \textit{Great Crested Flycatcher} in Fig.~\ref{Fig: IP in action}(b). IP proceeds by asking most informative queries about various bird attributes progressively making the posterior over the species labels more and more peaked. After $5$ queries, IP has gathered that the image is of a bird that has a perching-like shape, all-purpose beak and yellow belly, but does not have a yellow throat nor yellow upperparts. This results in a posterior concentrated on just $4$ species that exhibit these characteristics. IP then proceeds to discount \textit{Green Jay} and \textit{Scott Oriole} which have black breasts with query $6$. Likewise, \textit{Tropical Kingbirds} have grayish back and is segregated from \textit{Great Crested Flycatchers} which have buff-coloured backs with query $7$. Finally after $9$ queries, IP is $99\%$ confident about the current class. Such concept-based explanations are more accessible to non-experts, especially on fine-grained classification datasets, which typically require domain expertise. On average IP takes $14.7$ queries to classify a given bird image with $\epsilon = 0.007$ as the stopping criteria (See \eqref{eq: practical terminating condition}). 

\subsubsection{Word-based Queries}
\label{Section: Word-based Queries}
\myparagraph{Task and query set.} Our framework can also be successfully applied to other domains like NLP. As an example we consider the task of topic identification from newspaper 
extended headlines (headline + short description field) using the the Huffington Post News Category Dataset \cite{huffpostdataset}. We adopt a simple query set that consists of binary queries probing the existence of words in the extended headline. The words are chosen from a pre-defined vocabulary obtained by stemming all words in the HuffPost dataset and choosing the top-1,000 according to their tf-idf scores \cite{lavin2019analyzing}. We process the dataset to merge redundant categories (such as \textit{Style \& Beauty} and \textit{Beauty \& Style}), remove semantically ambiguous, HuffPost-specific categories (e.g. \textit{Impact} or \textit{Fifty}) and remove categories with few samples, arriving at 10 final categories (see Appendix \ref{Appendix: Tasks and Query Sets}). 

\myparagraph{IP in action.} Fig. \ref{Fig: IP in action}(c) shows an example run of IP on the HuffPost dataset. Note that positive responses to queries are very sparse, since each extended headline only contains $8.6$ words on average out of the 1,000 in the vocabulary. As a result, IP asks $125$ queries before termination. As discussed in \S \ref{sec:related_work}, such long decision paths would be impossible in decision trees due to data fragmentation and memory limitations. For clarity of presentation we only show the $20$ queries with the greatest impact on the estimated posterior (as measured by KL-divergence from previous posterior). Upon reaching the first positive query ``eat'', the probability mass concentrates on the categories \textit{Food \& Drink} and \textit{Wellness} with little mass on \textit{Travel}. However, as the queries about the existence of ``citi'', ``visit'', ``york'', and ``bar'' in the extended headline come back positive, the model becomes more and more confident that ``Travel'' is the correct class. IP requires about $199.3$ queries on average to predict the topic of the extended headline with $\epsilon = 10^{-3}$ as the stopping criteria (See \eqref{eq: practical terminating condition}). Additional details on the HuffPost query set are in Appendix \ref{Appendix: Tasks and Query Sets}.

Further examples of IP performing inference on all tasks can be found in Appendix \ref{Appendix: Additional Example Runs}.

\subsection{Quantitative Evaluation}

\subsubsection{Classification Accuracy} \label{Section: Experiments.Models}

We compare the classification accuracy of our model's prediction based on the query-answers gathered by IP until termination with several other baseline models. For each of the models considered, we first give a brief description and then comment on their performance with respect to IP. All the results are summarized in Table \ref{fig: results table}.



\textsc{Decision Tree} refers to standard classification trees learnt using the popular CART algorithm \cite{breiman2017classification}. In the Introduction, we mentioned that classical decision trees learnt using $Q$ to supply the node splitting functions will be intepretable by construction but are not competitive with state-of-the-art methods. This is illustrated in our results in Table \ref{fig: results table}. Across all datasets, IP obtains superior performance since it is based on an underlying generative model (VAE) and only computes the branch of the tree traversed by the input data in an online manner, thus it is not shackled by data fragmentation and memory limitations.

\begin{figure*}
    \centering
    \includegraphics[width=0.8\linewidth]{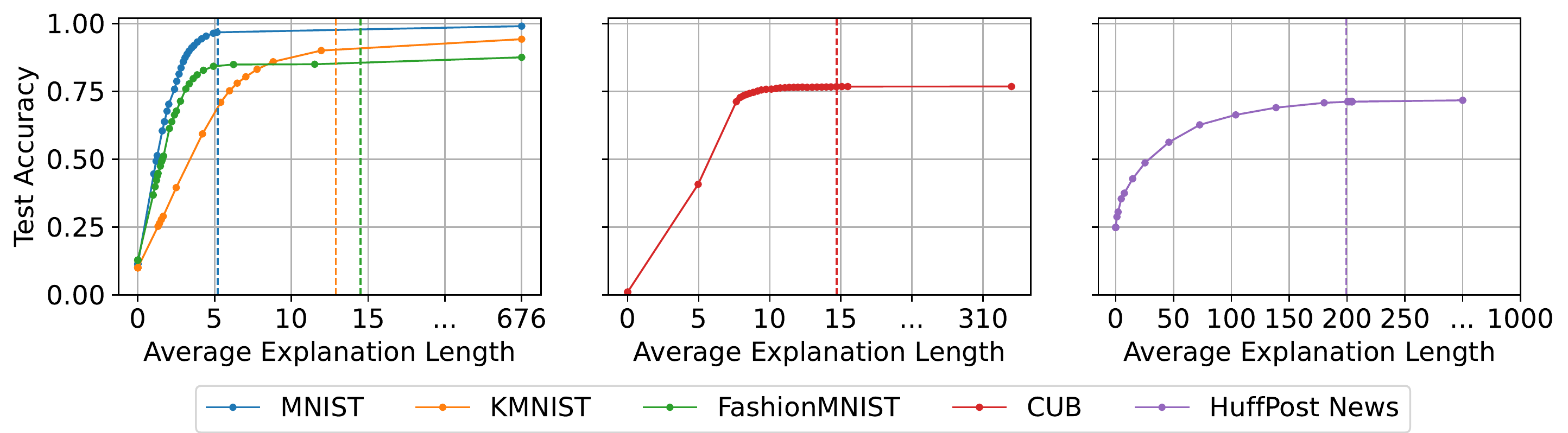}
    \caption{\textbf{Trade-off between predictive performance and explanation length} Different points along the curves correspond to different values of $\epsilon$ as the stopping criteria \eqref{eq: practical terminating condition} is varied. The colored dotted vertical line in each plot indicates the avg. explanation length v/s test accuracy at the $\epsilon$ value used as the stopping criteria for reporting results for the IP strategy in this work. For each plot, the x-axis ranges from $0$ to the size of the query set, $|Q|$, chosen for that task.}
    \label{fig:accuracy_minimality_plot}
\end{figure*}

\textsc{MAP using $Q$} refers to the Maximum A Posteriori estimate obtained using the posterior distribution over the labels given the answers to all the queries in $Q$ (for a given input). Recall, IP asks queries until the stopping criteria is reached (Equation \eqref{eq: practical terminating condition} \& Equation \eqref{eq: practical terminating condition for deterministic cases}). Naturally, there is a trade-off between the length of the explanations and the predictive performance observed. If  we ask all the queries then the resulting explanations of length $|Q|$ might be too long to be desirable. The results for IP reported in Table \ref{fig: results table} use different dataset-specific stopping criteria according to the elbow in their respective accuracy vs. explanation length curves (see Fig.~\ref{fig:accuracy_minimality_plot}). On the binary image datasets, (MNIST, KMNIST, and FashionMNIST) IP obtains an accuracy within $~3\%$ of the best achievable upon seeing all the query-answers with only about $2\%$ of the total queries in $Q$. Similarly for the CUB and Huffpost datasets, IP achieves about the same accuracy as \textsc{MAP using $Q$} but asks less than $5\%$ and $20\%$ of total possible queries respectively.

\textsc{Black-Box using $Q$} refers to the best performing deep network model we get by training on features supplied by evaluating all $q \in Q$ on input data from the various training datasets. For the binary image datasets, this is just a 4-layer CNN with ReLU activations. For CUB we use the results reported by the sequential model in \cite{koh2020concept}. For HuffPost, we found a single hidden layer with ReLU non-linearity give the best performance. Further architectural and training details are in Appendix \ref{Appendix: Comparison Models}.
In Table \ref{fig: results table} we show that across all datasets, the predictive performance obtained by \textsc{MAP Using $Q$} is on par with the best performance we obtained using black-box expressive non-interpretable networks \textsc{Black-Box using $Q$}. Thus, our generative models, which form the backbone for IP, are competitive with state-of-the-art prediction methods. 

\textsc{Black-Box} refers to the best performing black-box model on these datasets in terms of classification accuracy as reported in literature; to the best of our knowledge. In Table \ref{fig: results table}, we see a performance gap in each dataset when compared with \textsc{MAP using $Q$} which uses an interpretable query set. This is expected since explainability can be viewed as an additional constraint on learning. For example, on FashionMNIST we see an almost $8.5\%$ relative fall in accuracy due to binarization. This is because it is harder to decipher between some classes like shirts and pullovers at the binary level. On the other hand, binary patches are easily interpretable as edges, foregrounds and backgrounds. Similarly, there is a relative drop of accuracy of about $17\%$ for the HuffPost dataset since our queries regarding the existence of different words ignore their arrangement in sentences. Thus we lose crucial contextual information used by state-of-the-art transformer models \cite{devlin2018bert}. Ideally, we would like query sets to be easily interpretable, lead to short explanations and be sufficient to solve the task. Finding such query sets is nontrivial and will be explored in future work.

\begin{table}[h]
\centering
\begingroup
\renewcommand{\arraystretch}{1.5} 
\caption{\textbf{Classification accuracy of our model (Information Pursuit) relative to baselines on different test sets.} See \ref{Section: Experiments.Models} for details on each model.}
\label{fig: results table}
\resizebox{\columnwidth}{!}{\begin{tabular}{| l | c  | c | c | c | c |}
\hline
\textbf{Model} & \textbf{MNIST} & \textbf{KMNIST} & \textbf{Fashion} & \textbf{CUB} & \textbf{HuffPost} \\
\hline
\textsc{Information Pursuit} & 96.78\% & 91.02\% & 85.60\% & 76.73\% & 71.21\% \\
\textsc{Decision Tree}\cite{breiman2017classification} & 90.23\% & 78.00\% & 80.80\%  & 68.80\% & 63.00\%  \\
\hline
\textsc{MAP using $Q$} & 99.05\% & 94.25\% &  87.56\% &  76.80\% &  71.72\% \\
\textsc{Black-box using $Q$} & 99.15\% &  95.10\% &  88.43\% &  76.30\% &  71.48\% \\
\hline
\textsc{Black-box} & 99.83\%\cite{hu2018squeeze} &  98.83\% \cite{clanuwat2018deep} &  96.70\% \cite{fashion_github} &  82.70\% \cite{koh2020concept} & 86.45\% \tablefootnote{We fine-tuned a Bert Large Uncased Transformer model \cite{devlin2018bert} with the last layer replaced with a linear one. See Appendix \ref{Appendix: Black-Box} for details.} \\
\hline
\end{tabular}}
\endgroup
\end{table}

\subsubsection{Comparison to current attribution methods} \label{section: comparison to attribution methods}
At first glance, it might seem that using attribution methods/saliency maps can provide the same insights as to which parts of the image or more generally which queries in $Q$ were most influential in a decision made by a black-box model trained on input features supplied by all the query-answers. However, the unreliability of these methods in being faithful to the model they try to explain brings their utility into question \cite{adebayo2018sanity, yang2019benchmarking, shah2021input}. We conjecture that this is because current attribution methods are not designed to generate explanations that are sufficient statistics of the model's prediction. We illustrate this with a simple experiment using our binary image classification datasets.

For each input image $x$, we compute the corresponding attribution map $e(x)$ for the model's predicted class using two popular attribution methods, Integrated gradients (IG) \cite{sundararajan2017axiomatic} and DeepSHAP \cite{lundberg2017unified}. We then compute the $L$ most important $3 \times 3$ patches, where $L$ is the number of patches queried by IP for that particular input image. For computing the attribution/importance of a patch we average the attributions of all the pixels in that patch (following \cite{yang2019benchmarking}). We proceed as follows: {\it(i)} Given $e(x)$, compute the patch with maximum attribution and add these pixels to our explanation, {\it(ii)} Zero-out the attributions of all the pixels in the previously selected patch and repeat step {\it(i)} until $L$ patches are selected. The final explanation consists of $L$ possibly overlapping patches. Now, we evaluate the sufficiency of the generated explanation for the model's prediction by estimating the MAP accuracy of the posterior over labels given the intensities in the patches included in this explanation. This is done via a VAE trained to learn the joint distribution over image pixels and class labels. We experiment with both the raw attribution scores returned by IG and DeepSHAP and also the absolute values of the attribution scores for $e(x)$. The results are reported in Table \ref{Table: comparison_to_current_attrib_methods}. In almost all cases (with the exception of DeepSHAP on FashionMNIST), IP generates explanations that are more predictive of the class label than popular attribution methods.


\begin{table}[h]
\centering
\renewcommand{\arraystretch}{1.5} 
\caption{\textbf{MAP accuracy of explanations generated by Information Pursuit (IP) v/s other attribution methods.} IP explanations (in almost all cases) achieve a higher classification accuracy than explanations of the same length generated using baseline attribution methods. The (absolute) method refers to explanations generated using absolute values of the attribution map scores. On MNIST and KMNIST, IP explanations achieve a $10\%$ and $2.38\%$ relative improvement respectively over the best performing baseline method. On FashionMNIST, IP explanations are second best with a relative decrease of about $3.12\%$ from the best performing baseline.}
\label{Table: comparison_to_current_attrib_methods}
\begin{tabular}{| @{\;\;}l@{\;\;} | @{\;\;}c@{\;\;}  | @{\;\;}c@{\;\;} | @{\;\;}c@{\;\;} |}
\hline
\textbf{Explanation Method} & \textbf{MNIST} & \textbf{KMNIST} & \textbf{Fashion-MNIST} \\
\hline
\textsc{Information Pursuit} & \textbf{96.78\%} & \textbf{91.02\%} & 85.60\%  \\
\textsc{IG} & 78.48\% & 84.87\% & 78.49\%   \\
\textsc{IG (absolute)} & 70.39\% & 84.72\% & 64.95\%   \\
\textsc{DeepSHAP} & 87.98\% & 88.90\% &  \textbf{88.36}\%  \\
\textsc{DeepSHAP (absolute)} & 84.80\% & 84.56\% &  84.35\%  \\
\hline
\end{tabular}
\end{table}

\section{Conclusion}
\label{section: Conclusion}


We have presented a step towards building trustworthy interpretable machine learning models that respect the domain- and user-dependent nature of interpretability. We address this by composing user-defined, interpretable queries into concise explanations. Furthermore, unlike many contemporary attempts at explainability, our method is not post-hoc, but is \textit{interpretable by design} and guaranteed to produce faithful explanations. We formulate a tractable approach to implement this framework through deep generative models, MCMC algorithms, and the information pursuit algorithm. Finally, we demonstrate the effectiveness of our method across various vision and language tasks at generating concise explanations describing the underlying reasoning process behind the prediction. Future work will be aimed at extending the proposed framework to more complex tasks beyond classification such as scene parsing, image captioning, and sentiment analysis.

\ifCLASSOPTIONcompsoc
  \section*{Acknowledgments}
\else
  \section*{Acknowledgment}
\fi

The authors thank Mar\'{\i}a P\'erez Ortiz and John Shawe-Taylor for their contributions to the design of the experiments on document classification presented in Section \ref{Section: Word-based Queries}. This research was supported by the Army Research Office under the Multidisciplinary University Research Initiative contract W911NF-17-1-0304 and by the NSF grant 2031985.

\ifCLASSOPTIONcaptionsoff
  \newpage
\fi



\bibliographystyle{IEEEtran}
\bibliography{TPAMI_IP}

\begin{thebibliography}{10}
\providecommand{\url}[1]{#1}
\csname url@samestyle\endcsname
\providecommand{\newblock}{\relax}
\providecommand{\bibinfo}[2]{#2}
\providecommand{\BIBentrySTDinterwordspacing}{\spaceskip=0pt\relax}
\providecommand{\BIBentryALTinterwordstretchfactor}{4}
\providecommand{\BIBentryALTinterwordspacing}{\spaceskip=\fontdimen2\font plus
\BIBentryALTinterwordstretchfactor\fontdimen3\font minus
  \fontdimen4\font\relax}
\providecommand{\BIBforeignlanguage}[2]{{%
\expandafter\ifx\csname l@#1\endcsname\relax
\typeout{** WARNING: IEEEtran.bst: No hyphenation pattern has been}%
\typeout{** loaded for the language `#1'. Using the pattern for}%
\typeout{** the default language instead.}%
\else
\language=\csname l@#1\endcsname
\fi
#2}}
\providecommand{\BIBdecl}{\relax}
\BIBdecl

\bibitem{gunning2019xai}
D.~Gunning, M.~Stefik, J.~Choi, T.~Miller, S.~Stumpf, and G.-Z. Yang,
  ``Xai—explainable artificial intelligence,'' \emph{Science Robotics},
  vol.~4, no.~37, p. eaay7120, 2019.

\bibitem{rudin2019stop}
C.~Rudin, ``Stop explaining black box machine learning models for high stakes
  decisions and use interpretable models instead,'' \emph{Nature Machine
  Intelligence}, vol.~1, no.~5, pp. 206--215, 2019.

\bibitem{murdoch2019interpretable}
W.~J. Murdoch, C.~Singh, K.~Kumbier, R.~Abbasi-Asl, and B.~Yu, ``Interpretable
  machine learning: definitions, methods, and applications,'' \emph{arXiv
  preprint arXiv:1901.04592}, 2019.

\bibitem{eureg2019}
{European Commission}, ``Building trust in human-centric artificial
  intelligence,'' \emph{Communication from the Commission to the European
  Parliament, the Council, the European Economic and Social Committee and the
  Committee of the Regions}, vol. 168, 2019.

\bibitem{FDAminutes2021}
{United States Food and Drug Administration}, ``Virtual public workshop -
  transparency of artificial intelligence/machine learning-enabled medical
  devices,'' Transcript: \url{https://www.fda.gov/media/154423/download}, Oct.
  14, 2021.

\bibitem{johansson2011trade}
U.~Johansson, C.~S{\"o}nstr{\"o}d, U.~Norinder, and H.~Bostr{\"o}m, ``Trade-off
  between accuracy and interpretability for predictive in silico modeling,''
  \emph{Future medicinal chemistry}, vol.~3, no.~6, pp. 647--663, 2011.

\bibitem{wanner2021stop}
J.~Wanner, L.-V. Herm, K.~Heinrich, and C.~Janiesch, ``Stop ordering machine
  learning algorithms by their explainability! an empirical investigation of
  the tradeoff between performance and explainability,'' in \emph{Conference on
  e-Business, e-Services and e-Society}.\hskip 1em plus 0.5em minus 0.4em\relax
  Springer, 2021, pp. 245--258.

\bibitem{dovsilovic2018explainable}
F.~K. Do{\v{s}}ilovi{\'c}, M.~Br{\v{c}}i{\'c}, and N.~Hlupi{\'c}, ``Explainable
  artificial intelligence: A survey,'' in \emph{2018 41st International
  convention on information and communication technology, electronics and
  microelectronics (MIPRO)}.\hskip 1em plus 0.5em minus 0.4em\relax IEEE, 2018,
  pp. 0210--0215.

\bibitem{arrieta2020explainable}
A.~B. Arrieta, N.~D{\'\i}az-Rodr{\'\i}guez, J.~Del~Ser, A.~Bennetot, S.~Tabik,
  A.~Barbado, S.~Garc{\'\i}a, S.~Gil-L{\'o}pez, D.~Molina, R.~Benjamins
  \emph{et~al.}, ``Explainable artificial intelligence (xai): Concepts,
  taxonomies, opportunities and challenges toward responsible ai,''
  \emph{Information fusion}, vol.~58, pp. 82--115, 2020.

\bibitem{gunning2019darpa}
D.~Gunning and D.~Aha, ``Darpa’s explainable artificial intelligence (xai)
  program,'' \emph{AI magazine}, vol.~40, no.~2, pp. 44--58, 2019.

\bibitem{baehrens2010explain}
D.~Baehrens, T.~Schroeter, S.~Harmeling, M.~Kawanabe, K.~Hansen, and K.-R.
  M{\"u}ller, ``How to explain individual classification decisions,'' \emph{The
  Journal of Machine Learning Research}, vol.~11, pp. 1803--1831, 2010.

\bibitem{simonyan2013deep}
K.~Simonyan, A.~Vedaldi, and A.~Zisserman, ``Deep inside convolutional
  networks: Visualising image classification models and saliency maps,''
  \emph{arXiv preprint arXiv:1312.6034}, 2013.

\bibitem{kolek2022rate}
S.~Kolek, D.~A. Nguyen, R.~Levie, J.~Bruna, and G.~Kutyniok, ``A
  rate-distortion framework for explaining black-box model decisions,'' in
  \emph{International Workshop on Extending Explainable AI Beyond Deep Models
  and Classifiers}.\hskip 1em plus 0.5em minus 0.4em\relax Springer, 2022, pp.
  91--115.

\bibitem{shrikumar2017learning}
A.~Shrikumar, P.~Greenside, and A.~Kundaje, ``Learning important features
  through propagating activation differences,'' in \emph{International
  conference on machine learning}.\hskip 1em plus 0.5em minus 0.4em\relax PMLR,
  2017, pp. 3145--3153.

\bibitem{zeiler2014visualizing}
M.~D. Zeiler and R.~Fergus, ``Visualizing and understanding convolutional
  networks,'' in \emph{European conference on computer vision}.\hskip 1em plus
  0.5em minus 0.4em\relax Springer, 2014, pp. 818--833.

\bibitem{selvaraju2017grad}
R.~R. Selvaraju, M.~Cogswell, A.~Das, R.~Vedantam, D.~Parikh, and D.~Batra,
  ``Grad-cam: Visual explanations from deep networks via gradient-based
  localization,'' in \emph{Proceedings of the IEEE international conference on
  computer vision}, 2017, pp. 618--626.

\bibitem{smilkov2017smoothgrad}
D.~Smilkov, N.~Thorat, B.~Kim, F.~Vi{\'e}gas, and M.~Wattenberg, ``Smoothgrad:
  removing noise by adding noise,'' \emph{arXiv preprint arXiv:1706.03825},
  2017.

\bibitem{subramanya2019fooling}
A.~Subramanya, V.~Pillai, and H.~Pirsiavash, ``Fooling network interpretation
  in image classification,'' in \emph{Proceedings of the IEEE/CVF International
  Conference on Computer Vision}, 2019, pp. 2020--2029.

\bibitem{adebayo2018sanity}
J.~Adebayo, J.~Gilmer, M.~Muelly, I.~Goodfellow, M.~Hardt, and B.~Kim, ``Sanity
  checks for saliency maps,'' \emph{Advances in neural information processing
  systems}, vol.~31, 2018.

\bibitem{yang2019benchmarking}
M.~Yang and B.~Kim, ``Benchmarking attribution methods with relative feature
  importance,'' \emph{arXiv preprint arXiv:1907.09701}, 2019.

\bibitem{kindermans2019reliability}
P.-J. Kindermans, S.~Hooker, J.~Adebayo, M.~Alber, K.~T. Sch{\"u}tt,
  S.~D{\"a}hne, D.~Erhan, and B.~Kim, ``The (un) reliability of saliency
  methods,'' in \emph{Explainable AI: Interpreting, Explaining and Visualizing
  Deep Learning}.\hskip 1em plus 0.5em minus 0.4em\relax Springer, 2019, pp.
  267--280.

\bibitem{shah2021input}
H.~Shah, P.~Jain, and P.~Netrapalli, ``Do input gradients highlight
  discriminative features?'' \emph{Advances in Neural Information Processing
  Systems}, vol.~34, 2021.

\bibitem{slack2020fooling}
D.~Slack, S.~Hilgard, E.~Jia, S.~Singh, and H.~Lakkaraju, ``Fooling lime and
  shap: Adversarial attacks on post hoc explanation methods,'' in
  \emph{Proceedings of the AAAI/ACM Conference on AI, Ethics, and Society},
  2020, pp. 180--186.

\bibitem{koh2020concept}
P.~W. Koh, T.~Nguyen, Y.~S. Tang, S.~Mussmann, E.~Pierson, B.~Kim, and
  P.~Liang, ``Concept bottleneck models,'' in \emph{International Conference on
  Machine Learning}.\hskip 1em plus 0.5em minus 0.4em\relax PMLR, 2020, pp.
  5338--5348.

\bibitem{chen2019looks}
C.~Chen, O.~Li, D.~Tao, A.~Barnett, C.~Rudin, and J.~K. Su, ``This looks like
  that: deep learning for interpretable image recognition,'' \emph{Advances in
  neural information processing systems}, vol.~32, 2019.

\bibitem{rudin2022interpretable}
C.~Rudin, C.~Chen, Z.~Chen, H.~Huang, L.~Semenova, and C.~Zhong,
  ``Interpretable machine learning: Fundamental principles and 10 grand
  challenges,'' \emph{Statistics Surveys}, vol.~16, pp. 1--85, 2022.

\bibitem{janssen1997compositionality}
T.~M. Janssen and B.~H. Partee, ``Compositionality,'' in \emph{Handbook of
  logic and language}.\hskip 1em plus 0.5em minus 0.4em\relax Elsevier, 1997,
  pp. 417--473.

\bibitem{lakkaraju2016interpretable}
H.~Lakkaraju, S.~H. Bach, and J.~Leskovec, ``Interpretable decision sets: A
  joint framework for description and prediction,'' in \emph{Proceedings of the
  22nd ACM SIGKDD international conference on knowledge discovery and data
  mining}, 2016, pp. 1675--1684.

\bibitem{wan2021nbdt}
\BIBentryALTinterwordspacing
A.~Wan, L.~Dunlap, D.~Ho, J.~Yin, S.~Lee, S.~Petryk, S.~A. Bargal, and J.~E.
  Gonzalez, ``{\{}NBDT{\}}: Neural-backed decision tree,'' in
  \emph{International Conference on Learning Representations}, 2021. [Online].
  Available: \url{https://openreview.net/forum?id=mCLVeEpplNE}
\BIBentrySTDinterwordspacing

\bibitem{mu2020compositional}
J.~Mu and J.~Andreas, ``Compositional explanations of neurons,'' \emph{Advances
  in Neural Information Processing Systems}, vol.~33, pp. 17\,153--17\,163,
  2020.

\bibitem{jahangiri2017information}
E.~Jahangiri, E.~Yoruk, R.~Vidal, L.~Younes, and D.~Geman, ``Information
  pursuit: A bayesian framework for sequential scene parsing,'' \emph{arXiv
  preprint arXiv:1701.02343}, 2017.

\bibitem{bau2017network}
D.~Bau, B.~Zhou, A.~Khosla, A.~Oliva, and A.~Torralba, ``Network dissection:
  Quantifying interpretability of deep visual representations,'' in
  \emph{Proceedings of the IEEE conference on computer vision and pattern
  recognition}, 2017, pp. 6541--6549.

\bibitem{hernandez2022natural}
E.~Hernandez, S.~Schwettmann, D.~Bau, T.~Bagashvili, A.~Torralba, and
  J.~Andreas, ``Natural language descriptions of deep visual features,''
  \emph{arXiv preprint arXiv:2201.11114}, 2022.

\bibitem{breiman2017classification}
L.~Breiman, J.~H. Friedman, R.~A. Olshen, and C.~J. Stone, \emph{Classification
  and regression trees}.\hskip 1em plus 0.5em minus 0.4em\relax Routledge,
  2017.

\bibitem{quinlan1986induction}
J.~R. Quinlan, ``Induction of decision trees,'' \emph{Machine learning},
  vol.~1, no.~1, pp. 81--106, 1986.

\bibitem{amit1997shape}
Y.~Amit and D.~Geman, ``Shape quantization and recognition with randomized
  trees,'' \emph{Neural computation}, vol.~9, no.~7, pp. 1545--1588, 1997.

\bibitem{breiman2001random}
L.~Breiman, ``Random forests,'' \emph{Machine learning}, vol.~45, no.~1, pp.
  5--32, 2001.

\bibitem{caruana2006empirical}
R.~Caruana and A.~Niculescu-Mizil, ``An empirical comparison of supervised
  learning algorithms,'' in \emph{Proceedings of the 23rd international
  conference on Machine learning}, 2006, pp. 161--168.

\bibitem{fernandez2014we}
M.~Fern{\'a}ndez-Delgado, E.~Cernadas, S.~Barro, and D.~Amorim, ``Do we need
  hundreds of classifiers to solve real world classification problems?''
  \emph{The journal of machine learning research}, vol.~15, no.~1, pp.
  3133--3181, 2014.

\bibitem{xu2021deep}
H.~Xu, K.~A. Kinfu, W.~LeVine, S.~Panda, J.~Dey, M.~Ainsworth, Y.-C. Peng,
  M.~Kusmanov, F.~Engert, C.~M. White \emph{et~al.}, ``When are deep networks
  really better than decision forests at small sample sizes, and how?''
  \emph{arXiv preprint arXiv:2108.13637}, 2021.

\bibitem{kontschieder2015deep}
P.~Kontschieder, M.~Fiterau, A.~Criminisi, and S.~R. Bulo, ``Deep neural
  decision forests,'' in \emph{Proceedings of the IEEE international conference
  on computer vision}, 2015, pp. 1467--1475.

\bibitem{geman1996active}
D.~Geman and B.~Jedynak, ``An active testing model for tracking roads in
  satellite images,'' \emph{IEEE Transactions on Pattern Analysis and Machine
  Intelligence}, vol.~18, no.~1, pp. 1--14, 1996.

\bibitem{chattopadhay2018grad}
A.~Chattopadhay, A.~Sarkar, P.~Howlader, and V.~N. Balasubramanian,
  ``Grad-cam++: Generalized gradient-based visual explanations for deep
  convolutional networks,'' in \emph{2018 IEEE winter conference on
  applications of computer vision (WACV)}.\hskip 1em plus 0.5em minus
  0.4em\relax IEEE, 2018, pp. 839--847.

\bibitem{ribeiro2016should}
M.~T. Ribeiro, S.~Singh, and C.~Guestrin, ``" why should i trust you?"
  explaining the predictions of any classifier,'' in \emph{Proceedings of the
  22nd ACM SIGKDD international conference on knowledge discovery and data
  mining}, 2016, pp. 1135--1144.

\bibitem{lundberg2017unified}
S.~M. Lundberg and S.-I. Lee, ``A unified approach to interpreting model
  predictions,'' \emph{Advances in neural information processing systems},
  vol.~30, 2017.

\bibitem{kim2018interpretability}
B.~Kim, M.~Wattenberg, J.~Gilmer, C.~Cai, J.~Wexler, F.~Viegas \emph{et~al.},
  ``Interpretability beyond feature attribution: Quantitative testing with
  concept activation vectors (tcav),'' in \emph{International conference on
  machine learning}.\hskip 1em plus 0.5em minus 0.4em\relax PMLR, 2018, pp.
  2668--2677.

\bibitem{zhou2018interpretable}
B.~Zhou, Y.~Sun, D.~Bau, and A.~Torralba, ``Interpretable basis decomposition
  for visual explanation,'' in \emph{Proceedings of the European Conference on
  Computer Vision (ECCV)}, 2018, pp. 119--134.

\bibitem{yeh2020completeness}
C.-K. Yeh, B.~Kim, S.~Arik, C.-L. Li, T.~Pfister, and P.~Ravikumar, ``On
  completeness-aware concept-based explanations in deep neural networks,''
  \emph{Advances in Neural Information Processing Systems}, vol.~33, pp.
  20\,554--20\,565, 2020.

\bibitem{alvarez2018towards}
D.~Alvarez~Melis and T.~Jaakkola, ``Towards robust interpretability with
  self-explaining neural networks,'' \emph{Advances in neural information
  processing systems}, vol.~31, 2018.

\bibitem{bohle2021convolutional}
M.~Bohle, M.~Fritz, and B.~Schiele, ``Convolutional dynamic alignment networks
  for interpretable classifications,'' in \emph{Proceedings of the IEEE/CVF
  Conference on Computer Vision and Pattern Recognition}, 2021, pp.
  10\,029--10\,038.

\bibitem{wu2021optimizing}
M.~Wu, S.~Parbhoo, M.~C. Hughes, V.~Roth, and F.~Doshi-Velez, ``Optimizing for
  interpretability in deep neural networks with tree regularization,''
  \emph{Journal of Artificial Intelligence Research}, vol.~72, pp. 1--37, 2021.

\bibitem{pillai2021explainable}
V.~Pillai and H.~Pirsiavash, ``Explainable models with consistent
  interpretations,'' \emph{UMBC Student Collection}, 2021.

\bibitem{chen2020concept}
Z.~Chen, Y.~Bei, and C.~Rudin, ``Concept whitening for interpretable image
  recognition,'' \emph{Nature Machine Intelligence}, vol.~2, no.~12, pp.
  772--782, 2020.

\bibitem{de2019bias}
M.~De-Arteaga, A.~Romanov, H.~Wallach, J.~Chayes, C.~Borgs, A.~Chouldechova,
  S.~Geyik, K.~Kenthapadi, and A.~T. Kalai, ``Bias in bios: A case study of
  semantic representation bias in a high-stakes setting,'' in \emph{proceedings
  of the Conference on Fairness, Accountability, and Transparency}, 2019, pp.
  120--128.

\bibitem{galassi2020attention}
A.~Galassi, M.~Lippi, and P.~Torroni, ``Attention in natural language
  processing,'' \emph{IEEE Transactions on Neural Networks and Learning
  Systems}, vol.~32, no.~10, pp. 4291--4308, 2020.

\bibitem{pruthi2019learning}
D.~Pruthi, M.~Gupta, B.~Dhingra, G.~Neubig, and Z.~C. Lipton, ``Learning to
  deceive with attention-based explanations,'' \emph{arXiv preprint
  arXiv:1909.07913}, 2019.

\bibitem{craven1995extracting}
M.~Craven and J.~Shavlik, ``Extracting tree-structured representations of
  trained networks,'' \emph{Advances in neural information processing systems},
  vol.~8, 1995.

\bibitem{dancey2004decision}
D.~Dancey, D.~A. McLean, and Z.~A. Bandar, ``Decision tree extraction from
  trained neural networks,'' in \emph{Proceedings of the Nineteenth Conference
  on Artificial Intelligence}.\hskip 1em plus 0.5em minus 0.4em\relax American
  Association for Artificial Intelligence, 2004.

\bibitem{frosst2017distilling}
N.~Frosst and G.~Hinton, ``Distilling a neural network into a soft decision
  tree,'' \emph{arXiv preprint arXiv:1711.09784}, 2017.

\bibitem{roy2016monocular}
A.~Roy and S.~Todorovic, ``Monocular depth estimation using neural regression
  forest,'' in \emph{Proceedings of the IEEE conference on computer vision and
  pattern recognition}, 2016, pp. 5506--5514.

\bibitem{biccici2018conditional}
U.~C. Bi{\c{c}}ici, C.~Keskin, and L.~Akarun, ``Conditional information gain
  networks,'' in \emph{2018 24th International Conference on Pattern
  Recognition (ICPR)}.\hskip 1em plus 0.5em minus 0.4em\relax IEEE, 2018, pp.
  1390--1395.

\bibitem{murthy2016deep}
V.~N. Murthy, V.~Singh, T.~Chen, R.~Manmatha, and D.~Comaniciu, ``Deep decision
  network for multi-class image classification,'' in \emph{Proceedings of the
  IEEE conference on computer vision and pattern recognition}, 2016, pp.
  2240--2248.

\bibitem{vilalta1997global}
R.~Vilalta, G.~Blix, and L.~Rendell, ``Global data analysis and the
  fragmentation problem in decision tree induction,'' in \emph{European
  Conference on Machine Learning}.\hskip 1em plus 0.5em minus 0.4em\relax
  Springer, 1997, pp. 312--326.

\bibitem{tishby2000information}
N.~Tishby, F.~C. Pereira, and W.~Bialek, ``The information bottleneck method,''
  \emph{arXiv preprint physics/0004057}, 2000.

\bibitem{chaloner1995bayesian}
K.~Chaloner and I.~Verdinelli, ``Bayesian experimental design: A review,''
  \emph{Statistical Science}, pp. 273--304, 1995.

\bibitem{sznitman2010active}
R.~Sznitman and B.~Jedynak, ``Active testing for face detection and
  localization,'' \emph{IEEE Transactions on Pattern Analysis and Machine
  Intelligence}, vol.~32, no.~10, pp. 1914--1920, 2010.

\bibitem{cuturi2020noisy}
M.~Cuturi, O.~Teboul, Q.~Berthet, A.~Doucet, and J.-P. Vert, ``Noisy adaptive
  group testing using bayesian sequential experimental design,'' \emph{arXiv
  preprint arXiv:2004.12508}, 2020.

\bibitem{branson2014ignorant}
S.~Branson, G.~Van~Horn, C.~Wah, P.~Perona, and S.~Belongie, ``The ignorant led
  by the blind: A hybrid human--machine vision system for fine-grained
  categorization,'' \emph{International Journal of Computer Vision}, vol. 108,
  no.~1, pp. 3--29, 2014.

\bibitem{mnih2014recurrent}
V.~Mnih, N.~Heess, A.~Graves \emph{et~al.}, ``Recurrent models of visual
  attention,'' in \emph{Advances in neural information processing systems},
  2014, pp. 2204--2212.

\bibitem{elsayed2019saccader}
G.~Elsayed, S.~Kornblith, and Q.~V. Le, ``Saccader: improving accuracy of hard
  attention models for vision,'' in \emph{Advances in Neural Information
  Processing Systems}, 2019, pp. 702--714.

\bibitem{li2016glance}
M.~Li, S.~S. Ge, and T.~H. Lee, ``Glance and glimpse network: A stochastic
  attention model driven by class saliency,'' in \emph{Asian Conference on
  Computer Vision}.\hskip 1em plus 0.5em minus 0.4em\relax Springer, 2016, pp.
  572--587.

\bibitem{li2019visual}
H.~Li, P.~Wang, C.~Shen, and A.~v.~d. Hengel, ``Visual question answering as
  reading comprehension,'' in \emph{Proceedings of the IEEE/CVF Conference on
  Computer Vision and Pattern Recognition}, 2019, pp. 6319--6328.

\bibitem{malinowski2017ask}
M.~Malinowski, M.~Rohrbach, and M.~Fritz, ``Ask your neurons: A deep learning
  approach to visual question answering,'' \emph{International Journal of
  Computer Vision}, vol. 125, no.~1, pp. 110--135, 2017.

\bibitem{mao2019neuro}
J.~Mao, C.~Gan, P.~Kohli, J.~B. Tenenbaum, and J.~Wu, ``The neuro-symbolic
  concept learner: Interpreting scenes, words, and sentences from natural
  supervision,'' \emph{arXiv preprint arXiv:1904.12584}, 2019.

\bibitem{shih2016look}
K.~J. Shih, S.~Singh, and D.~Hoiem, ``Where to look: Focus regions for visual
  question answering,'' in \emph{Proceedings of the IEEE conference on computer
  vision and pattern recognition}, 2016, pp. 4613--4621.

\bibitem{lu2016hierarchical}
J.~Lu, J.~Yang, D.~Batra, and D.~Parikh, ``Hierarchical question-image
  co-attention for visual question answering,'' \emph{Advances in neural
  information processing systems}, vol.~29, 2016.

\bibitem{andreas2016neural}
J.~Andreas, M.~Rohrbach, T.~Darrell, and D.~Klein, ``Neural module networks,''
  in \emph{Proceedings of the IEEE conference on computer vision and pattern
  recognition}, 2016, pp. 39--48.

\bibitem{laurent1976constructing}
H.~Laurent and R.~L. Rivest, ``Constructing optimal binary decision trees is
  np-complete,'' \emph{Information processing letters}, vol.~5, no.~1, pp.
  15--17, 1976.

\bibitem{belghazi2018mine}
M.~I. Belghazi, A.~Baratin, S.~Rajeswar, S.~Ozair, Y.~Bengio, A.~Courville, and
  R.~D. Hjelm, ``Mine: mutual information neural estimation,'' \emph{arXiv
  preprint arXiv:1801.04062}, 2018.

\bibitem{kingma2013auto}
D.~P. Kingma and M.~Welling, ``Auto-encoding variational bayes,'' \emph{arXiv
  preprint arXiv:1312.6114}, 2013.

\bibitem{doucet2009tutorial}
A.~Doucet, A.~M. Johansen \emph{et~al.}, ``A tutorial on particle filtering and
  smoothing: Fifteen years later,'' \emph{Handbook of nonlinear filtering},
  vol.~12, no. 656-704, p.~3, 2009.

\bibitem{jalal2021instance}
A.~Jalal, S.~Karmalkar, A.~Dimakis, and E.~Price, ``Instance-optimal compressed
  sensing via posterior sampling,'' \emph{Proceedings of Machine Learning
  Research}, vol. 139, 2021.

\bibitem{nijkamp2020anatomy}
E.~Nijkamp, M.~Hill, T.~Han, S.-C. Zhu, and Y.~N. Wu, ``On the anatomy of
  mcmc-based maximum likelihood learning of energy-based models,'' in
  \emph{Proceedings of the AAAI Conference on Artificial Intelligence},
  vol.~34, no.~04, 2020, pp. 5272--5280.

\bibitem{durmus2019high}
A.~Durmus and E.~Moulines, ``High-dimensional bayesian inference via the
  unadjusted langevin algorithm,'' \emph{Bernoulli}, vol.~25, no.~4A, pp.
  2854--2882, 2019.

\bibitem{welling2011bayesian}
M.~Welling and Y.~W. Teh, ``Bayesian learning via stochastic gradient langevin
  dynamics,'' in \emph{Proceedings of the 28th international conference on
  machine learning (ICML-11)}.\hskip 1em plus 0.5em minus 0.4em\relax Citeseer,
  2011, pp. 681--688.

\bibitem{lecun1998gradient}
Y.~LeCun, L.~Bottou, Y.~Bengio, and P.~Haffner, ``Gradient-based learning
  applied to document recognition,'' \emph{Proceedings of the IEEE}, vol.~86,
  no.~11, pp. 2278--2324, 1998.

\bibitem{xiao2017fashion}
H.~Xiao, K.~Rasul, and R.~Vollgraf, ``Fashion-mnist: a novel image dataset for
  benchmarking machine learning algorithms,'' \emph{arXiv preprint
  arXiv:1708.07747}, 2017.

\bibitem{clanuwat2018deep}
T.~Clanuwat, M.~Bober-Irizar, A.~Kitamoto, A.~Lamb, K.~Yamamoto, and D.~Ha,
  ``Deep learning for classical japanese literature,'' \emph{arXiv preprint
  arXiv:1812.01718}, 2018.

\bibitem{wah2011caltech}
C.~Wah, S.~Branson, P.~Welinder, P.~Perona, and S.~Belongie, ``The caltech-ucsd
  birds-200-2011 dataset,'' 2011.

\bibitem{huffpostdataset}
R.~Misra, ``News category dataset,'' 06 2018.

\bibitem{lavin2019analyzing}
M.~Lavin, ``Analyzing documents with tf-idf,'' 2019.

\bibitem{devlin2018bert}
J.~Devlin, M.-W. Chang, K.~Lee, and K.~Toutanova, ``Bert: Pre-training of deep
  bidirectional transformers for language understanding,'' \emph{arXiv preprint
  arXiv:1810.04805}, 2018.

\bibitem{hu2018squeeze}
J.~Hu, L.~Shen, and G.~Sun, ``Squeeze-and-excitation networks,'' in
  \emph{Proceedings of the IEEE conference on computer vision and pattern
  recognition}, 2018, pp. 7132--7141.

\bibitem{fashion_github}
H.~Xiao, K.~Rasul, and R.~Vollgraf, ``Fashion-mnist: A mnist-like fashion
  product database,'' in \emph{GitHub}, 2017.

\bibitem{sundararajan2017axiomatic}
M.~Sundararajan, A.~Taly, and Q.~Yan, ``Axiomatic attribution for deep
  networks,'' in \emph{International conference on machine learning}.\hskip 1em
  plus 0.5em minus 0.4em\relax PMLR, 2017, pp. 3319--3328.

\end{thebibliography}
%

%

\begin{IEEEbiography}[{\includegraphics[width=1in,height=1.25in,clip,keepaspectratio]{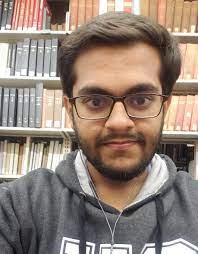}}]{Aditya Chattopadhyay}
is a PhD student in the Computer Science Department, Johns Hopkins University. He received the Bachelor of Technology degree in Computer Science and Master of Science by Research degree in Computational Natural Sciences from the International Institute of Information Technology, Hyderabad in 2016 and 2018 respectively. His research interests include explainable AI, probabilistic graphical models and Bayesian inference.

\end{IEEEbiography}

\begin{IEEEbiography}[{\includegraphics[width=1in,height=1.25in,clip,keepaspectratio]{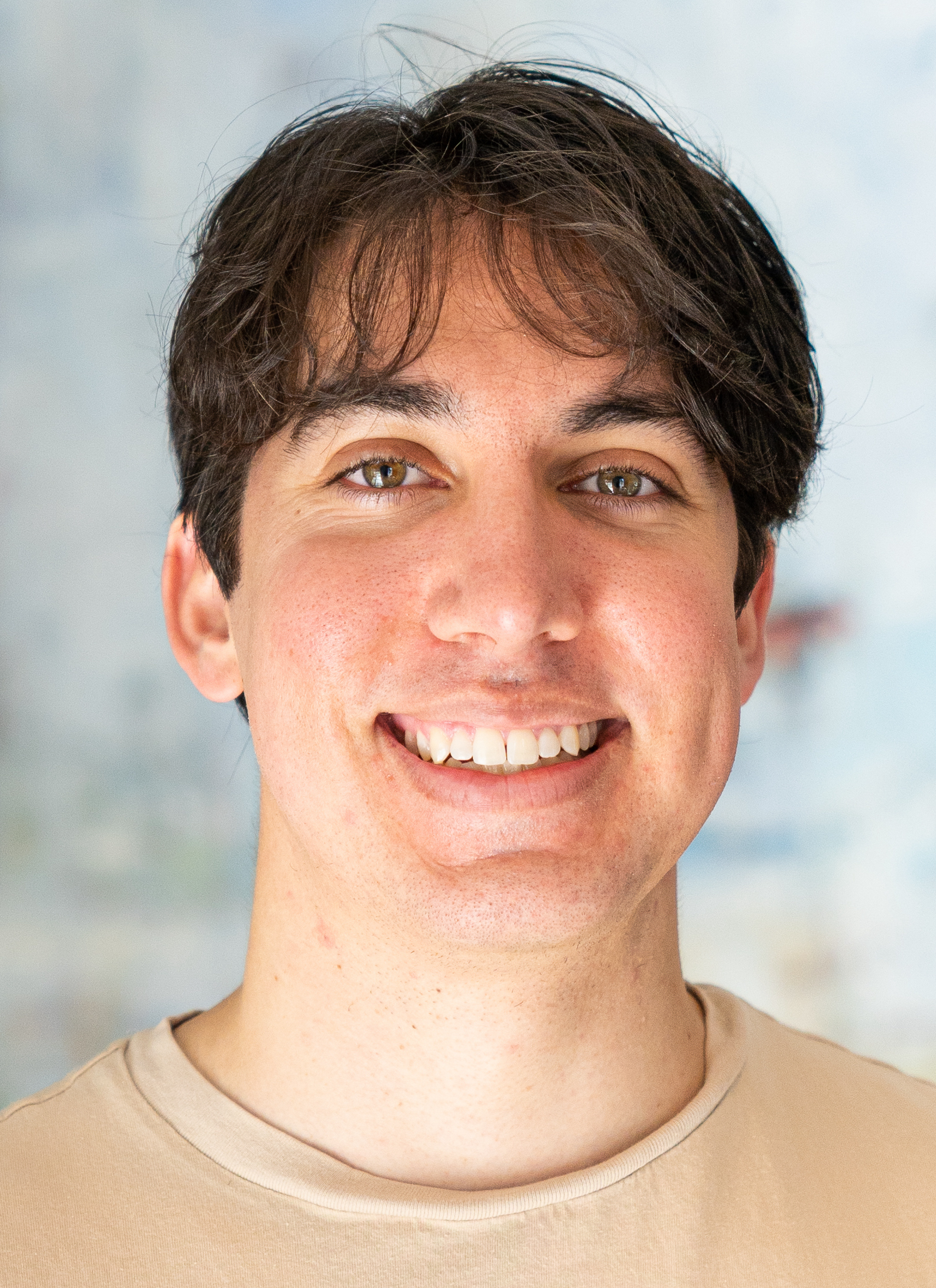}}]{Stewart Slocum} received his BS in computer science and applied mathematics from Johns Hopkins University in 2021. His research interests center on principled deep learning methods with performance and robustness guarantees.
\end{IEEEbiography}
\vspace{-1mm}
\begin{IEEEbiography}[{\includegraphics[width=1in,height=1.25in,clip,keepaspectratio]{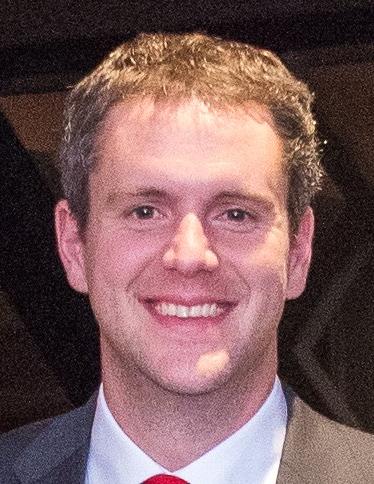}}]{Benjamin Haeffele}
is an Associate Research Scientist in the Mathematical Institute for Data Science at Johns Hopkins University. His research interests involve developing theory and algorithms for processing high-dimensional data at the intersection of machine learning, optimization, and computer vision. In addition to basic research in data science he also works on a variety of applications in medicine, microscopy, and computational imaging. He received his Ph.D. in Biomedical Engineering at Johns Hopkins University in 2015 and his B.S. in Electrical Engineering from the Georgia Institute of Technology in 2006.
\end{IEEEbiography}


\begin{IEEEbiography}[{\includegraphics[width=1in,height=1.25in,clip,keepaspectratio]{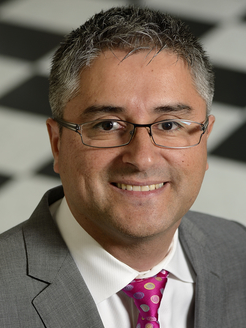}}]{Ren\'e Vidal}
received his B.S. degree in Electrical Engineering (valedictorian) from the Pontificia Universidad Cat\'olica de Chile in 1997 and his M.S.
and Ph.D. degrees in Electrical Engineering and Computer Science from
the University of California at Berkeley in 2000 and 2003, respectively. He is currently the Director of the Mathematical Institute for Data Science (MINDS) and the Hershel L. Seder Professor of Department of Biomedical Engineering at The Johns Hopkins University, where he has been since
2004. He is co-author of the book “Generalized Principal Component Analysis” (Springer 2016), co-editor of the book “Dynamical Vision” (Springer 2006) and co-author of over 300 articles in machine learning, computer vision, signal and image processing, biomedical image analysis,
hybrid systems, robotics and control. He is or has been Associate Editor in Chief of the IEEE Transactions on Pattern Analysis and Machine
Intelligence and Computer Vision and Image Understanding, Associate Editor or Guest Editor of Medical Image Analysis, the IEEE Transactions
on Pattern Analysis and Machine Intelligence, the SIAM Journal on Imaging Sciences, Computer Vision and Image Understanding, the
Journal of Mathematical Imaging and Vision, the International Journal
on Computer Vision and Signal Processing Magazine. He has received
numerous awards for his work, including 
the 2021 Edward J. McCluskey Technical Achievement Award,
the 2016 D'Alembert Faculty Fellowship, 
the 2012 IAPR J.K. Aggarwal Prize, the 2009 ONR Young
Investigator Award, the 2009 Sloan Research Fellowship and the 2005
NSF CAREER Award. He is a Fellow of the IEEE, Fellow of IAPR, Fellow
of AIMBE, and a member of the ACM and SIAM.
\end{IEEEbiography}

\begin{IEEEbiography}[{\includegraphics[width=1in,height=1.25in,clip,keepaspectratio]{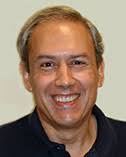}}]{Donald Geman}
(Life Senior Member, IEEE) received the B.A. degree in literature from the University of Illinois and the Ph.D. degree in mathematics from Northwestern University. He was a Distinguished Professor with the University of Massachusetts until 2001, when he joined the Department of Applied Mathematics and Statistics, Johns Hopkins University, where he is currently a member of the Center for Imaging Science and the Institute for Computational Medicine. His current research interests include statistical learning, computer vision, and computational biology. He is a member of the National Academy of Sciences and a fellow of the IMS and SIAM.
\end{IEEEbiography}




\onecolumn
\appendices
\newtheorem*{proposition*}{Proposition}
\newtheorem*{lemma*}{Lemma}
\newtheorem*{theorem*}{Theorem}
\section{}
In proofs of propositions and lemmas we rewrite the statement (un-numbered) for convenience. 
\subsection{Characterizing the optimal strategy $\pi^*$}
\label{Appendix: Characterizing the optimal strategy}

In this subsection we characterize the optimal strategy for any such arbitrary query set chosen by the user. 

\begin{proposition*}
Assuming $Q$ is finite and when $d$ is taken to be the KL-divergence then objective \eqref{eq: complexity definition} can be rewritten as,
\begin{align}
     &H_Q^{\epsilon}(X; Y) := \min_{\pi} \E_{X}\left[|expl^{\pi}_Q(X)|\right] \\
    &\textrm{s.t.} \;  \sum_{k=1}^{\tau^{\pi}}I(Y; S^{\pi}_k(X) \mid S^{\pi}_{k - 1}(X)) \geq I(X; Y) - \epsilon \nonumber
\end{align}
where, $\tau^{\pi} = \max\{t^{\pi}(x): x \in \gX\}$ and $t^{\pi}(X)$ is defined as the number of queries selected by $\pi$ for input $X$ until $q_{STOP}$. We define $S^{\pi}_{k}(X)$ as a random variable where any realization $S^{\pi}_{k}(x^{\textrm{obs}})$, $x^{\textrm{obs}} \in \gX$, denotes the event 
\[ S^{\pi}_k(x^{\textrm{obs}}) := \{x' \in \gX \mid \{q_i, q_i(x^{\textrm{obs}})\}_{1:k} = \{q_i, q_i(x')\}_{1:k} \},\]
where $q_i$ is the $i^{th}$ query selected by $\pi$ for input $x^{\textrm{obs}}$.
Here we use the convention that $S^{\pi}_{0}(X) = \Omega$ (the entire sample space) and $S^{\pi}_{l}(X) = S^{\pi}_{t^{\pi}(X)}(X)$ $\; \forall l > t^{\pi}(X)$.

\end{proposition*}
\begin{proof}
 We begin by reformulating our sufficiency constraint in \eqref{eq: complexity definition} in terms of entropy by taking the ``distance" between probability distributions as the KL divergence.\footnote{In favour of a clearer exposition, we abuse notation here and use $p(Y \mid expl^{\pi}_Q(X))$ to denote $P(Y \mid S^{\pi}_{t^{\pi}(X)}(X))$. In reality, $expl^{\pi}_Q(x)$ refers to the sequence of query-answer pairs chosen for $x$ as defined in \eqref{def: expl} whereas $S^{\pi}_{t^{\pi}(X = x)}(X = x)$ refers to the event which is the set of all possible data-points that agree on the first $t^{\pi}(x)$ query-answers observed for $x$.} The $\epsilon$-Sufficiency constraint can then be rewritten as,
\begin{equation*}
    \begin{aligned}
    \epsilon &\geq \mathbb{E}_X[KL\left(p(Y \mid X), p(Y \mid expl^{\pi}_Q(X)\right)] \\
    &= \mathbb{E}_X\left[\mathbb{E}_{Y}\left[\log \frac{p(Y \mid X)}{p(Y \mid expl^{\pi}_Q(X))} \mid X\right]\right] \\
     &= \mathbb{E}_X\left[\mathbb{E}_{Y}\left[\log \frac{p(Y \mid X)}{P(Y)} \mid X\right]\right] + \mathbb{E}_X\left[\mathbb{E}_{Y}\left[\log \frac{p(Y)}{P(Y \mid expl^{\pi}_Q(X))} \mid X\right]\right] \\
     &= I(X; Y) - I(expl^{\pi}_Q(X); Y) \\
     &= H(Y \mid expl^{\pi}_Q(X)) - H(Y \mid X)
    \end{aligned}
\end{equation*}
In the third equality we multiplied the term inside the $\log$ by the identity $\frac{P(Y)}{P(Y)} = 1$. The fifth inequality is by definition of mutual information. Thus, we can rewrite \eqref{eq: complexity definition} as,



\begin{align}
    \label{eq: complexity definition 2}
    H^{\epsilon}_Q(X; Y) := &\min_{\pi} \E_{X}\left[|expl^{\pi}_Q(X)|\right] \\
    &\textrm{s.t.} \; H(Y \mid expl^{\pi}_Q(X)) - H(Y \mid X) \leq \epsilon \quad (\epsilon\textit{-Sufficiency)} \nonumber
\end{align}

Let's define $t^{\pi}(X)$ as the number of queries selected by $\pi$ for input $X$ until $q_{STOP}$. Define $\tau^{\pi} = \max\{t^{\pi}(X): X \in \gX\}$. For the purpose of analysis we can vacuously modify $\pi$ such that for any given $x^{obs} \in \gX$, $\pi$ asks a fixed $\tau^{\pi}$ number of queries by filling the remaining $\tau^{\pi} - t^{\pi}(x^{obs})$ queries with $q_{STOP}$. An immediate consequence of this modification is that $S^{\pi}_{l}(X) = S^{\pi}_{t^{\pi}(X)}(X)$ $\; \forall l > t^{\pi}(X)$.

We will now show that the sufficiency criteria $H(Y \mid expl^{\pi}_Q(X)) - H(Y \mid X)$ can be rewritten as a sum of successive mutual information terms. 

\begin{equation}
    \begin{aligned}
        &H(Y \mid expl^{\pi}_Q(X)) - H(Y \mid X) \\
        &= (H(Y) - H(Y \mid X)) - (H(Y) - H(Y \mid expl^{\pi}_Q(X)))\\
        &= I(X; Y) - (H(Y) -  H(Y \mid S^{\pi}_{\tau}(X))\\
    \end{aligned}
\end{equation}

The third equality uses the fact that $H(Y \mid expl^{\pi}_Q(X)) = H(Y \mid S^{\pi}_{\tau}(X))$ since the $\forall x \in X$, $S^{\pi}_{\tau}(x) = S^{\pi}_{t^{\pi}(x)}(x)$.

We can now write $H(Y) -  H(Y \mid S^{\pi}_{\tau}(X))$ as a telescoping series,

\begin{equation}
    \begin{aligned}
        H(Y) -  H(Y \mid S^{\pi}_{\tau}(X) &= H(Y) - H(Y \mid S^{\pi}_1(X)) + H(Y \mid S^{\pi}_1(X)) 
        - H(Y \mid S^{\pi}_1(X), S^{\pi}_2(X)) \\&+ H(Y \mid S^{\pi}_1(X), S^{\pi}_2(X)) \ldots + H(Y \mid S^{\pi}_1(X), S^{\pi}_2(X), \ldots, S^{\pi}_{\tau - 1}(X)) - H(Y \mid S^{\pi}_{\tau}(X)) \\
        &= I(Y; S^{\pi}_1(X)) + I(Y; S^{\pi}_2(X) \mid S^{\pi}_1(X))) + \ldots + I(Y; S^{\pi}_{\tau}(X) \mid S^{\pi}_{\tau - 1}(X))
    \end{aligned}
\end{equation}

The last equality is obtained by noticing that 
\begin{enumerate}
    \item $H(Y \mid S^{\pi}_1(X), \ldots, S^{\pi}_{k - 1}(X)) - H(Y \mid S^{\pi}_1(X), \ldots, S^{\pi}_{k}(X)) = I(Y; S^{\pi}_{k}(X) \mid S^{\pi}_{k - 1}(X))$,
    \item $H(Y \mid S^{\pi}_{\tau}(X)) = H(Y \mid S^{\pi}_{1}(X), S^{\pi}_{2}(X), \ldots, S^{\pi}_{\tau}(X))$ since the events $\{S^{\pi}_{k}(x)\}_{1:\tau}$ are nested $\forall x \in \gX$.
\end{enumerate}

Putting it all together we can rewrite \eqref{eq: complexity definition 2} as, 
\begin{align}
    &H_Q^{\epsilon}(X; Y) := \min_{\pi} \E_{X}\left[|expl^{\pi}_Q(X)|\right] \\
    &\textrm{s.t.} \; I(X; Y) - \sum_{k=1}^{\tau^{\pi}}I(Y; S^{\pi}_k(X) \mid S^{\pi}_{k - 1}(X)) \leq \epsilon \nonumber
\end{align}

\end{proof}

\subsection{Approximation guarantees for \textrm{IP}}
\label{Appendix: proof of performance guarantee IP}
In Proposition \ref{prop: IP guarantees} we prove that the $\textrm{IP}$ strategy comes within $1$ bit of $H(Y)$ (the entropy of $Y$) under the assumption that one has access to all possible binary functions of $X$, that are also binary functions of $Y$, as queries. Given such a query set, it is well-known that the optimal strategy $\pi^*$ is given by the Huffman Code for $Y$ which is also within $1$ bit of $H(Y)$. Thus, the result is immediate that $\textrm{IP}$ asks at most 1 query more than $\pi^*$ on average. We restate Proposition \ref{prop: IP guarantees} below for ease.

\begin{proposition*} Let $Y$ be discrete. Let $\tilde{H}_Q(X; Y)$ be the expected description length obtained by the $\textrm{IP}$ strategy. If $H(Y | X) = 0$ and $Q$ is the set of all possible binary functions of $X$ such that $H(q(X) \mid Y) = 0$ $\forall q \in Q$, then $H(Y) \leq \tilde{H}_Q(X; Y) \leq H(Y) + 1$.\!\!
\end{proposition*}

We make two remarks before turning to the proof.\\
\noindent
{\bf Remark 1:} 
We have observed data $X$, a categorical discrete r.v. $Y$, and binary queries $\{q(X), q \in {\mathcal Q}\}$ from which $Y$ can be estimated.
In fact, let's suppose that $Y$ is determined by $X$, say $Y=f(X)$; equivalently, $H(Y|X)=0$.
We assume that $Y$ represents some high-level, possibly semantic, interpretation of $X$ which can only be seen through the
eyes of the queries $q \in Q$.  Ideally, we would like to be able to query the membership of $Y$ in any subset of $\mathcal Y$
(the set of possible values of $Y$); this is the information we would have in coding $Y$.   We will say that the query $I_{Y \in D}$ is {\it realizable} for some
$D \subset {\mathcal Y}$ if the query $I_{X \in f^{-1}(D)}$ is in ${\mathcal Q}$, i.e., we can test for $Y \in D$ by one of our
observable data queries.  In general, not all subsets of $\mathcal Y$ can be associated with attributes or features of $X$, e.g., ``Napolean'' or
``Dead'' in ``20 Questions'', or ``black beak'' in bird species classification.  If there was a query $q(X)$ for every subset $D$ of values of $Y$,
then our theorem says that mean number of queries needed to determine the state of $Y$ with IP is bounded below by $H(Y)$ and above by
$H(Y)+1$.

\noindent
{\bf Remark 2:} The sequence of queries $q_1,q_2,...$ generated by the
IP algorithm for a particular data point can be seen as one branch,
root to leaf, of a decision tree constructed by the standard machine
learning strategy based on successive reduction of uncertainty about $Y$ as
measured by mutual information: $q_1 =
\argmax_{q \in Q} I(q(X);Y), q_{k+1}= \argmax_{q \in
Q}I(q(X);Y|S^{\textrm{IP}}_k(x^0))$ where the $S^{\textrm{IP}}_k(x^0)$ is the event that for
the first $k$ questions the answers agree with those for $x^0$.  We
stop as soon as $Y$ is determined. Whereas a decision tree 
accommodates all $x$ simultaneously, the questions along the branch depends
on having a particular, fixed data point.  But the learning problem
in the branch version (``active testing'') is exponentially
simpler.

\medskip

\begin{proof}
The lower bound $H(Y) \leq \tilde{H}_Q(X; Y)$ comes from Shannon's source coding theorem for stochastic source $Y$.\\ Now for the upper bound,  
since $I(q(X);Y \mid S^{\textrm{IP}}_k(x^0))=H(q(X)|S^{\textrm{IP}}_k(x^0))-H(q(X)|Y,S^{\textrm{IP}}_k(x^0))$ and
since $Y$ determines $q(Y)$ and hence also $q(X)$, the second
entropy term is zero (since given $H(q(X)\mid Y) = 0)$.  So our problem is maximize the conditional entropy
of the binary random variable $q(X)$ given $S^{\textrm{IP}}_k(x^0)$. So
the IP algorithm is clearly just
``divide and conquer'':
\[ q_1 = \argmax_{q \in Q} H(q(X)),\]
\[ q_{k+1}= \argmax_{q \in Q}H(q(X)|S^{\textrm{IP}}_k(x^0)).\]
Equivalently, since entropy of a binary random variable $\rho$ is maximized when $P(\rho) = \frac{1}{2}$,
\[  q_{k+1} = \arg \min_{q \in Q}|P(q(X)= 1|S^{\textrm{IP}}_k(x^0)) - \frac{1}{2}|.\]

Let ${\mathcal Y}_k$ be the set of ``active hypotheses'' after
$k$ queries (denoted as $\gA_k$), namely those $y$ with positive posterior probability:
$P(Y=y|S^{\textrm{IP}}_k(x^0))>0$.  
Indeed, 
\begin{equation*}
\begin{aligned}
P(Y=y|S^{\textrm{IP}}_k(x^0)) &= \frac{P(S^{\textrm{IP}}_k(x^0)|Y=y)p(y)}{\sum_y P(S^{\textrm{IP}}_k(x^0)|Y=y)p(y)} \\
&= \frac{1_{{\mathcal Y}_k} p(k)}{\sum_{y \in {\mathcal A}_k}p(y)}
\end{aligned}
\end{equation*}
since 
\[P(S^{\textrm{IP}}_k(x^0)|Y=y) = \left\{ \begin{array}{ll}
1, & \textrm{if $y \in {\mathcal A}_k$} \\
0, & y \notin {\mathcal A}_k \end{array}\right.\]

In particular, the classes in the active set 
have the {\it same relative weights} as at the outset.  In summary: 


\begin{center}
$p(y|S^{\textrm{IP}}_k(x^0)) = \left\{ \begin{array}{ll}
p(y)/\sum_{{\mathcal A}_k}p(l), \,\, 
y \in {\mathcal A}_k\\

	0, & \textrm{otherwise}
	\end{array}\right.$
\end{center}

The key observation to prove the theorem
is that if a hypothesis $y$ generates the same answers
to the first $m$ or more questions as $y^0$,
and hence is active at step $m$, then
its prior likelihood $p(y)$ is at most
$2^{-(m-1)},\,\, m =1,2,\ldots$.
This is intuitively clear: if $y$ has the same answer as
$y^0$ on the first question, and $p(y^0) > \frac{1}{2}$, then
only one question is needed and the active set is empty
at step two; if ${q_1}(y)={q_1}(y^0)$ and ${q_2}(y)={q_2}(y^0)$
and $p(y^0) > \frac{1}{4}$, then
only two question are needed and the active set is empty
at step three, etc.

Finally, since $C$, the code length,
takes values in the non-negative integers
$\{0,1,\ldots, \}$:
\begin{eqnarray*}
\tilde{H}_Q(X; Y) &:=& \E[C] \\ &=& \sum_{m=1}^\infty P(C \geq m) \\
		&\leq& \sum_{m=1}^\infty P(p(Y) < 2^{-(m-1)}) \\
		&=& \sum_{m=1}^\infty \sum_{y: p(y) < 2^{-(m-1)}} p(y)\\
                &=& \sum_{y \in {\mathcal Y}} \sum_{m=1}^\infty 
1_{\{p(y) < 2^{-(m-1)}\}}p(k) \\
	        &=& \sum_{y \in {\mathcal Y}}p(k) (1 - \log p(k)) \\
		& = & H(Y) + 1 
\end{eqnarray*}
\end{proof}

\subsection{Termination Criteria for $\textrm{IP}$}
\label{Appendix: termination criteria for IP}
We would first analyze the termination criteria for the exact case, that is, $\epsilon = 0$ in \eqref{eq: complexity definition}, and then move on to the more general case.

\myparagraph{Termination Criteria when $\boldsymbol{\epsilon = 0}$} Ideally for a given input $x^{\textrm{obs}}$, we would like to terminate ($\textrm{IP}$ outputs $q_{STOP}$) after $L$ steps if 
\begin{equation}
    p(y \mid x^{\textrm{obs}}) = p(y \mid x') 
   \; \forall x' \in S^{\textrm{IP}}_L(x^{\textrm{obs}}),\ y \in \gY
    \label{eq: terminating criteria}
\end{equation}
Recall, $S^{\textrm{IP}}_L(x^{\textrm{obs}}) = \{x' \in \gX \mid \{q_i, q_i(x')\}_{1:L} = \{q_i, q_i(x^{\textrm{obs}})\}_{1:L}\}$. In other words, its the event consisting of all $x' \in \gX$ which share the first $L$ query-answer pairs with $x^{\textrm{obs}}$. 

If \eqref{eq: terminating criteria} holds for all $x^{\textrm{obs}} \in \gX$, the it is easy to see that this is equivalent to the sufficiency constraint in the case $\epsilon=0$,
\[p(y \mid x) = p(y \mid expl^{\textrm{IP}}_Q(x)) \;  \forall (x, y) \times (\gX \times \gY)\]
where $p(y \mid expl^{\textrm{IP}}_Q(x)) := p(y \mid S^{\textrm{IP}}_{t^{\textrm{IP}}(x^{\textrm{obs}})}(x^{\textrm{obs}})) \;  \forall (x, y) \in (\gX \times \gY)$ and $t^{\textrm{IP}}(x)$ is the number of iterations IP takes on input $x$ before termination. 

Unfortunately, detecting \eqref{eq: terminating criteria} is difficult in practice. Instead we have the following lemma which justifies our stopping criteria for IP. 

\begin{lemma}
For a given input $x^{\textrm{obs}}$ if event $S^{\textrm{IP}}_L(x^{\textrm{obs}})$ (after asking $L$ queries) satisfies the condition specified by \eqref{eq: terminating criteria} then for all subsequent queries $q_m$, $m \geq L$, $\max_{q \in Q} I({q}(X); Y|S^{\textrm{IP}}_m(x^{\textrm{obs}})) = 0$.
\label{lemma: termination_criterion}
\end{lemma}

Refer to Appendix \ref{Appendix: proof of terminating criteria} for a proof.

Inspired from Lemma \ref{lemma: termination_criterion} we formulate an optimistic stopping criteria as,
\begin{equation}
L = \inf\{k \in \{1, 2, ..., |Q|\}: \max_{q \in Q} I({q}(X); Y|S^{\textrm{IP}}_m(x^{\textrm{obs}})) = 0 \ \forall m \geq k, m \leq |Q|\} 
\label{eq: terminating condition based on lemma exact}
\end{equation}
Evaluating \eqref{eq: terminating condition based on lemma exact} would be computationally costly since it would involve processing all the queries for every input $x$. We employ a more practically amenable criteria 
\begin{equation}
q_{L+1} = q_{STOP} \quad \textrm{ if } \quad \max_{q \in Q} I({q}(X); Y|S^{\textrm{IP}}_m(x^{\textrm{obs}})) = 0 \; \forall m \in \{L, L + 1, ..., L + T\}
\label{eq: terminating condition based on lemma exact practical}
\end{equation}
$T > 0$ is a hyper-parameter chosen via cross-validation. 
Note, it is possible that there does not exist any informative query in one iteration, but upon choosing a question there suddenly appears informative queries in the next iteration. For example, consider the XOR problem. $X \in \sR^2$ and $Y \in \{0, 1\}$. Let $Q$ be the set to two axis-aligned half-spaces. Both half-spaces have zero mutual information with $Y$. However, upon choosing any one as $q_1$, the other half-space is suddenly informative about $Y$. Equation \eqref{eq: terminating condition based on lemma exact practical} ensures that we do not stop prematurely.\\

\myparagraph{Termination Criteria for general $\epsilon$ when $d$ is taken as the KL-divergence}
For a general $\epsilon > 0$ we would like IP to terminate such that on average, 
\begin{equation}
    \mathbb{E}_X[KL\left(p(Y \mid X), p(Y \mid expl^{\textrm{IP}}_Q(X)\right)] \leq \epsilon
    \label{eq: random stuff}
\end{equation}

Detecting this is difficult in practice since IP is an online algorithm and only computes query-answers for a given input $x$. So it is not possible to know apriori when to terminate such that in expectation the KL divergence would be less than $\epsilon$. Instead we opt for the stronger requirement that,
\begin{equation}
    KL\left(p(Y \mid x), p(Y \mid expl^{\textrm{IP}}_Q(x)\right) \leq \epsilon \quad \forall x \in \gX.
    \label{eq: stronger_requirement_general_epsilon}
\end{equation}
It is easy to see that \eqref{eq: stronger_requirement_general_epsilon} implies \eqref{eq: random stuff}.

As before, $p(y \mid expl^{\textrm{IP}}_Q(x)) := p(y \mid S^{\textrm{IP}}_{t^{\textrm{IP}}(x^{\textrm{obs}})}(x^{\textrm{obs}})) \;  \forall (x, y) \in (\gX \times \gY)$ and $t^{\textrm{IP}}(x)$ is the number of iterations IP takes on input $x$ before termination. We have the following lemma (analogous to the $\epsilon = 0$ case). 

\begin{lemma} We make the following assumptions:
\begin{enumerate}
    \item $\gY$ is a countable set (recall $Y \in \gY$).
    \item For any $x^{\textrm{obs}} \in \gX$ and $x_1, x_2 \in S^{\textrm{IP}}_{t^{\textrm{IP}}(x^{\textrm{obs}})}(x^{\textrm{obs}})$, we have $p(Y \mid x_1)$ and $p(Y \mid x_2)$ have the same support.
\end{enumerate}
Then, for given input $x^{\textrm{obs}}$ if event $S^{\textrm{IP}}_{t^{\textrm{IP}}(x^{\textrm{obs}})}(x^{\textrm{obs}})$ (after asking $t^{\textrm{IP}}(x^{\textrm{obs}})$ queries) satisfies the condition specified by \eqref{eq: stronger_requirement_general_epsilon} then for all subsequent queries $q_m$, $m \geq t^{\textrm{IP}}(x^{\textrm{obs}})$, $\max_{q \in Q} I({q}(X); Y|S^{\textrm{IP}}_m(x^{\textrm{obs}})) \leq \epsilon'$, where $\epsilon' =  C\epsilon$ for some constant $C > 0$. 
\label{lemma: termination_criterion general}
\end{lemma}
Refer to Appendix \ref{Appendix: proof of terminating criteria general} for a proof. The assumption of $\gY$ being countable is typical for supervised learning (the scenario considered in this paper) where the set of labels is often finite. The second assumption intuitively means that $P(y \mid x_2) = 0 \implies P(y \mid x_1) = 0$ for any $y \in \gY$. This is a reasonable assumption since we envision practical scenarios in which $\epsilon$ is close to $0$ and thus different inputs which share the same query-answers until termination by $\textrm{IP}$ are expected to have very ``similar" posteriors.\footnote{We refer to the distribution $p(y \mid x)$ for any $x \in \gX$ as the posterior distribution of $x$.}

Inspired from Lemma \ref{lemma: termination_criterion general} we formulate an optimistic stopping criteria $\forall x^{\textrm{obs}} \in \gX$ as,
\begin{equation}
t^{\textrm{IP}}(x^{\textrm{obs}}) = \inf\{k \in \{1, 2, ..., |Q|\}: \max_{q \in Q} I({q}(X); Y|S^{\textrm{IP}}_m(x^{\textrm{obs}})) \leq \epsilon' \ \forall m \geq k, m \leq |Q|\} 
\label{eq: terminating condition based on lemma}
\end{equation}
Evaluating \eqref{eq: terminating condition based on lemma} would be computationally costly since it would involve processing all the queries for every input $x^{\textrm{obs}}$. We employ a more practically amenable criteria 
\begin{equation}
q_{t^{\textrm{IP}}(x^{\textrm{obs}}) + 1} = q_{STOP} \quad \textrm{ if } \quad \max_{q \in Q} I({q}(X); Y|S^{\textrm{IP}}_m(x^{\textrm{obs}})) \leq \epsilon' \; \forall m \in \{L, L + 1, ..., L + T\}
\end{equation}
$T > 0$ is a hyper-parameter chosen via cross-validation.

\subsection{Proof of Lemma \ref{lemma: termination_criterion}}
\label{Appendix: proof of terminating criteria}

\begin{proof}
Recall each query $q$ partitions the set $\gX$ and $S^{\textrm{IP}}_L(x^{\textrm{obs}})$ is the event $\{x' \in \gX \mid \{q_i, {q_i}(x^{\textrm{obs}})\}_{1:L} = \{q_i, {q_i}(x')\}_{1:L}\}$. It is easy to see that if $S^{\textrm{IP}}_L(x)$ satisfies the condition specified by \eqref{eq: terminating criteria} then 
\begin{equation}
    P(y  \mid S^{\textrm{IP}}_m(x^{\textrm{obs}})) = P(y \mid x') \ \forall x' \in S^{\textrm{IP}}_m(x^{\textrm{obs}}) \ \forall m \geq L, \ \forall q \in Q  
\label{eq: equivalence class partitioning}
\end{equation}
This is because subsequent query-answers partition a set in which all the data points have the same posterior distributions. Now, $\forall q \in Q, \ \forall a \in Range(q), \ y \in \gY$

\begin{equation}
\begin{aligned}
    p({q}(X) = a, y|S^{\textrm{IP}}_m(x^{\textrm{obs}})) = p({q}(X) = a \mid S^{\textrm{IP}}_m(x^{\textrm{obs}})) p(y \mid {q}(X) = a, S^{\textrm{IP}}_m(x^{\textrm{obs}}))  
\end{aligned}
\label{eq: chain rule of prob}
\end{equation}
\eqref{eq: chain rule of prob} is just an application of the chain rule of probability. The randomness in $q(X)$ is entirely due to the randomness in $X$. For any $a \in Range(q)$, $y \in \gY$
\begin{equation}
\begin{aligned}
    p(y \mid {q}(X) = a, S^{\textrm{IP}}_m(x^{\textrm{obs}})) &= \sum_{x'} p(y, X = x' \mid a, S^{\textrm{IP}}_m(x^{\textrm{obs}})) \\
    &= \sum_{x'} p(y \mid X = x', a, S^{\textrm{IP}}_m(x^{\textrm{obs}}))p(X = x' \mid a, S^{\textrm{IP}}_m(x^{\textrm{obs}})) \\
    &= \sum_{x'} p(y \mid X = x')p(X = x' \mid a, S^{\textrm{IP}}_m(x^{\textrm{obs}})) \\
    &= p(y \mid S^{\textrm{IP}}_m(x^{\textrm{obs}}))\sum_{x'} p(X = x' \mid a, S^{\textrm{IP}}_m(x^{\textrm{obs}})) \\
    &= p(y \mid S^{\textrm{IP}}_m(x^{\textrm{obs}}))
\end{aligned}
\label{eq: y indep a given sufficient history}
\end{equation}

The first equality is an application of the law of total probability, third due to conditional independence of the history and the hypothesis given $X=x'$ (assumption) and the fourth by invoking (\eqref{eq: equivalence class partitioning}).

Substituting \eqref{eq: y indep a given sufficient history} in \eqref{eq: chain rule of prob} we obtain $Y \indep q(X) \mid S^{\textrm{IP}}_m(x^{\textrm{obs}}))$ $\forall m \geq L, q \in Q$. This implies that for  all subsequent queries $q_m$, $m > L$, $\max_{q \in Q} I({q}(X); Y|S^{\textrm{IP}}_m(x^{\textrm{obs}}))) = 0$. Hence, Proved.

\end{proof}

\subsection{Proof of Lemma \ref{lemma: termination_criterion general}}
\label{Appendix: proof of terminating criteria general}

\begin{proof}
\myparagraph{Condition \eqref{eq: stronger_requirement_general_epsilon} implies bounded KL divergence between inputs on which IP has identical query-answer trajectories} Recall $S^{\textrm{IP}}_{t^{\textrm{IP}}(x^{\textrm{obs}}))}(x^{\textrm{obs}}))$ is the event $\{x' \in \gX \mid \{q_i, {q_i}(x^{\textrm{obs}}))\}_{1:t^{\textrm{IP}}(x^{\textrm{obs}}))} = \{q_i, {q_i}(x')\}_{1:t^{\textrm{IP}}(x^{\textrm{obs}}))}\}$. If $S^{\textrm{IP}}_{t^{\textrm{IP}}(x^{\textrm{obs}}))}(x^{\textrm{obs}}))$ satisfies \eqref{eq: stronger_requirement_general_epsilon} then using Pinsker's inequality we conclude, 
\begin{equation}
    \delta(p(Y \mid x^{\textrm{obs}})), p(Y \mid S^{\textrm{IP}}_{t^{\textrm{IP}}(x^{\textrm{obs}}))}(x^{\textrm{obs}}))) \leq \sqrt{\frac{\epsilon}{2}}
    \label{eq: pinsker 1}
\end{equation}
Here $\delta$ is the total variational distance between the two distributions. Since $\delta$ is a metric we conclude for any $x_1, x_2 \in S^{\textrm{IP}}_{t^{\textrm{IP}}(x^{\textrm{obs}}))}(x^{\textrm{obs}}))$,
\begin{equation}
\begin{aligned}
\delta(p(Y \mid x_1), p(Y \mid x_2)) &\leq  \delta(p(Y \mid x_1), p(Y \mid S^{\textrm{IP}}_{t^{\textrm{IP}}(x^{\textrm{obs}})}(x^{\textrm{obs}})) + \delta(p(Y \mid x_2), p(Y \mid S^{\textrm{IP}}_{t^{\textrm{IP}}(x^{\textrm{obs}})}(x^{\textrm{obs}}))  \\
&\leq \sqrt{2\epsilon}
\end{aligned}
\label{eq: metric tv}
\end{equation}

Since $\gY$ is countable, define $\eta = \min\{p(y \mid \hat{x}): y \in \gY, p(y \mid \hat{x})) > 0, \hat{x} \in \gX\}$. Then, by the reverse Pinsker's inequality we conclude,
\begin{equation}
   KL(p(Y \mid x_1), p(Y \mid x_2))) \leq \frac{\epsilon}{\eta} =: \epsilon' \quad \forall x_1, x_2 \in S^{\textrm{IP}}_{t^{\textrm{IP}}(x^{\textrm{obs}})}(x^{\textrm{obs}})
   \label{eq: revere pinsker}
\end{equation}
Note, the above upper bound holds since by assumption 2, $p(Y \mid x_1)$ and $p(Y \mid x_2))$ have the same support.

\myparagraph{Bounded KL divergence between inputs implies subsequent queries have mutual information bounded by $\boldsymbol{\epsilon}$}
For any subsequent query $q \in Q$ that IP asks about input $x^{\textrm{obs}}$ we have $\forall x \in S^{\textrm{IP}}_{t^{\textrm{IP}}(x^{\textrm{obs}})}(x^{\textrm{obs}})$,
\begin{equation}
  KL\left(p(Y \mid x), p(Y \mid S^{\textrm{IP}}_{t^{\textrm{IP}}(x^{\textrm{obs}}) + 1}(x^{\textrm{obs}}))\right) = \sum_Y  p(Y \mid x) \log  p(Y \mid x)  - \sum_Y  p(Y \mid x) \log p(Y \mid S^{\textrm{IP}}_{t^{\textrm{IP}}(x^{\textrm{obs}}) + 1}(x^{\textrm{obs}}))
  \label{eq. KL espression for statistic}
\end{equation}
where, $S^{\textrm{IP}}_{t^{\textrm{IP}}(x^{\textrm{obs}}) + 1}(x^{\textrm{obs}}) := S^{\textrm{IP}}_{t^{\textrm{IP}}(x^{\textrm{obs}})}(x^{\textrm{obs}}) \cap \{x' \in \gX: q(x') = q(x)\}$. For brevity, we denote $S^{\textrm{IP}}_{t^{\textrm{IP}}(x^{\textrm{obs}}) + 1}(x^{\textrm{obs}})$ as $\textrm{B}$,

\begin{equation}
\begin{aligned}
  \sum_Y  p(Y \mid x) \log p(Y \mid \textrm{B}) &= \sum_Y  p(Y \mid x) \log \left[\sum_{x' \in \textrm{B}} p(Y \mid x', \textrm{B})P(x' \mid \textrm{B}) \right]  \\
  &= \sum_Y  p(Y \mid x) \log \left[\sum_{x' \in \textrm{B}} p(Y \mid x')P(x' \mid \textrm{B})  \right] \\
  &\geq \sum_Y  p(Y \mid x) \sum_{x' \in \textrm{B}} p(x' \mid \textrm{B}) \log p(Y \mid x') \\
  &= \sum_{x' \in \textrm{B}} p(x' \mid \textrm{B}) \sum_Y  p(Y \mid x) \log  p(Y \mid x')
  \label{eq. second term of KL espression for statistic}
\end{aligned}
\end{equation} 
In the third inequality Jensen's inequality was used. 
Substituting \eqref{eq. second term of KL espression for statistic} in \eqref{eq. KL espression for statistic},

\begin{equation}
\begin{aligned}
  KL\left(p(Y \mid x) || p(Y \mid \textrm{B})\right) &\leq \sum_{x' \in \textrm{B}} p(x' \mid \textrm{B}) \sum_Y  p(Y \mid x) \log  \frac{p(Y \mid x)}{ p(Y \mid x')} \\ 
  &\leq  \epsilon' \sum_{x'} p(x' \mid \textrm{B}) \\
  &= \epsilon'
  \label{eq: bounding KL divergence}
\end{aligned}   
\end{equation}
In the second inequality we substituted from \eqref{eq: revere pinsker} since $x, x' \in \textrm{B} \subseteq S^{\textrm{IP}}_{t^{\textrm{IP}}(x^{\textrm{obs}})}(x^{\textrm{obs}})$. In the third equality we used the identity $\sum_{x' \in \textrm{B}} p(x' \mid \textrm{B}) = 1$,

It is easy to see that \eqref{eq: bounding KL divergence} holds for all $x' \in S^{\textrm{IP}}_{t^{\textrm{IP}}(x^{\textrm{obs}}) + 1}(x^{\textrm{obs}})$ and thus $I(X; Y \mid S^{\textrm{IP}}_{t^{\textrm{IP}}(x^{\textrm{obs}}) + 1}(x^{\textrm{obs}})) \leq \epsilon$

Since $Y \rightarrow X \rightarrow q(X)$ we can apply the data-processing inequality to obtain,
\[I(q(X); Y \mid S^{\textrm{IP}}_{t^{\textrm{IP}}(x^{\textrm{obs}}) + 1}(x^{\textrm{obs}})) \leq I(X; Y \mid S^{\textrm{IP}}_{t^{\textrm{IP}}(x^{\textrm{obs}}) + 1}(x^{\textrm{obs}})) \leq \epsilon' \quad \forall q \in Q.\]

This implies that for  all subsequent queries $q_m$, $m > t^{\textrm{IP}}(x^{\textrm{obs}})$, $\max_{q \in Q} I({q}(X); Y|S^{\textrm{IP}}_m(x^{\textrm{obs}})) \leq \epsilon$. Hence, Proved.
\end{proof}



\subsection{Complexity of the Information Pursuit Algorithm}
\label{Appendix: complexity analysis}
For any given input $x$, the per-iteration cost of the IP algorithm is $\mathcal{O}(N + |Q|m)$, where $|Q|$ is the total number of queries, $N$ is the number of ULA iterations, and $m$ is cardinality of the product sample space $q(X) \times Y$. For simplicity we assume that the output hypothesis $Y$ and query-answers $q(X)$ are finite-valued and also that the number of values query answers can take is the same but our framework can handle more general cases.

More specifically, to compute $q_{k+1} = \argmax_{q \in Q} I({q}(X); Y \mid S^{\textrm{IP}}_{k}(x^{\textrm{obs}}))$. We first run ULA for $N$ iterations to get samples from $p(z \mid y, S^{\textrm{IP}}_k(x))$ which are then used to estimate the distribution $p(q(x), y \mid S^{\textrm{IP}}_k(x))$ (using \eqref{eq: compute intermediate conditionals}) for every query $q \in Q$ and every possible query-answer hypothesis, $(q(x), y)$, pair. This incurs a cost of $\mathcal{O}(N + |Q|m)$. We then numerically compute the mutual information between query-answer $q(X)$ and $Y$ given history for every $q \in Q$  as described in \eqref{eq: numerical MI}. This has a computational complexity $\mathcal{O}(|Q|m)$. Finally, we search over all $q \in Q$ to find the query with maximum mutual information (refer \eqref{def. IP algorithm}).

It is possible to reduce $N$ by using advanced MCMC sampling methods which converge faster to the required distribution $p(z \mid y, S^{\textrm{IP}}_k(x)))$. The linear cost of searching of all queries can also be reduced by making further assumptions about the structure of the query set. For example, we conjecture that this cost can be reduced to $\log |Q|$ using hierarchical query sets where answers to queries would depend upon to answers to queries higher up in the hierarchy. Note that if the query answers $q(X)$ and $Y$ are continuous random variables then we would need to resort to sampling to construct stochastic estimates of the mutual information between $q(X)$ and $Y$ instead of carrying our explicit numerical computations. We would explore these directions in future work.

\subsection{Network Architectures and Training Procedure}
\label{Appendix: Network architectures and training}
Here we describe the architectures and training procedures for the $\beta$-VAEs used to calculate mutual information as described in Section \ref{section: Information Pursuit}.

\subsubsection{Architectures}
\begin{figure}[H]
    \centering
    \includegraphics[scale=0.4]{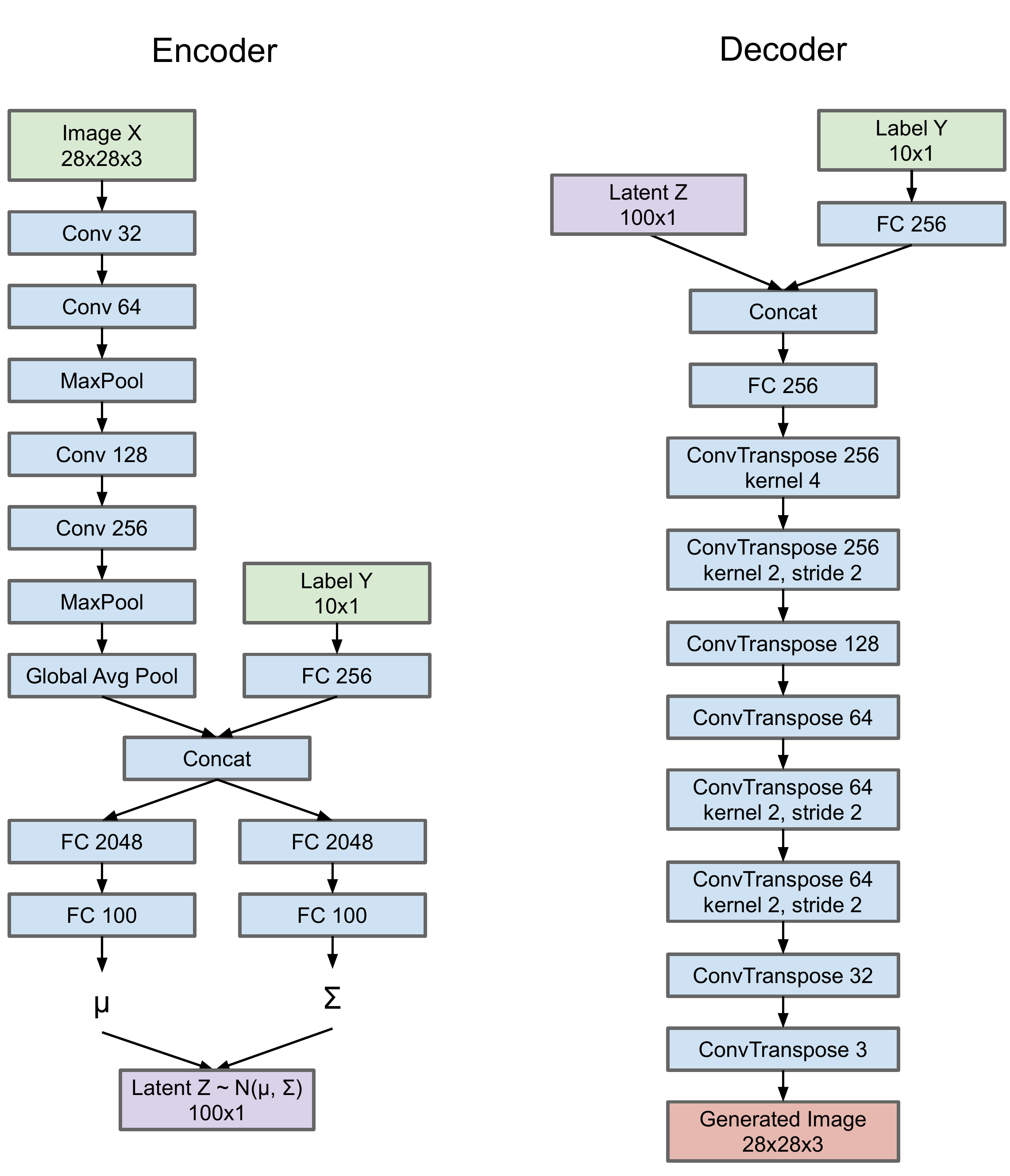}
    \caption{\textit{Binary Image VAE}}
\end{figure}
\myparagraph{Binary Image Classification} In the encoder, all convolutional layers use kernel size 3 and stride 1, and max pool layers use a pooling window of size 2. In the decoder all transposed convolutions use kernel size 3 and stride 1 unless otherwise noted. In both the encoder and decoder, all non-pooling layers are followed by a BatchNorm layer and LeakyReLU activation (with slope -0.3) except for the final encoder layer (no nonlinearities) and the final decoder layer (sigmoid activation).

\begin{figure}[H]
    \centering
    \begin{tabular}{c@{\hskip 1cm}c}
        \includegraphics[width=0.45\textwidth]{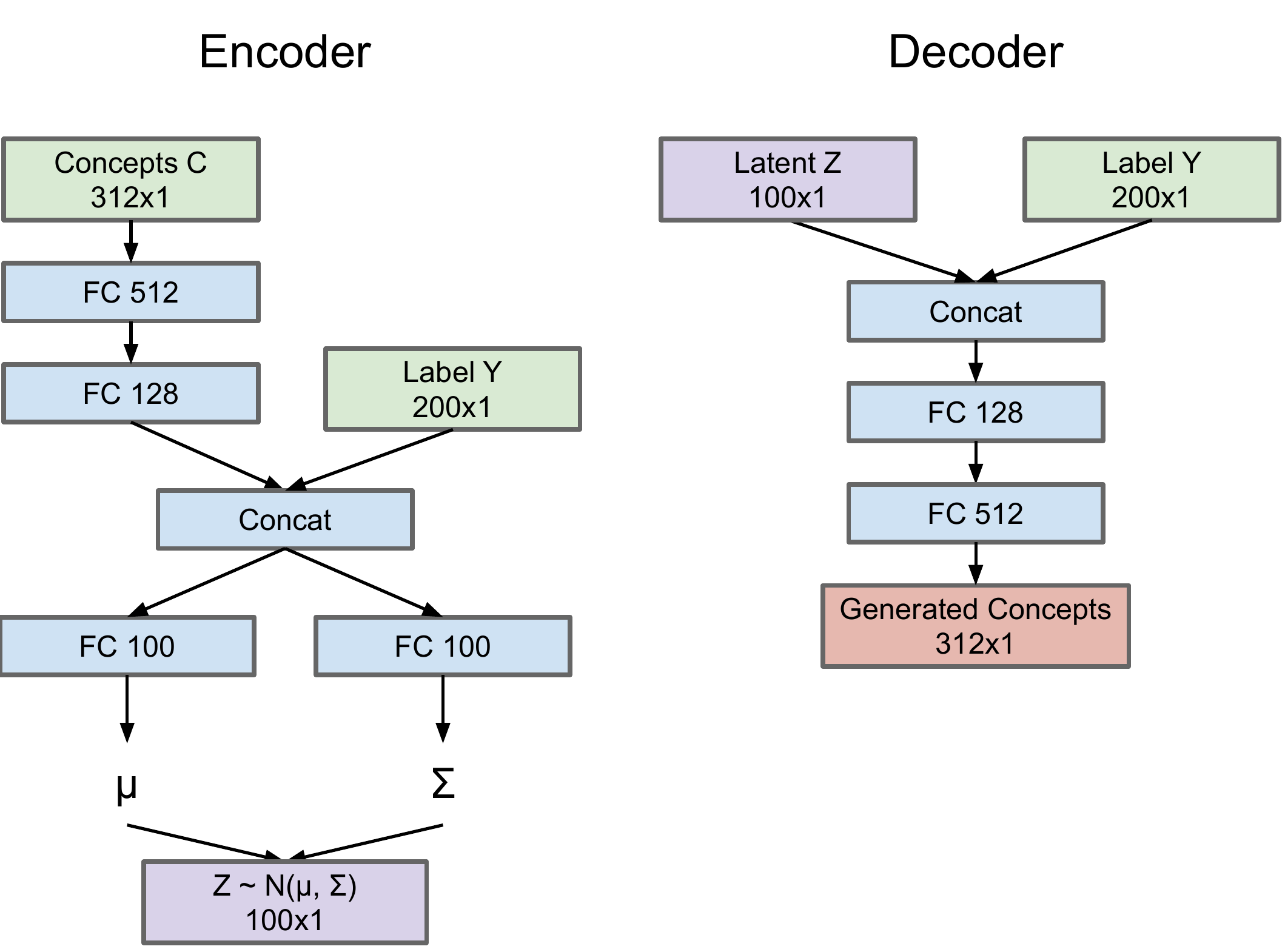} & \includegraphics[width=0.45\textwidth]{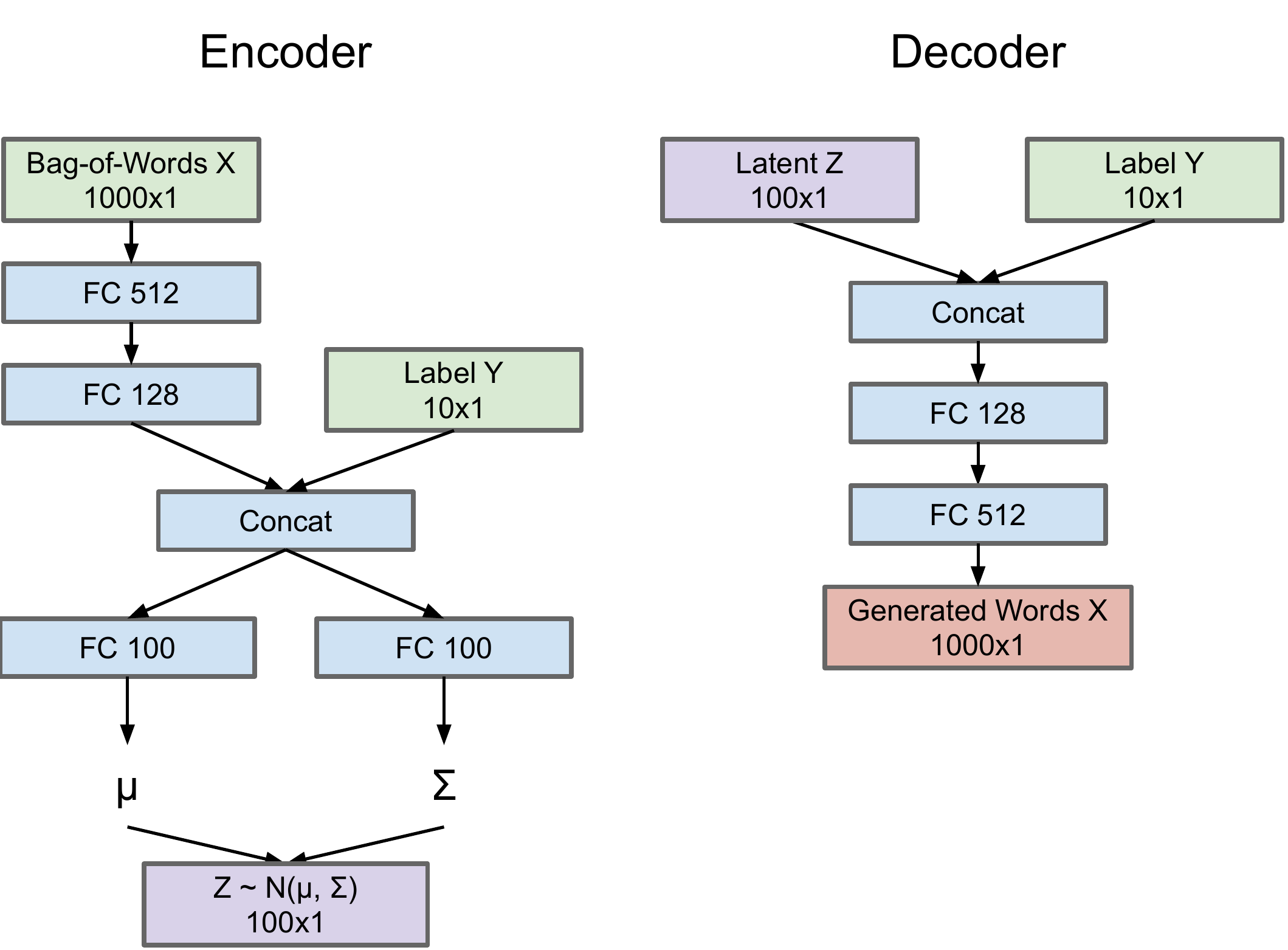} \\
        \small \textit{(a) CUB Concepts VAE} & \small \textit{(b) HuffPost News Headlines VAE}
    \end{tabular}
    \caption{\textit{CUB Concepts and HuffPost News Headlines VAEs}}
\end{figure}
\myparagraph{CUB Bird Species Classification and HuffPost News Headline Classification} For CUB and HuffPost News, we use essentially the same VAE architecture, only designed to handle different-sized inputs $X$ and one-hot labels $Y$. All layers are followed by a BatchNorm layer and ReLU activation except for the final encoder layer (no nonlinearities) and the final decoder layer (sigmoid activation).

\subsubsection{Training}
The $\beta$-VAE was trained by optimizing the Evidence Lower BOund (ELBO) objective
\begin{equation}
\label{eq:ELBO} 
\begin{aligned}
    \max_{\omega, \phi} \text{ELBO}(\omega, \phi) = \sum_{i=1}^n \left[ \mathbb{E}_{p_{\phi}(z \mid y^{(i)}, x^{(i)})}[\log p_{\omega}(x^{(i)} \mid z, y^{(i)})] - \beta D_{KL}(p_{\phi}(z \mid y^{(i)},
x^{(i)}) \| p(z)) \right]
\end{aligned}
\end{equation} 
where $p_{\phi}(z \mid y^{(i)}, x^{(i)})$ denotes the encoder and $p_{\omega}(x^{(i)} \mid z, y^{(i)})$ the decoder. The prior over latents $p(z)$ is taken to be standard Gaussian.

For all binary image datasets, we trained our VAE for 200 epochs using Adam with learning rate $0.001$ and the $\beta$-VAE parameter $\beta = 5.0$. We also trained the CUB VAE for 200 epochs using Adam with learning rate $0.001$ but with $\beta = 1.0$. Finally, we trained the HuffPost News VAE for 100 epochs using Adam with learning rate $0.0002$ and $\beta = 1.0$.

\section{}
\subsection{HuffPost News Task and Query Set} \label{Appendix: Tasks and Query Sets}

The Huffington Post News Category dataset consists of 200,853 articles published by the Huffington Post between 2012 and 2018. Each datapoint contains the article ``headline'', a ``short description" (a one to two-sentence-long continuation of the headline), and a label for the category/section it was published under in the newspaper. We concatenate the article headline and short description to form one extended headline. Additionally, many of the 41 category labels are redundant (due to changes in how newspaper sections were named over the years), are semantically ambiguous and HuffPost-specific (e.g. ``Impact'', ``Fifty'', ``Worldpost''), or have very few articles. Therefore, we combine category labels for sections with equivalent names (e.g. ``Arts \& Culture'' with ``Culture \& Arts''), remove ambiguous HuffPost-specific categories, and then keep only the 10 most frequent categories to ensure that each category has an adequate number of samples. This leaves us with a final dataset of size 132,508 and category labels ``Entertainment'', ``Politics'', ``Queer Voices'', ``Business'', ``Travel'', ``Parenting'', ``Style \& Beauty'', ``Food \& Drink'', ``Home \& Living'', and ``Wellness''.

\subsection{Comparison Models} \label{Appendix: Comparison Models}
\subsubsection{MAP using $Q$} For IP, we use ULA to sample from $p(z \mid y, S^{\textrm{IP}}_k(x))$, but now that we have access to the all the query answers, $Q(x)$, we can improve performance by making use of the VAE's encoder instead. Following equation \ref{eq: marginalize z}, we draw many samples from the encoder $p(z \mid Q(x), y)$ and then decode these samples to estimate the VAE's posterior distribution $p(y | Q(x))$. For each problem, we set the number of samples to be the same as what we draw during each iteration of IP (12,000 for binary image tasks, 12,000 for CUB, 10,000 for HuffPost News). We expect the accuracy of this model to serve as an upper bound for what we can achieve with IP given that it uses all queries. Our task-specific VAE architectures can be found in Appendix \ref{Appendix: Network architectures and training}.

\subsubsection{Black-Box Using $Q$} We also compare to non-interpretable supervised models which receive all queries $Q(x)$ as input and try to predict the associated label $y$. This allows a comparison between the accuracy of the posterior of our generative model and traditional supervised approaches on the chosen interpretable query set.

For the binary image datasets, we use a simple CNN where all convolutional layers use kernel size $3 \times 3$ and a stride of 1, all max pooling uses a kernel size of 2. For CUB and HuffPost News we use simple MLPs.
\begin{table}[H]
\centering
\begingroup
\renewcommand{\arraystretch}{1.5} 

\caption{\textit{Binary Image CNN Architecture}}
\begin{tabular}{| l | l | l | l |}
\hline
Layer & Input Size/Channels & Output Size/Channels & Nonlinearity \\
\hline
Convolution & 3 & 32 & BatchNorm + ReLU \\
Convolution & 32 & 64 & BatchNorm + ReLU + MaxPool \\
Convolution & 64 & 128 & BatchNorm + ReLU \\
Convolution & 128 & 256 & BatchNorm + ReLU + MaxPool + Global Avg Pool \\
Fully-connected & 256 & 2048 & BatchNorm + ReLU \\
Fully-connected & 2048 & 10 & Sigmoid \\
\hline
\end{tabular}
\endgroup
\end{table}

\begin{table}[H]
\centering
\begingroup
\renewcommand{\arraystretch}{1.5} 

\caption{\textit{CUB Attributes MLP Architecture}}
\begin{tabular}{| l | l | l | l |}
\hline
Layer & Input Size/Channels & Output Size/Channels & Nonlinearity \\
\hline
Fully-connected & 312 & 100 & ReLU \\
Fully-connected & 100 & 25 & ReLU \\
Fully-connected & 25 & 200 & Sigmoid \\
\hline
\end{tabular}

\endgroup
\end{table}

\begin{table}[H]
\centering
\begingroup
\renewcommand{\arraystretch}{1.5} 

\caption{\textit{HuffPost Bag-of-Words MLP Architecture}}
\begin{tabular}{| l | l | l | l |}
\hline
Layer & Input Size/Channels & Output Size/Channels & Nonlinearity \\
\hline
Fully-connected & 1000 & 100 & ReLU \\
Fully-connected & 100 & 25 & ReLU \\
Fully-connected & 25 & 10 & Sigmoid \\
\hline
\end{tabular}

\endgroup
\end{table}

\subsubsection{Black-Box} \label{Appendix: Black-Box} Since we ourselves pre-processed the cleaned 10-class version of HuffPost News, there are no reported accuracies in the literature to compare IP with. Therefore, as a strong black-box baseline, we fine-tune a pre-trained Bert Large Uncased Transformer model \cite{devlin2018bert} with an additional dropout layer (with dropout probability 0.3) and randomly initialized fully-connected layer. Our implementation is publicly available at https://www.kaggle.com/code/stewyslocum/news-classification-using-bert.

\newpage
\subsection{Additional Example Runs} \label{Appendix: Additional Example Runs}
\subsubsection{IP with Various Patch Scales} \label{Appendix: IP Patch Scales}
For binary image classification, we also experimented with patch queries of sizes other than $3 \times 3$, from single pixel $1 \times 1$ queries up to $4 \times 4$ patches. 

\begin{figure}[H]
    \centering
    \includegraphics[width=0.7\textwidth]{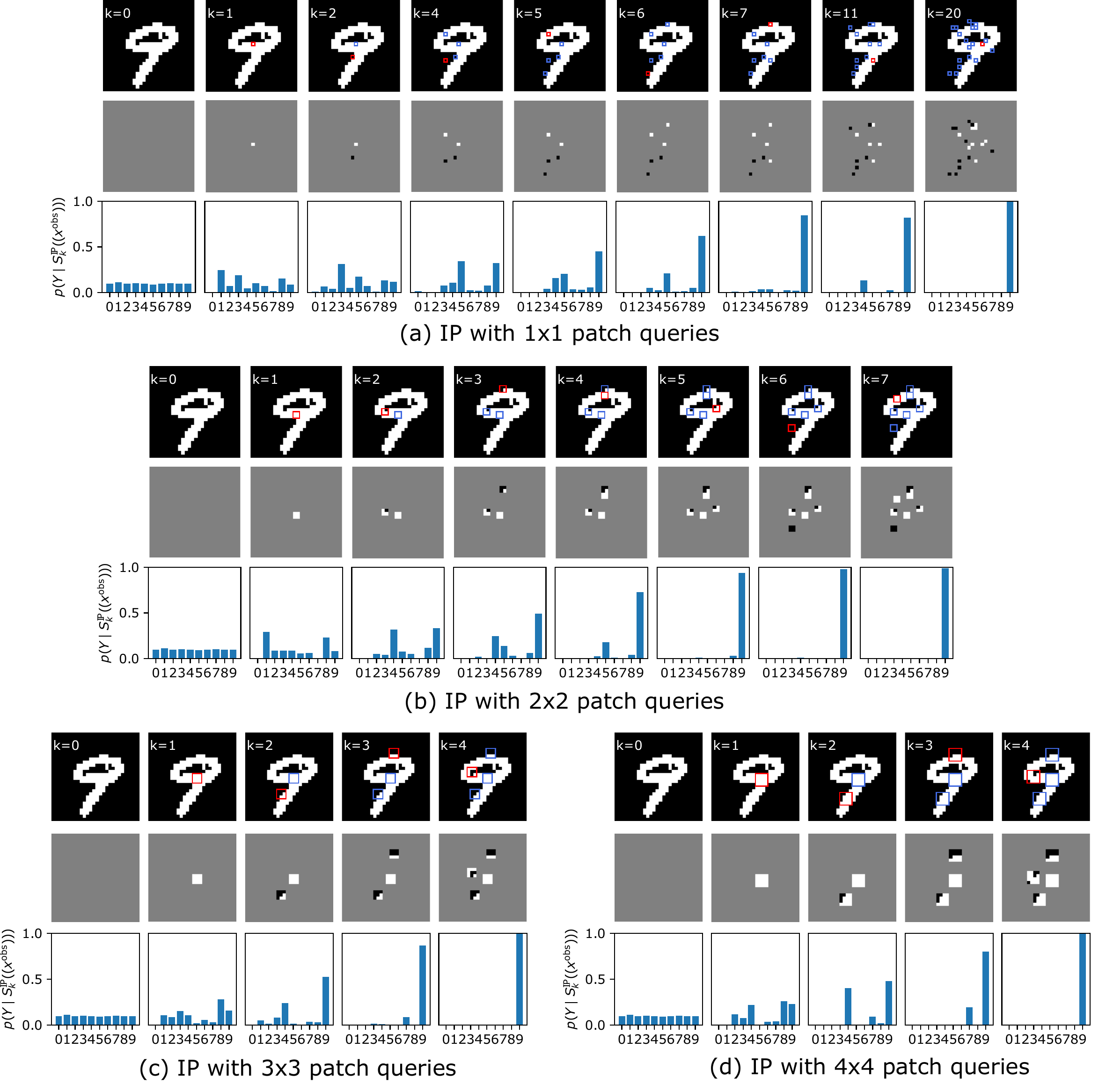}
    \caption{\textbf{IP on MNIST with different sized patch queries.} In each subfigure, the top row displays the test image with red boxes denoting the current queried patch and blue boxes denoting previous patches. The second row shows the revealed portion of the image that IP gets to use at each query. The final row shows the model's estimated posteriors at each query. For conciseness in (a), we only display the 8 iterations of IP with highest KL divergence between successive posteriors (i.e. the most influential iterations). Observe that at all patch sizes, the queries chosen by IP cover roughly the same parts of the image, illustrating the importance of this region during classification. Reaching the stopping criteria of $99\%$ posterior confidence takes 20 queries with $1\times1$ patches, 7 queries with $2\times2$ patches, 4 queries with $3\times3$ patches, and 4 queries with $4\times4$ patches.}
    \label{fig: effect of scale}
\end{figure}

\begin{table}[H]
\centering
\begingroup
\renewcommand{\arraystretch}{1.5} 

\caption{\textit{Number of queries and pixels gleaned by IP (until termination) using query sets of different patch scales on MNIST.}}
\begin{tabular}{| l | l | l | l | l |}
\hline
Patch Size & $1\times1$ & $2\times2$ & $3\times3$ & $4\times4$ \\
\hline
\# Queries & 21.1 & 9.6 & 5.2 & 4.6 \\
\# Pixels & 21.1 & 32.8 & 44.7 & 54.7 \\
\% Pixels & 2.7 & 4.2 & 5.7 & 6.9 \\
\hline
\end{tabular}

\endgroup
\end{table}

While $1\times1$ patch queries use the smallest total number of pixels and most similarly resemble existing post-hoc attribution maps, they lead to a large number of scattered queries that are hard to interpret individually. On the other extreme, as the patch size grows larger, the number of total queries decreases, but queries becomes harder to interpret since each patch would contain many image features. On the MNIST dataset, we found $3\times3$ patches to be a sweet spot where explanations tended to be very short, but were also at a level of granularity where each patch could be individually interpreted as a single edge or stroke. Remarkably, at all chosen patch scales, only a very small fraction (2-7\%) of the image needs to be revealed to classify the images with high confidence.

\newpage
\subsubsection{IP for Binary Image Classification}
Now we provide additional example runs of IP for each of the three binary image classification datasets.

As in Figure \ref{Fig: IP in action} in the main paper, in each plot, the top row displays the test image with red boxes denoting the current queried patch and blue boxes denoting previous patches. The second row shows the revealed portion of the image that IP gets to use at each query. The final row shows the model's estimated posteriors at each query.

\begin{figure}[H]
    \begin{center}\includegraphics[height=4.8cm]{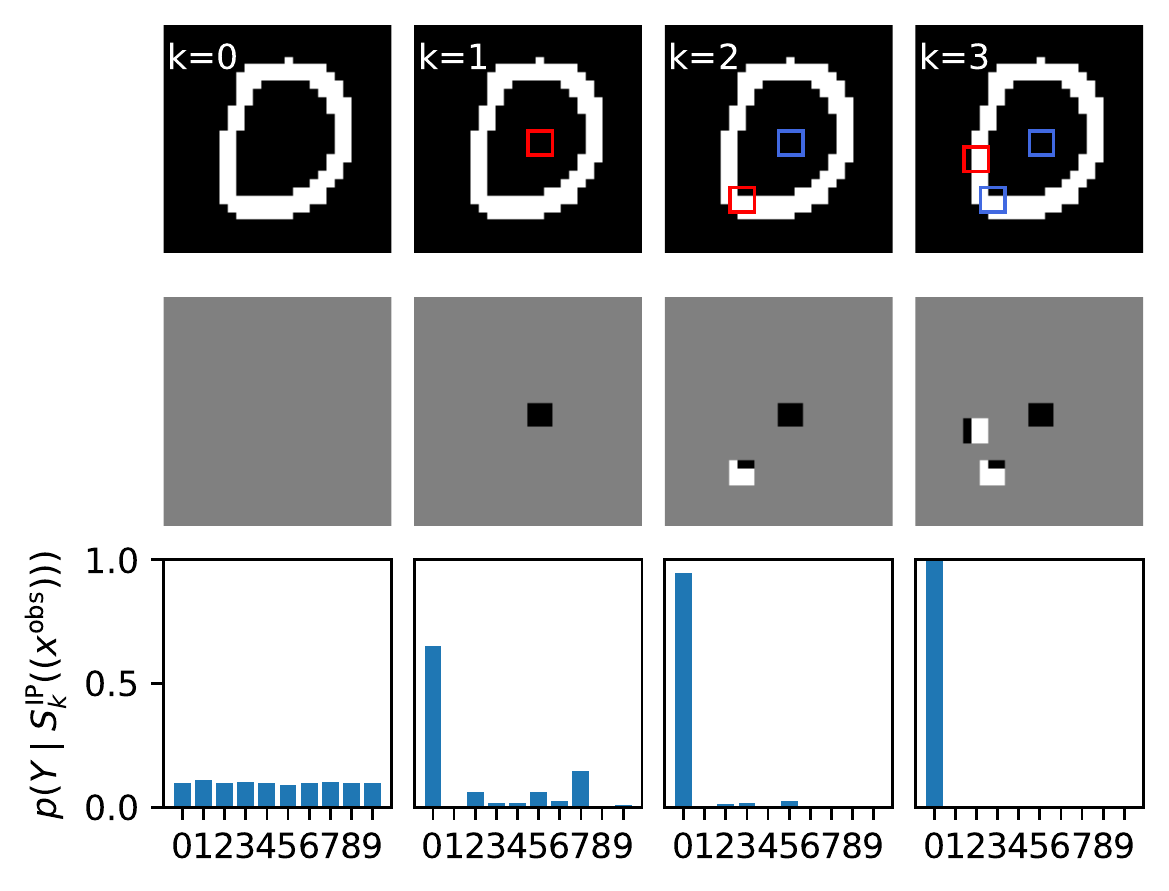}\end{center}
    \textit{(a) Each query reveals the patch with maximum mutual information with $Y$, conditioned on query history. This is initially independent of the particular image and asks for the pixel intensities in the center patch (see k = 1 in row 1). After the first query reveals that the center patch is all black, the posterior concentrates on ``0'' and ``7''. After observing a white corner in the bottom left (which would be black for a ``7''), the model becomes confident that the image is a ``0'', and even more so after one final query when it reaches termination.}
    
    \begin{center}\includegraphics[height=4.8cm]{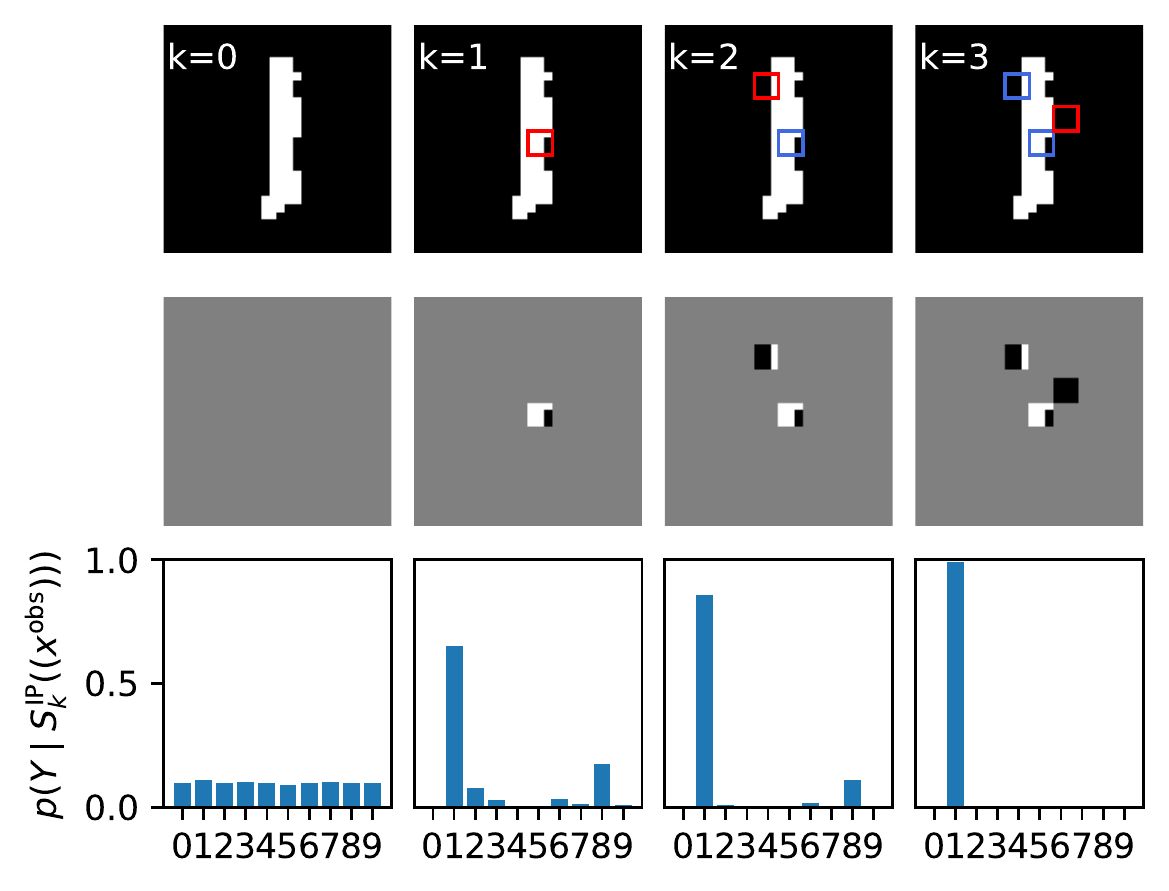}\end{center}
    \textit{(b) The first query reveals a vertical white stroke in the center of the image, leading to a concentration of the posterior on ``1''. In the next two queries, IP determines that there is a single long vertical stroke center stroke taking up the entire height of the image, and so it reaches a 99\% confidence of the image being a ``1''.}
    
    \begin{center}\includegraphics[height=4.8cm]{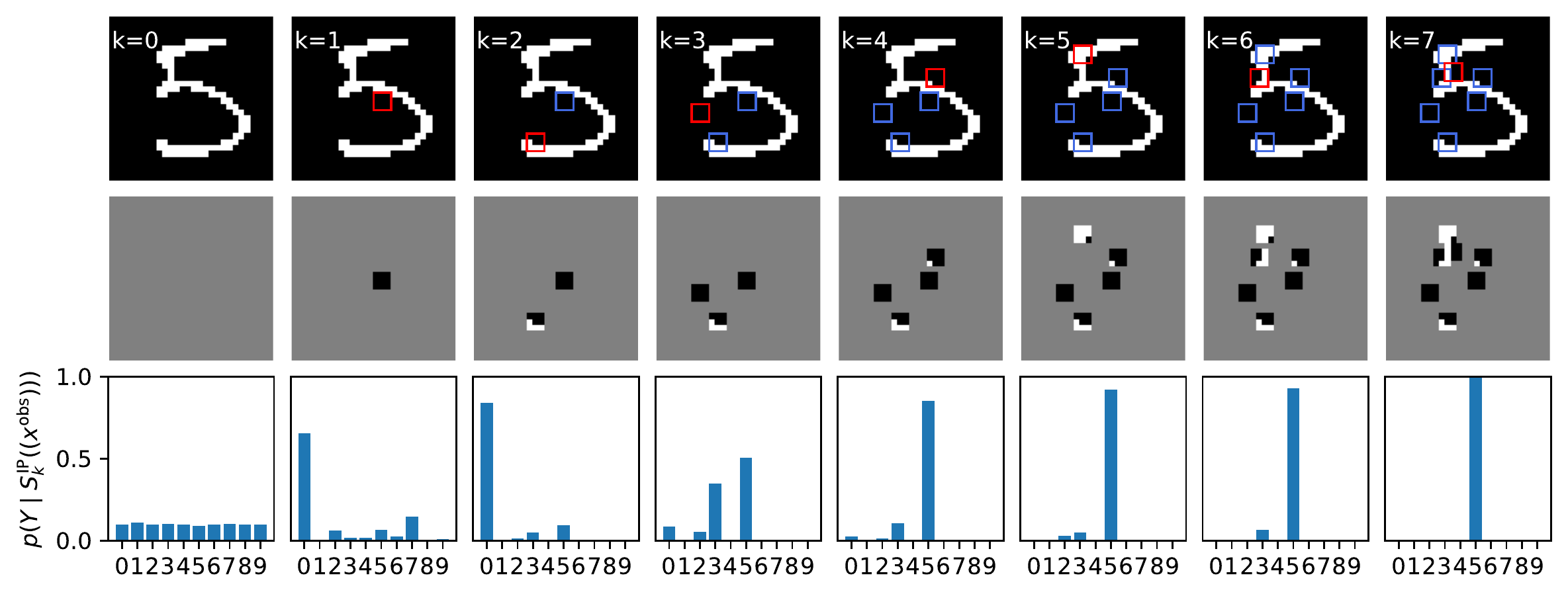}\end{center}
    \textit{(c) As in the top left example, the first query is all black, and the posterior mass shifts onto ``0''. Because the answer to this first query is identical as in the top left example, the second query chosen, in the bottom left of the image, is also the same. The response to this query is also a white corner as in the top left example, and so the posterior continues to concentrate on ``0'', and the third query is also in the same left area of the image. However, this third query reveals a black patch, indicating that the image might be a ``5'' instead of a ``0''. In the remaining four queries, IP discovers other portions of the ``5'' digit and finally arrives at the right answer with high confidence.}
    
    \caption{\textit{Additional Examples of IP on MNIST}}
\end{figure}
\newpage
Recall that in order to improve performance on the KMNIST and FashionMNIST datasets (at the expense of asking a few more queries) we modified IP's termination criteria to include a stability condition: terminate when the original criterion ($\max_Y p(Y | S^{\textrm{IP}}_k(x)) \geq 0.99$) is true for 5 queries in a row.

\begin{figure}[H]
    \begin{center}\includegraphics[width=\textwidth]{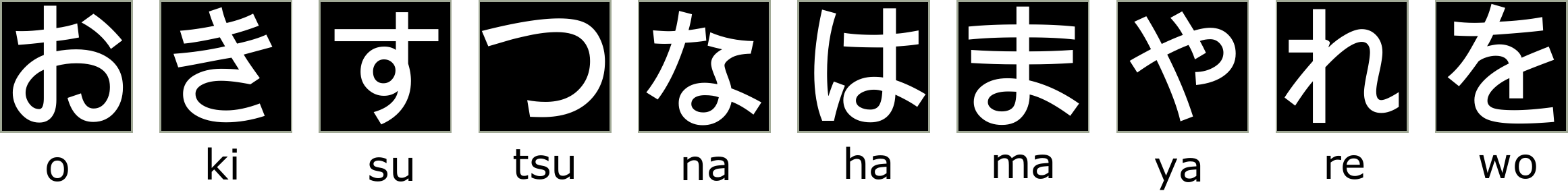}\end{center}
    \textit{(a) KMNIST is a 10-class dataset of handwritten Japanese Hiragana characters. To assist the reader in understanding the examples below, we display each typed character along with its romanized name.}

    \begin{center}\includegraphics[height=6cm]{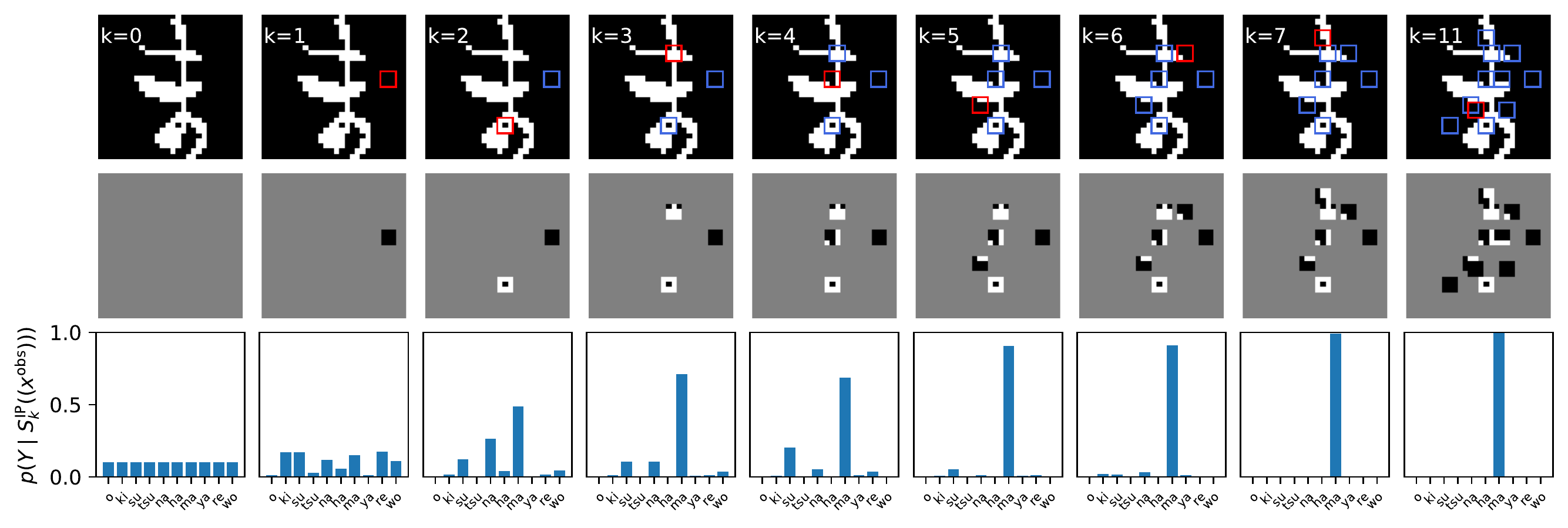}\end{center}
    \textit{(b) For each image, IP selects the first query to be in the middle right of the image, where several characters are likely to have a stroke. Upon finding none, IP rules out ``o'', ``tsu'', and ``ya'' but otherwise distributes probability mass rather equally. On the second query, IP discovers a closed loop in the bottom of the character, a clear sign of ``na'' and ``ma'', which increase the most. The discovery of the double crosses in the remaining queries concentrate the posterior on ``ma'' until termination. For conciseness, we only display the 8 iterations of IP with highest KL divergence between successive posteriors (i.e. the most influential iterations).}
    
    \begin{center}\includegraphics[height=6cm]{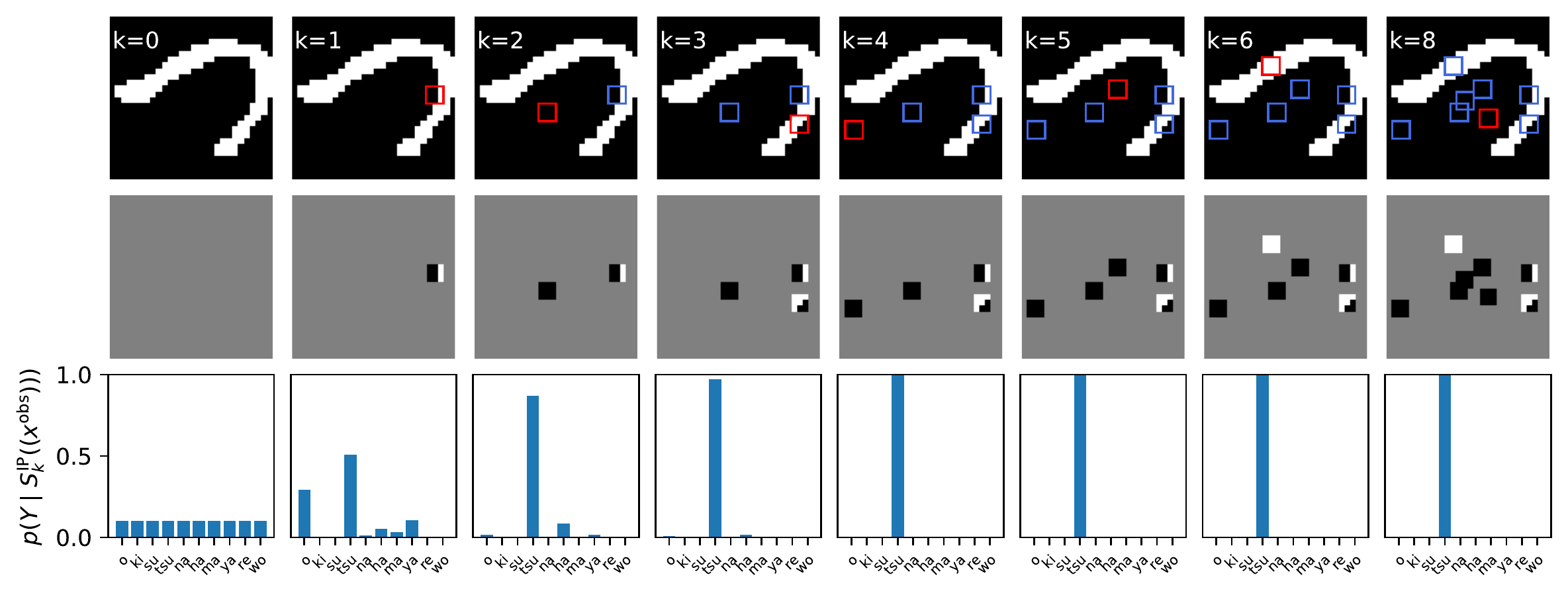}\end{center}
    \textit{(c) In the first query, IP discovers a left edge, hinting at the presence of a large loop on the right side of the image, features of ``o'' and ``tsu''. The second query is likely intended to disambiguate these two characters as ``o'' contains white strokes in this region. Upon finding a black patch here, IP is already confident, and reaches 99\% confidence in just four queries.}
    
    \caption{\textit{Additional Examples of IP on KMNIST}}
\end{figure}%
\newpage
\begin{figure}[ht]\ContinuedFloat
    \begin{center}\includegraphics[height=6cm]{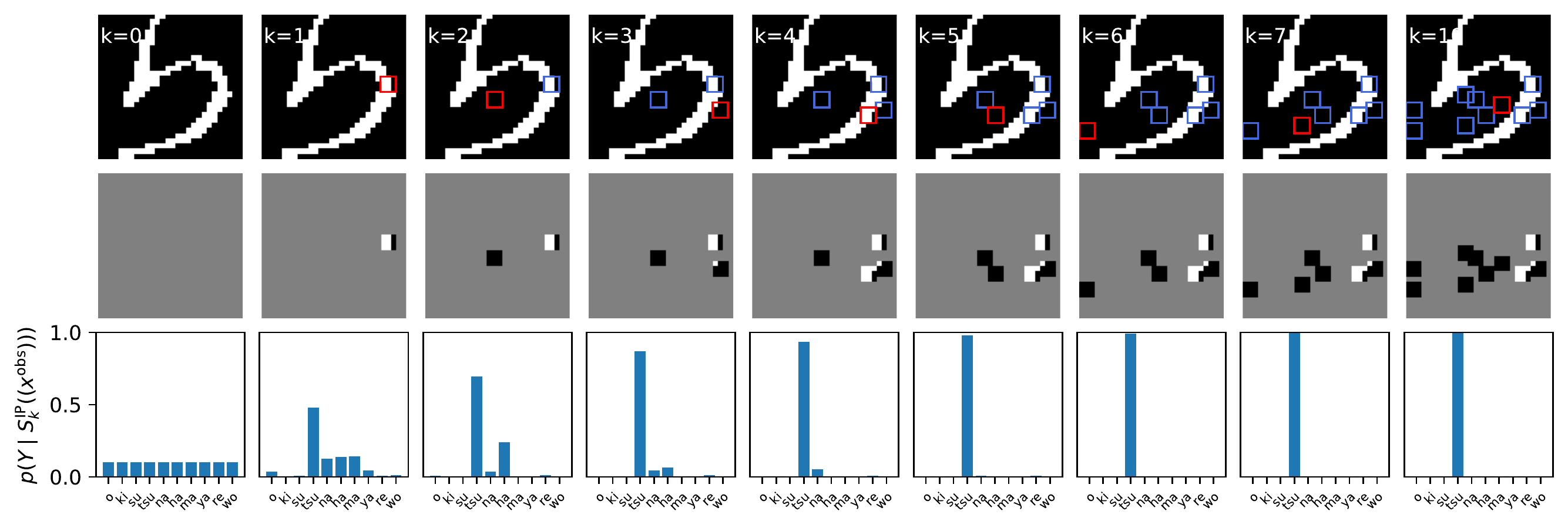}\end{center}
    \textit{(d) The first query reveals a right edge, suggesting a slightly smaller loop on the right side of the image, a feature shared by several characters. However, most of these queries have a busy center region except for ``tsu'', whose probability increases after the second query reveals a black patch. The next two queries outline the shape of the loop which is very large, a distinctive characteristic of ``tsu''.}
    
    \caption{\textit{Additional Examples of IP on KMNIST}}
\end{figure}
\newpage
\begin{figure}[H]
    \begin{center}\includegraphics[height=6cm]{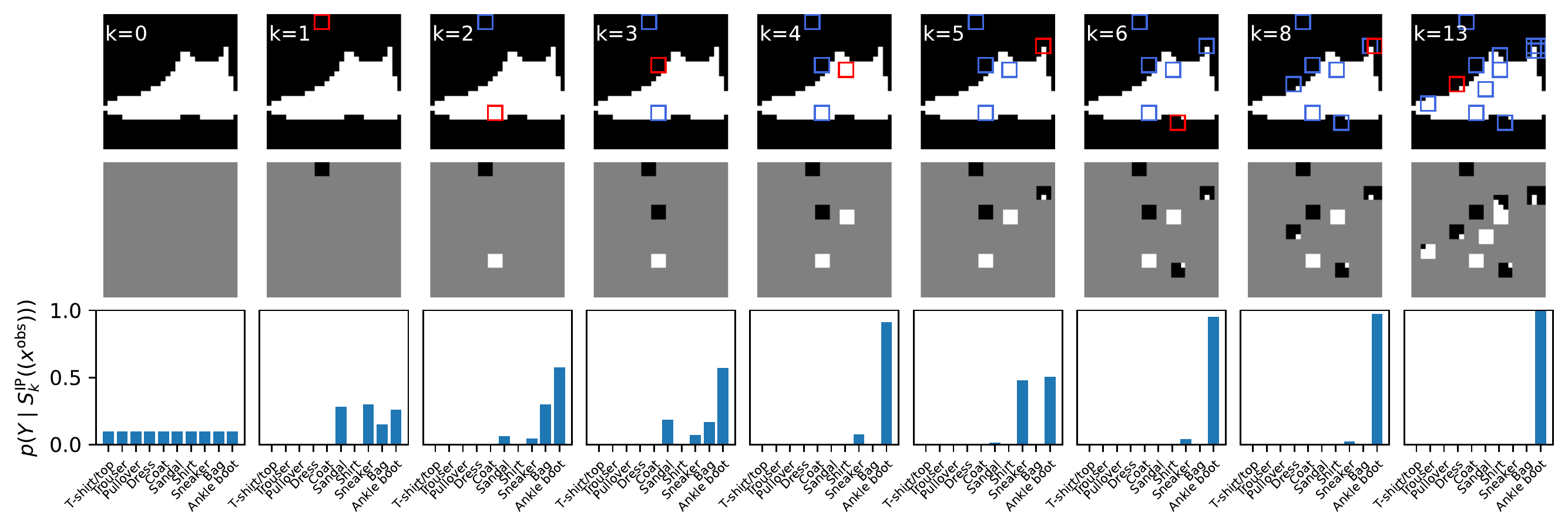}\end{center}
    \textit{(a) On this dataset, IP always selects the first query to be the patch in the top center, which being all black in this case rules out the possibility of the object being a type of pant or upper body garment, which would take up the entire height of the image. Over the next several queries, IP focuses on queries that would allow it to distinguish types of shoes from each other, in particular finding the shoe to have a high top and a small heel, eventually causing the posterior to concentrate on the correct ``Ankle Boot" category. For conciseness, we only display the 8 iterations of IP with highest KL divergence between successive posteriors (i.e. the most influential iterations).}
    
    \begin{center}\includegraphics[height=6cm]{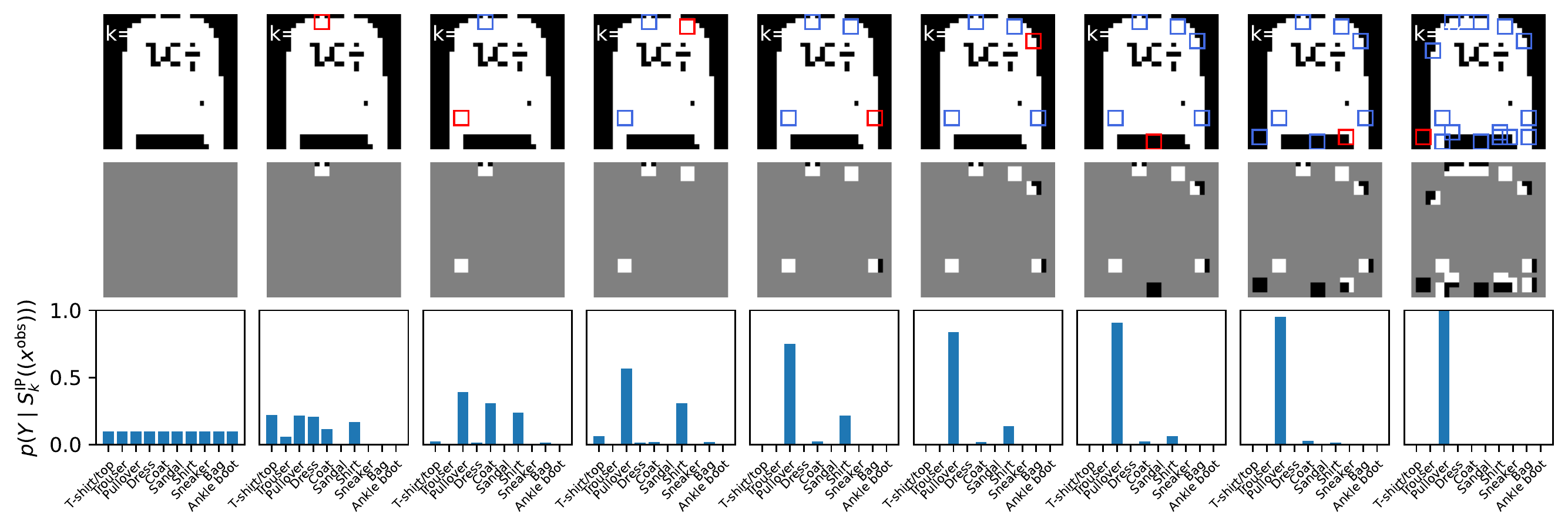}\end{center}
    \textit{(b) The first query detects a white corner in the top center of the image, which hints at the presence of a collar, causing the posterior mass to move to the ``Coat'' and ``Shirt'' categories. Determining between these two categories is relatively difficult however, especially with binary images. But in general, coats tend to be bulkier than shirts. Therefore, after finding a white patch in iteration $k=2$, the probability of ``Coat'' slightly increases, but as more partially black queries along the outside of the shirt are revealed and the slimmer outline of the shirt comes into view, the posterior converges on the correct category of ``Shirt''.}
    
    \begin{center}\includegraphics[height=6cm]{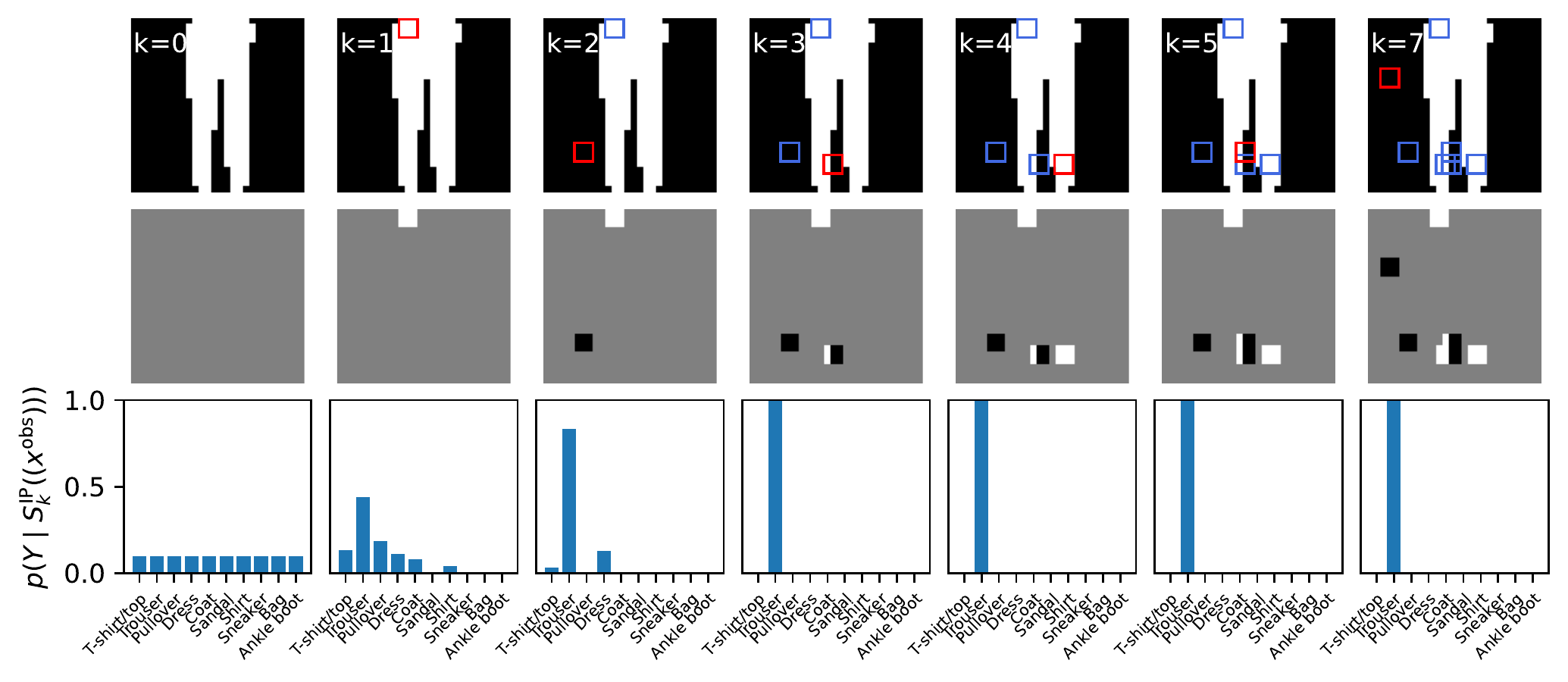}\end{center}
    \textit{(c) Detecting an all white patch in the top center, IP rules out the possibility of the image being some type of shoe. The second query in the lower left returns a black patch, suggesting that the image is also not an upper body garment, which would take up the width of the image. In query $k=3$, IP queries the center bottom of the image, discovering the space in between the two legs of the trouser, causing the posterior probability of ``Trouser'' to jump to nearly 100\%, which remains stable over the last few queries.}
    
    \caption{\textit{Additional Examples of IP on FashionMNIST}}
\end{figure}

\newpage
\subsubsection{IP for Bird Species Classification}
As in the main text, for each plot, on the left, we show the input image and on the right we have a heatmap of the estimated class probabilities per iteration. For readability, we only show the top 10 most probable classes out of the 200. To the right, we display the queries asked at each iteration, with red indicating a ``no'' response and green a ``yes'' response.

\begin{figure}[H]
    \begin{center}\includegraphics[height=5cm]{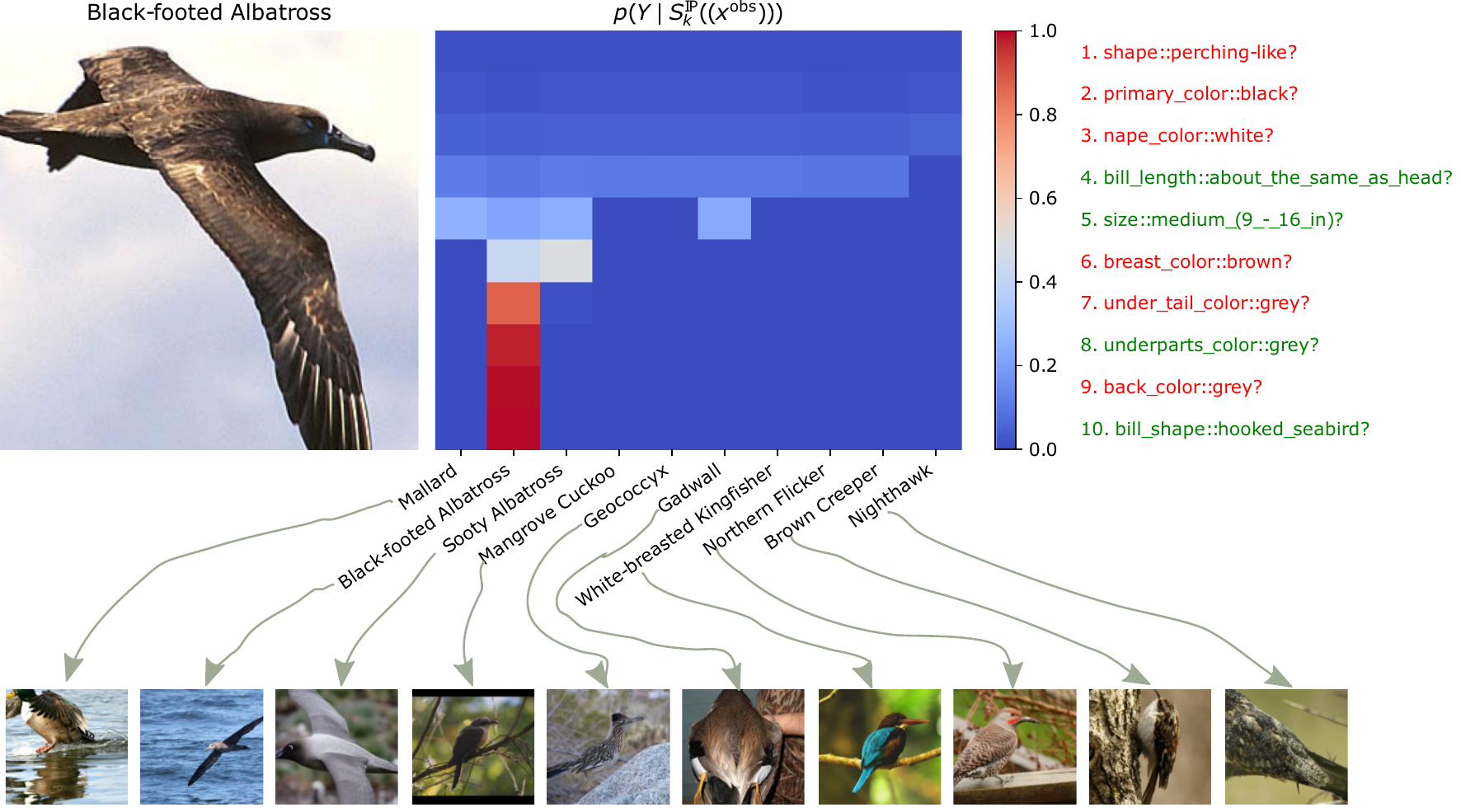}\end{center}
    \textit{(a) The first few queries narrow down the potential species from 200 down to a small number. Since we only show the 10 most probable classes, the first few queries increase the probability of all shown classes. The first queries that distinguish among these classes concern bill length (which rules out the short-billed Nighthawk) and size (which rule out the smaller Mangrove Cuckoo, Geococcyx, Kingfisher, Northern Flicker, and Brown Creeper). The remaining birds are quite similar, all medium-sized brown water birds. The next two color queries suggest that the bird in question is a Black-footed Albatross, which is confirmed by the answers to the next few queries, which all match up with the characteristics of that bird.}
    
    \begin{center}\includegraphics[height=5cm]{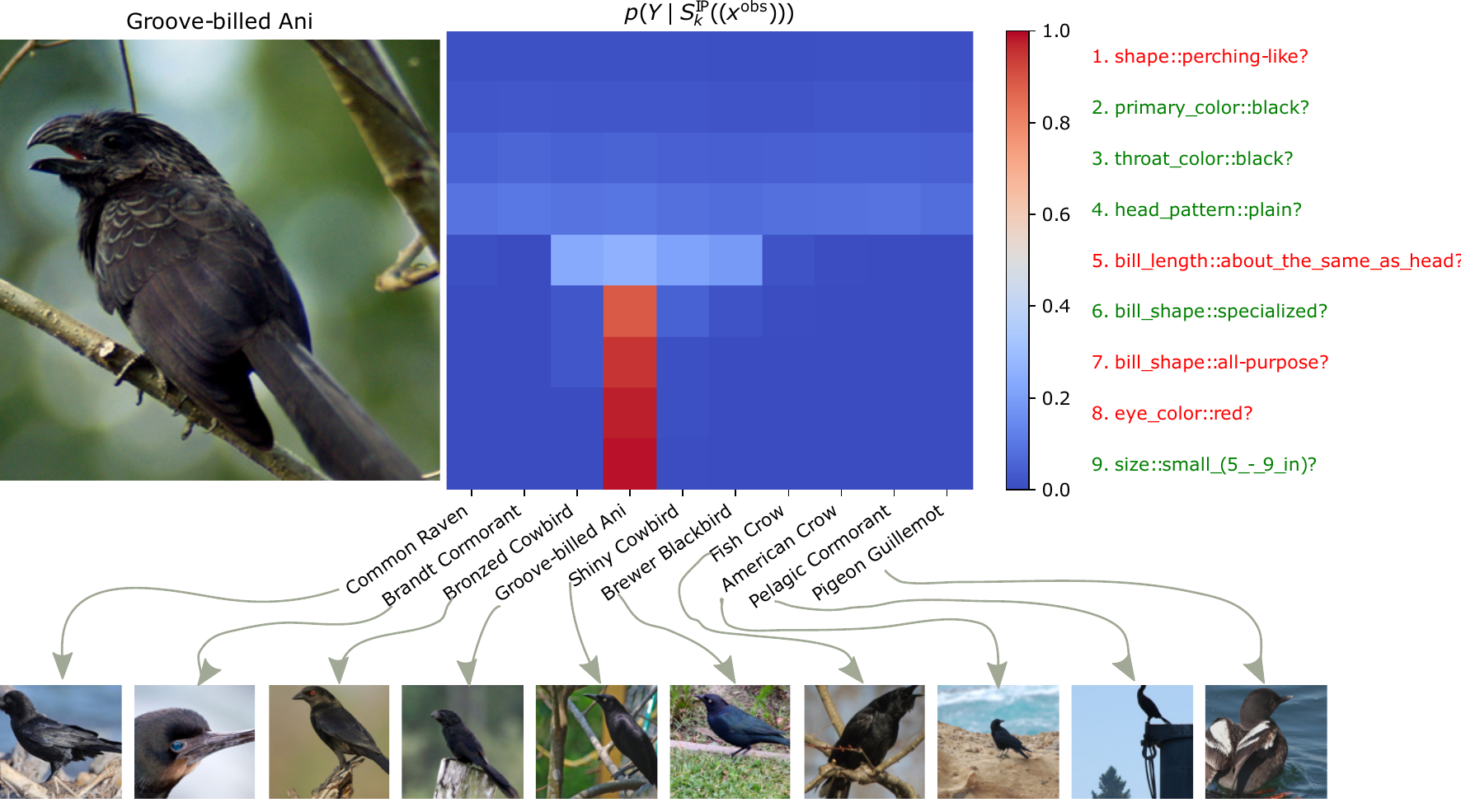}\end{center}
    \textit{(b) Again, the first few queries narrow down the likely species into the top 10 displayed classes. In just two queries (bill length and bill shape), IP distinguishes among these similar-looking, black, plain-headed birds that are hard for non-expert humans to differentiate between. Again, the last few queries serve to confirm the posterior prediction that the bird is a Groove-billed Ani.}
    
    \begin{center}\includegraphics[height=5cm]{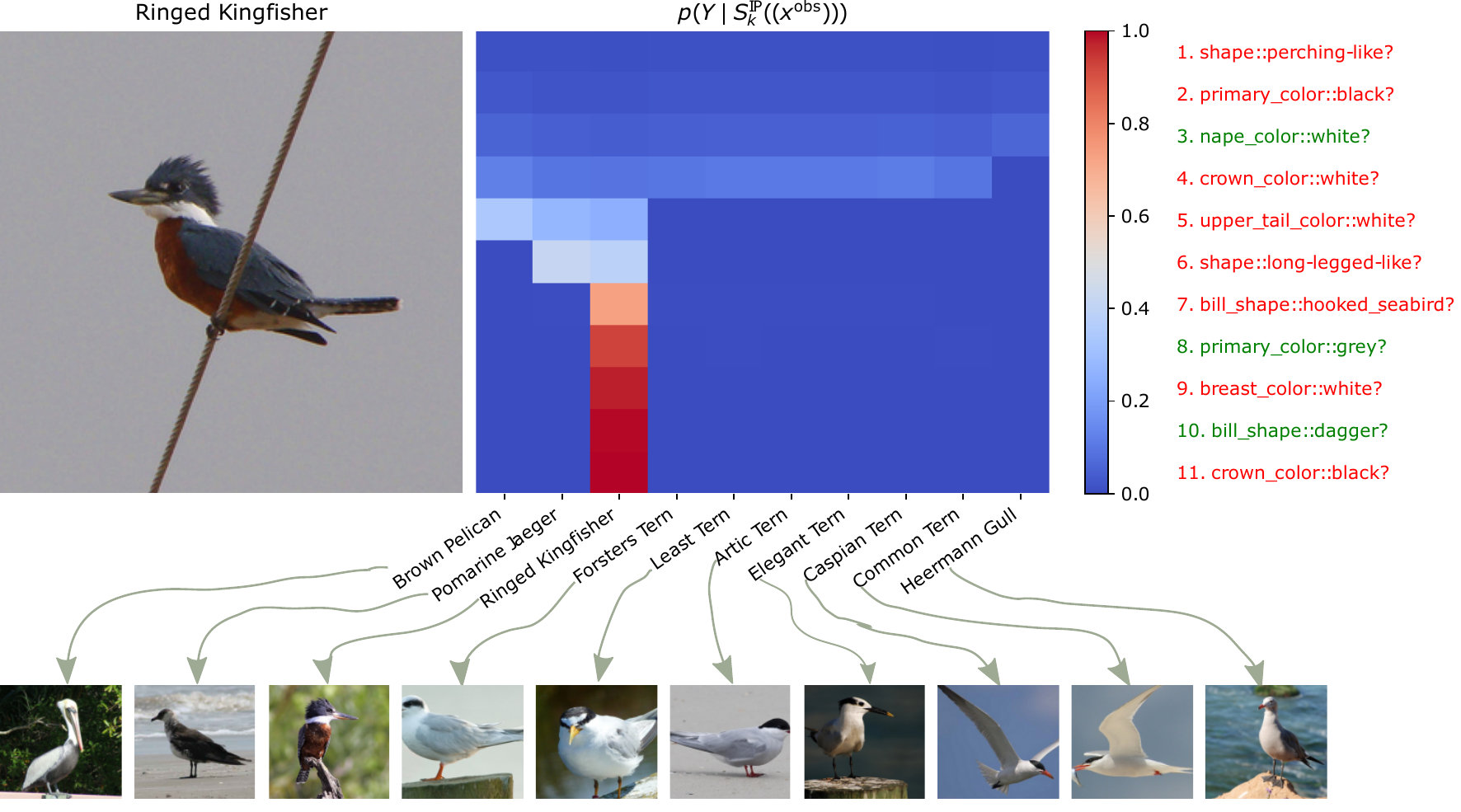}\end{center}
    \textit{(c) After establishing the top few most probable classes, IP converges on the class Ringed Kingfisher after just 7 queries. The last four queries simply serve to increase its confidence in its prediction.}
    
    \caption{\textit{Additional Examples of IP on CUB Bird Species Identification}}
\end{figure}

\subsubsection{IP for HuffPost News Headline Category Classification}

\begin{figure}[H]
    \centering
    \includegraphics[width=\textwidth]{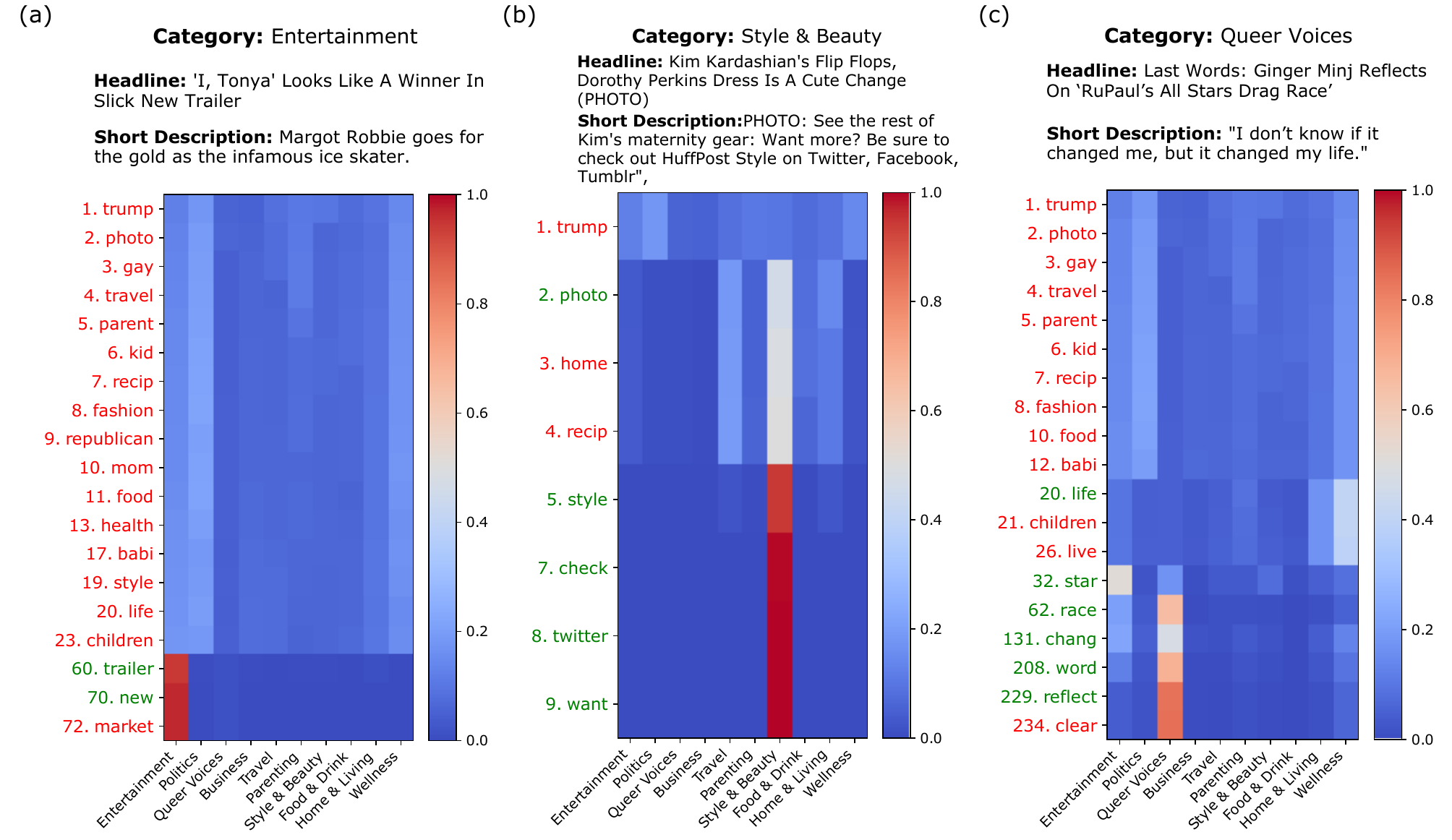}
    \caption{\textbf{Additional Examples of IP on HuffPost News Headline Classification.} As before, when more than 20 queries were asked, we only display the 20 queries that led to greatest KL divergence between successive posteriors.
    \textbf{(a)} Because of the sparse structure of natural language, it typically takes a significant number of queries before the first word that is present in the sentence is found. Until this point, no query is particularly informative, and the posterior distribution remains mostly unchanged from the prior. However, at query 60, IP asks the word ``trailer'', which is present in the extended headline. Naturally, the posterior shifts heavily towards ``Entertainment'', and a few queries later IP reaches its termination criteria. Analyzing this run, we can say that IP reached its decision primarily because of the presence of the word ``trailer'', leading us to say that this is a reasonable and trustworthy prediction.
    \textbf{(b)} This is an example of a relatively short explanation as IP happens to discover words present in the sentence after just two queries. Initially, the presence of ``photo'' causes the categories ``Travel'', ``Home \& Living'', and ``Style \& Beauty'' to become more probable. Several words later however, the word ``style'' is found, which is very strongly associated with the ``Style \& Beauty'' category.
    \textbf{(c)} The posterior remains mostly unchanged until IP discovers the word ``life'', which reasonably, shifts probability mass onto the ``Wellness'' category. However, several queries later, ``star'' is found to be present, which shifts the posterior away from ``Wellness'' onto ``Entertainment'' and ``Queer Voices''. After discovering several more words that are present, ``race'', ``change'', ``word'', ``reflect'', the posterior progressively converges on ``Queer Voices'', but still with relatively high uncertainty, likely because it never came across identifying words such as ``drag'', which was not present in the vocabulary.}
\end{figure}

\end{document}